\documentclass[11pt]{article}
\usepackage{moresize}
\usepackage{graphicx}
\graphicspath{figures}
\usepackage{aux/macros}

\def\wt{\tilde}
\def\n{^{(n)}}
\def\Real{{\mathbb{R}}}
\usepackage[colorinlistoftodos, textsize=tiny]{todonotes}
\makeatletter
\AtBeginDocument{\@mparswitchfalse\normalmarginpar}
\makeatother

\newcommand{\compactmat}[1]{\text{\footnotesize\renewcommand{\arraystretch}{0.5}\setlength{\arraycolsep}{2pt}%
  $\begin{bmatrix}#1\end{bmatrix}$}}
 
\usepackage[margin=1in]{geometry}
\DeclareMathOperator{\clim}{c-lim}
\newcommand{\sqrtm}{^{\nicefrac12}}
\newcommand{\isqrtm}{^{-\nicefrac12}}
\title{
Precise Asymptotics of 
Non-Convex M-Estimators
}
\author{
Shirley Xiaoqi Liu\thanks{Department of Statistics, University of Oxford} 
\qquad
Tirthankar Adhikari\thanks{School of Technology and Computer Science, TIFR Mumbai} 
\qquad
Parthe Pandit\thanks{Center for Machine Learning and Data Science, 
IIT Bombay}~\footnote{Correspondence to \texttt{pandit@iitb.ac.in}}
}
\date{}
\usepackage{caption}
\def\BVloc{\mathrm{BV}_{\sf loc}}
\def\SBVloc{\mathrm{SBV}_{\sf loc}}

\def\op{\mathrm{op}}
\def\dif{\mathrm{d}}

\crefname{enumi}{part}{parts}
\Crefname{enumi}{Part}{Parts}
\Crefname{lemma}{Lemma}{Lemmas}
\crefname{assumption}{assumption}{assumptions}
\Crefname{assumption}{Assumption}{Assumptions}
\AddToHook{cmd/appendix/before}{
\crefalias{section}{appendix}
}

\renewcommand{\bar}[1]{\overline{#1}}

\begin{document}

\maketitle

Precise asymptotic theory for  M-estimators in 
high-dimensional settings  
is largely limited to convex or mildly non-convex  formulations, yet non-convex loss functions as well as regularizers are widely used in practice. We study  M-estimation in generalized linear models (GLMs) where the loss function or the regularizer can be non-convex. Under rotationally invariant design matrices in the proportional asymptotic regime, 
we analyze the problem through  Vector 
Approximate Message Passing (VAMP), 
whose iterates are described   
by a low-dimensional 
deterministic recursion called State Evolution (SE). 
Existing   proofs of SE require the proximal 
operators of the loss function and the regularizer to be Lipschitz, but a non-convex loss or regularizer can have a discontinuous proximal operator, and the usual Onsager terms in VAMP then miss the contribution of the discontinuities.
Assuming that the proximal operators are special functions of locally bounded variation, 
we derive a correction to the Onsager terms using a distributional Stein's lemma. 
We prove that    SE holds for the corrected VAMP. 
We apply the theory 
to analyze group sparse regression with an $\ell_0$-type penalty 
and to robust regression with the  Cauchy loss.

\tableofcontents

\section{Introduction}

Generalized linear models (GLMs) are a fundamental class of models   in statistics and machine learning. In these models, for a design matrix $\Psi\in \mathbb{R}^{m\times p}$ and we observe $m$ measurements $y=(y_1, \ldots, y_m)\in \mathbb{R}^{m\times d}$, where each measurement $y_i\in\Real^d$ depends on an unknown signal 
$\theta_0\in \mathbb{R}^{p\times d}$
only through the corresponding row of the linear predictor $\beta_0=\Psi \theta_0\in \mathbb{R}^{m\times d}$, together with noise that is independent across measurements. That is, $y_i$ is drawn from a likelihood $$p(y_i\,|\,  (\beta_0)_i)\quad \text{with}\quad \beta_0=\Psi\theta_0$$ independently over $i$, given $\beta_0$.  

A standard way to estimate the unknown signal $\theta_0$ is to minimize a penalized loss: 
\begin{align}\label{opt:theta_beta_Phi}
    \widehat{\theta}\in \underset{\theta\in\Real^{p\times d}}{\arg\min}~\sum_{i=1}^m \ell_i(\beta_i) + \sum_{j=1}^p \pen_j(\theta_j)\qquad\text{subject to }\beta=\Psi\theta\,,
\end{align}
where  $\ell_i:\Real^d \to\Real$ is a loss function, $ \theta_j$ is the $j^{\rm th}$  row of $\theta$, and $\pen_j:\Real^d\to\Real$ is a regularization function. Estimators defined as minimizers of such objectives are called M-estimators, i.e., maximum likelihood-type estimators. Indeed, choosing $\ell_i(\beta_i)=-\log p(y_i\,|\,  \beta_i)$ gives penalized maximum likelihood. However, both the loss $\ell_i$ and the regularizer $\pen_j$ are design choices: the loss need not be the true likelihood and may instead be chosen to achieve other properties, such as  robustness to outliers via the Huber or Cauchy loss, 
while the regularizer  encodes structure we expect in $\theta_0$, such as sparsity via the $\ell_1$ or $\ell_0$-penalty. 

The examples of the Cauchy loss and the $\ell_0$ penalty involve solving a \emph{non-convex} problems which are generally less well understood than convex ones. In this paper, we allow {either} (i) $\ell_i$ is non-convex and $\pen_j(t)=\norm{t}^{2}/2$, \textit{or} (ii) $\ell_i$ is the square loss and $\pen_j$ is non-convex. Our framework can handle the case where both the loss function and the regularizer are non-convex, as well as the multi-layer VAMP \cite{pandit2019asymptotics,pandit2020matrix} with some additional book-keeping. However we present the minimal versions in this paper and defer the discussions about these extensions for a future version.

 We study the optimization problem in~\eqref{opt:theta_beta_Phi} through  the Vector Approximate Message Passing (VAMP) algorithm  \cite{rangan2019vector}.\footnote{More precisely, we study matrix-VAMP from \cite{pandit2020matrix} that handles matrix- or tensor-valued variables $\theta $ and $\beta$.} VAMP is an iterative algorithm that alternates between a step that handles the loss and a step that handles the regularizer, where the two steps are coupled through the singular vectors of the design~matrix~$\Psi$. Crucially, the fixed points of VAMP are stationary points of \eqref{opt:theta_beta_Phi}. 
When $\Psi$ is drawn from a rotationally invariant ensemble, 
the  statistical behavior of the  VAMP iterates, in the limit of the problem size $m,p\to \infty$ with $m/p$ fixed, can be precisely described  by a low-dimensional deterministic recursion
called  State Evolution (SE).  
Moreover, VAMP belongs to the  family of Approximate Message Passing (AMP) algorithms
 \cite{donoho2009message}; see   \cite{feng2022unifying} for a recent tutorial. 
 While the SE of VAMP holds for  rotationally invariant designs with arbitrary singular value spectra, the SE of standard AMP is restricted to i.i.d.\ Gaussian designs, and  the Orthogonal AMP  \cite{ma2017orthogonal} is closer to VAMP.

In each iteration, VAMP applies the proximal operator of   the loss $\ell_i$ in one step and that of the regularizer  $\pen_j$ in another.  Following the VAMP literature, we call these maps \emph{denoisers}.  Existing  proofs of  the  validity of SE require  the denoisers to be   Lipschitz continuous, which is guaranteed   when  $\ell_i$ and $\pen_j$  are convex.    When they are non-convex, however, the denoisers  can be discontinuous, 
which invalidates the   so-called \emph{Onsager terms} in VAMP. 
Each Onsager  term,  computed from  averaging the Jacobian  of its associated denoiser,   ensures the input to the other   denoiser behaves like a scaled version of the signal plus independent Gaussian noise, a property that the low-dimensional SE can  track. A discontinuous  denoiser still has a Jacobian away from its jumps, but averaging this Jacobian misses the effect of the jumps. 
The standard Onsager terms are therefore incorrect, and  the standard SE is no longer a valid description of the asymptotic properties of VAMP when the denoisers are discontinuous. 

\begin{figure}[h]
\centering
\begin{subfigure}{\linewidth}
\begin{minipage}[c]{0.55\linewidth}
    \centering
    \caption{Cauchy loss, $d=2$, $\gamma=0.1\compactmat{
1&0.3\\0.3&1.5
}$. 
Axes are the  two  coordinates of the input. 
   The jump set consists of 1D curves. For isotropic $\gamma\!=\!\gamma_0I_2$ with $\gamma_0\!<\!\frac{1}{8}$, the jump set is instead a circle 
(\Cref{section:correction cauchy}).}
    \label{fig:cauchy-jump}
\end{minipage}\hfill
\begin{minipage}[c]{0.20\linewidth}
\includegraphics[width=\linewidth]{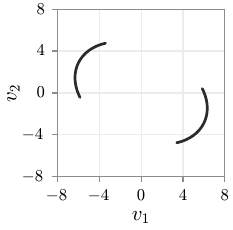}
\end{minipage}\hspace{0.13\linewidth}
\end{subfigure}
\begin{subfigure}{\linewidth}
\begin{minipage}[c]{0.55\linewidth}
    \centering
    \caption{Rank penalty, 
    $d=3$, $\gamma=
\compactmat{
1 & 0.20 & 0.12\\
0.20 & 1.60 & 0.25\\
0.12 & 0.25 & 0.65
}
$. %
    Axes are the coordinates $(t,x,y)$  parameterizing matrices 
$\compactmat{t+x&y\\y&t-x }$.  The jump set consists of  2D surfaces; the legend gives the ranks of the output on either side of the jump. For isotropic $\gamma $,   the proximal operator is  singular value hard thresholding:  only the blue and orange jump surfaces occur. 
}
    \label{fig:evt-jump}
\end{minipage}\hfill
\begin{minipage}[c]{0.44\linewidth}
\includegraphics[width=\linewidth]{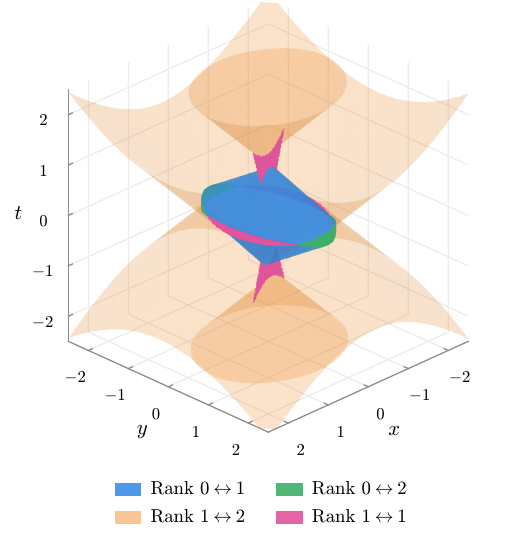}
\end{minipage}
\end{subfigure}
\label{fig:jumps}
\vspace{-0.2cm}
 \caption{Jump sets, i.e., the inputs where a proximal 
operator is discontinuous, for (a) the Cauchy loss with inputs in $\Real^2$, and (b) rank penalty on  $2\!\,\times\!\, 2$ symmetric matrices, a 3D space.  Each jump set has one dimension fewer than the input 
space, and   the Onsager term   
integrates over it. In both,  the weight matrix $\gamma$ on the quadratic term of the proximal operator (see \Cref{def:prox})   is   non-isotropic.}
\vspace{-0.5cm}
\end{figure}

Moreover, for multivariate denoisers, i.e., $d>1$, 
these discontinuities are not points but curves and surfaces, whose geometry can be complicated.  \Cref{fig:jumps} shows two examples. For the Cauchy loss, used in robust regression, 
the proximal operator on  $\Real^2$ jumps across curves (\Cref{fig:cauchy-jump}). For the rank penalty on $2\!\,\times \!\,2$ symmetric matrices, a 3D space, it jumps across surfaces  made of several pieces, distinguished by the ranks of the output on either side of the jump (\Cref{fig:evt-jump}). A valid Onsager term must account for the contribution of these jumps.

\paragraph{Main contributions.} In this paper, we extend  VAMP and SE to    discontinuous denoisers, 
by  adding the contribution of the jumps  to the Onsager terms as a correction. A recent paper \cite{ranasinghe2026state}  has studied  an analogous problem  for AMP with scalar denoisers, i.e., $d=1$, using a  correction  that  requires  knowledge of the signal   prior. 
In contrast, we study VAMP in  the general  $d>1$ setting. Our updated VAMP algorithm  requires no   signal prior and is   a standalone solver  for  non-convex  problems of the  form  \eqref{opt:theta_beta_Phi}. More specifically, we make the following contributions:

\begin{itemize}
    \item 
First, unlike the $d=1$ case, 
the correction terms for $d>1$ become integrals over jump surfaces 
with respect to the surface measure, or   formally, the $(d-1)$-dimensional Hausdorff measure. 
We derive  the correction terms  from  a distributional version of Stein's lemma (\Cref{lemma:stein}), which generalizes the scalar result of \cite[Lemma 4]{tibshirani2015degrees} used in  \cite{ranasinghe2026state}. 
This lemma applies to denoisers that belong to a class of   special functions of locally bounded variation \cite{alberti1997note} ($\SBVloc$,    \Cref{def:SBVloc}), whose
 distributional derivative  consists of the  almost-everywhere Jacobian    and a jump part on its  discontinuity surfaces,
 with no Cantor part. We  discuss this  in detail in \Cref{sec:correction_onsager} and  \Cref{section:sbvloc}.

\item 
Second, the correction term of \cite{ranasinghe2026state}  uses the  density  
of the input to the discontinuous denoiser,  computed from a known prior.
This is natural in the main application of \cite{ranasinghe2026state}, MIMO detection, where the signal distribution is known. 
In the M-estimation setting we study, however, the denoisers are proximal operators of the loss $\ell_i$ and regularizer $\pen_j$,   and the distribution of the true signal $\theta_0$ is unknown in practice.  A correction that requires a prior  would therefore  prevent VAMP from serving as a standalone solver for problem~\eqref{opt:theta_beta_Phi}.

To overcome this, we   estimate the input density with a Gaussian kernel  density estimator (KDE)   
over the rows of the input to the denoiser. 
A similar kernel estimate of the density at the jumps was used by  \cite{deledalle2013stein} to estimate the risk of scalar hard thresholding.  Our setting is more demanding because (i) the KDE is recomputed at every VAMP iteration from rows that need not be independent, and (ii) the  correction integrates the estimated density over the jump surfaces,  which may move across iterations, weighted by the jump of the denoiser and the surface normal.

\item 
Finally, while \cite{ranasinghe2026state} proves a valid SE for AMP with i.i.d.\ Gaussian designs and entrywise-separable denoisers fixed in advance, we prove an analogous SE (\Cref{theorem:state evolution}) for   our updated VAMP with rotationally invariant designs, allowing denoisers that act jointly on the $d$ entries of each row 
or jump surfaces that are data-dependent rather than   fixed in advance. 
The key new proof step is that the KDE-based corrected Onsager term converges to its SE counterpart (\Cref{lemma:consistency of onsager terms}). 

\end{itemize}

\paragraph{Related work.} Much work on AMP and VAMP uses  posterior-mean (Bayes-optimal) denoisers, where each denoiser is the minimum mean-square-error (MMSE) estimate of the signal from a Gaussian-corrupted observation, computed from a known prior, e.g.,  \cite{barbier2019optimal, rangan2019vector, zdeborova2016statistical, montanari2021estimation, liu2026many}. 
Such denoisers do \emph{not} have discontinuities, since the posterior mean under Gaussian noise is a smooth function of its input, regardless of the prior. 

Existing analyses of M-estimation via AMP and VAMP, in contrast, require the  proximal operators to be  Lipschitz. This holds for convex losses and penalties, e.g.,  \cite{Bayati2012lasso, donoho2016high, gerbelot2020asymptotic},  
while  \cite{rangan2011generalized} argues that SE can also hold for non-convex problems under this Lipschitz condition. A recent analysis   of non-convex single-index models ($d=1$) \cite{vilucchio2025asymptotics} makes the condition more explicit. The Lipschitz proximal operator assumption limits how non-convex the loss and regularizer can be relative to a scale set by the solution (see \cite[Assumption 2.4]{vilucchio2025asymptotics}). 
This condition excludes the discontinuous proximal operators we study.

Previous efforts at dealing with discontinuous proximal operators such as those of $\ell_p$ penalties for $p\in[0,1)$ have  smoothed the proximal operator itself to establish the SE. For example, 
\cite{zheng2017ellp} establishes state evolution for AMP with a smoothed proximal map by taking the large-system limit before sending the smoothing-bandwidth to zero. In contrast, we establish SE for VAMP with   discontinuous proximal operator itself, using smoothing only to estimate the Onsager term.

 \paragraph{Organization of the paper.}

\Cref{sec:preliminaries} introduces  notation and the space $\SBVloc$ of discontinuous functions we work with. \Cref{section:vamp} describes VAMP with the corrected Onsager term  and shows that the fixed points of VAMP are stationary points of our optimization problem of interest.
\Cref{section:examples} applies     VAMP     to two non-convex M-estimation problems, group sparse regression and Cauchy regression, along with numerical experiments.  \Cref{section:state evolution} states our main result, \Cref{theorem:state evolution}, which establishes the validity of the State Evolution, and  \Cref{section:proof sketch} presents its proof. The key technical contribution is \Cref{lemma:consistency of onsager terms} which shows the KDE-based corrected Onsager term converges to its SE counterpart.

\section{Preliminaries}\label{sec:preliminaries}

\subsection{Notation}\label{sec:notation}
\paragraph{Spaces and   operators:}   $(E,\inner{\cdot,\cdot})$  denotes a Euclidean space, e.g., $\Real^{d}$, $\Real^{d_1\times d_2}$, or a space of higher-order tensors, and $d:=\dim(E)$. We refer to elements of $E$ as tensors even if they may be vectors or matrices. For $x\in E, \norm{x}=\sqrt{\inner{x,x}}$ is the standard Frobenius norm.
Where the space is not clear from context, we indicate it by a subscript, e.g., $\norm{\cdot}_{E}$ for the norm on $E$; $0_E$ and $I_E$ for the zero tensor and identity operator on $E$; and $0_{ d} $ 
and   $I_d$  for the zero vector in $\Real^{ d}$ 
and    identity matrix in $\Real^{d\times d}$.

For (possibly different) Euclidean spaces $E_1,E_2,E_3$, we denote by $L(E_1;E_2)$ the space of bounded linear operators  $E_1\to E_2$. For $E_1 = \mathbb R^d$ and $E_2 = \mathbb R^{d'}$, these are the $d\times d'$ matrices.  We regard linear operators  as acting from the \emph{right}: for $u\in E_1$ and $A\in L(E_1; E_2)$ we write  $uA:= A(u) \in E_2$, so that  $u(AB)=(uA)B$ for  $B\in L(E_2; E_3)$. The adjoint  $A^*\in L(E_2;E_1)$ is defined by $\inner{uA, w}_{E_2} =\inner{u, wA^*}_{E_1} $ for all $u\in E_1$ and $w\in E_2$. For $a\in E_1$ and $b\in E_2$,   the rank-one operator $a\otimes b\in L(E_1;E_2)$ is defined by
$u(a\otimes b):=\inner{u,a}_{E_1} b $ for all $u\in E_1$, and   $a^{\otimes 2}:= a\otimes a \in L(E_1;E_1)$.

We denote by $\mc E_+\subset L(E;E)$  the set of self-adjoint positive definite operators on $E$, i.e., symmetric $\delta$ with $\inner{v\delta, v}=\inner{v,v\delta}>0$ for all $v\in E\backslash \{0_E\}$, and  by  $\mc E_\ge\subset L(E;E)$  the set of self-adjoint positive semidefinite operators on $E$. 
For $\delta\in \mc{E}_+$ and $v\in E$, we write   $\norm{v}_\delta^2:=\inner{v\delta, v}$. 
For $\delta\in\mc E_+$ and $a,b\in E$, operators compose as  $(a\otimes b)\delta=a\otimes (b\delta)$ and $\delta (a\otimes b)=(a\delta)\otimes b.$ When $E=\Real^d$, we write $\mathbb{R}^{d\times d}_\succ := \mathcal{E}_+$ and $\mathbb{R}^{d\times d}_\succeq := \mathcal{E}_\geq$ for the sets of symmetric positive definite and positive semidefinite $d\times d$ matrices, respectively.

We write function spaces of tensor fields in  $(\text{domain};\text{codomain})$ format. For example,   $g\in\SBVloc(E_1;E_2)$  (see \Cref{def:SBVloc} below) is a function $g:E_1\to E_2$, and its derivative $\Dif g\in \mc M_{\sf loc}(E_1;L(E_1;E_2))$ is a locally finite measure on $E_1$ that takes values in  $L(E_1;E_2)$ so that $\Dif g(A)\in L(E_1;E_2)$ for every bounded measurable $A\subset E$. 

\paragraph{Indexing:} 
An element $x \in E^n$ is an array with $n$ rows $x_i \in E$; for $E = \mathbb R^d$, $E^n \equiv \mathbb R^{n\times d}$. For $x \in E^n$ and $\delta \in \mathcal E_+$, we write $\norm{x}_\delta^2 := \sum_{i=1}^n \norm{x_i}_\delta^2$ and $x\otimes x:=\sum_{i=1}^n x_i\otimes x_i\in L(E;E)$.
For $n\in \mathbb{N}\backslash \{0\}$, write  $[n]:=\{1, 2, \ldots, n\}$.  Subscripts denote both row  index and iteration index. We reserve $i\in[n]$ for coordinate indices, and $t,j\in\Natural$ for iteration indices. 
For $x \in E^n$, we denote by $x_i \in E$ its $i^{\rm th}$ row. For an iterate $x_t \in E^n$, we write $(x_t)_i \in E$ for its $i^{\rm th}$ row.

\paragraph{Probability:} For $\tau\in\mc E_+$, we denote by $\varphi_\tau:E\to\Real$ the  density function of a centered Gaussian $E$-valued random variable with covariance matrix $\tau$, i.e., $\varphi_\tau(v):=\det (2\pi \tau)\isqrtm\exp\round{- \norm{v}_{\tau\inv}^2/2}$ for all $v\in E$. Its gradient is  $\mathrm{D}\varphi_\tau(v) = -v\tau^{-1}\varphi_\tau(v)$. 
For an $E$-valued $X$ and an $E_2$-valued $Y$, $\operatorname{Cov}(X, Y) := \Exp\big[(X - \Exp X) \otimes (Y - \Exp Y)\big] \in L(E_1;E_2)$ and $\Exp X^{\otimes 2}:=\operatorname{Cov}(X) := \operatorname{Cov}(X, X)\in\mc E_\ge.$

\subsection{Proximal operators}
The denoisers of the VAMP algorithm are proximal operators, in which the quadratic term is weighted by a
precision matrix that VAMP updates across iterations. To this end, we provide the following definition.
\begin{definition}[Variable-metric proximal operator]\label{def:prox} Let $f:E\to(-\infty, +\infty]$  be proper, lower semicontinuous and bounded below
and  let  $\gamma \in \mc E_+$,  define $\prox_f(\cdot,\gamma):E\to E$ by 
\begin{align*}
\prox_f(u,\gamma)\in \underset{v\in E}{\arg\min}~\tfrac12\norm{v-u}_\gamma^2 + f(v)\qquad \text{for all}\quad  u\in E,
\end{align*}
where, if the minimizer  is not unique, $\prox_f(u, \gamma)$ is  a fixed measurable selection. 
The same definition applies on $E^n$, with $\norm{x}_\gamma^2 = \sum_{i=1}^n \norm{x_i}_\gamma^2$. That is, if  $F:E^n\to (-\infty, +\infty]$ satisfies $F(x)=\sum_{i=1}^n F_i(x_i)$, then $\prox_F(\cdot,\gamma):E^n\to E^n$ satisfies  $(\prox_F(u,\gamma))_i = \prox_{F_i}(u_i,\gamma)$ for~all~$u\in E^n$.
\end{definition}

\subsection{\texorpdfstring{$\SBVloc$}{SBVloc}: Special functions of locally bounded variation}\label{sec:SBV_main_text}

When the denoisers are the proximal operators of non-convex functions, they are typically smooth except across surfaces where they have discontinuities, i.e., jumps. We therefore  work with their distributional derivative, which involves both the Jacobian and the jumps. This section lays down  the notation for it, which  \Cref{sec:correction_onsager} uses to define the corrected Onsager term.

Let $\mc L$ denote Lebesgue measure on $E_1$, and  let 
$\mc H$ denote the $(d-1)$-dimensional Hausdorff measure, which measures the surface area of $(d-1)$-dimensional sets.  A function $\zeta \in L^1_{\sf{loc}}(E_1;E_2)$, i.e., a locally integrable function, has locally  bounded variation, written $\zeta\in \BVloc(E_1;E_2)$, if its   distributional derivative $\mathrm D\zeta\in\mc M_{\sf loc}(E_1;L(E_1;L_2))$ is a locally finite $L(E_1;E_2)$-valued measure on $E_1$.  That is, for every bounded Borel set $A \subset E_1$, $\mathrm D\zeta(A) \in L(E_1;E_2)$ is well defined and its total variation $|\mathrm D\zeta|(A)$ is finite. 
To describe $\mathrm D\zeta$, we first locate the jumps of $\zeta$.
\begin{definition}[Jump set]\label{def:jump_set}
Let $\zeta\in \BVloc(E_1; E_2)$. 
 A point $v\in E_1$ belongs to the jump set $J_\zeta\subset E_1$ if  there exist  a unit tensor
$\mathbf n_\zeta(v)\in E_1$ and distinct tensors
$\zeta^+(v),\zeta^-(v)\in E_2$  such that 
\begin{equation*}
    \lim_{\rho\downarrow0}
    \frac{1}{\mc L(\mathbb{B}_\rho^\pm(v))}
    \int_{\mathbb{B}_\rho^\pm(v)}
    \norm{\zeta(u)-\zeta^\pm(v)}_{E_2}\,
    \dif u
    =0,
\end{equation*}
where 
\[
    \mathbb{B}_\rho^\pm(v)
    :=
    \left\{
        u\in E_1:
        \norm{u-v}_{E_1}<\rho,\quad
        \pm\inner{ u-v,\mathbf n_\zeta(v)}_{E_1}>0
    \right\}
\]
is the  half-ball on the $\pm\mathbf n_\zeta(v)$ side of $v$. That is, as $\rho\downarrow 0$, on average $\zeta(u)$ is close to $\zeta^+(v)$ on one half and to $\zeta^-(v)$ on the other. 
We call  $\zeta^\pm(v)$   the
 one-sided  traces,
$\mathbf n_\zeta(v)$   the    jump normal,
and
\[
    [\zeta](v):=\zeta^+(v)-\zeta^-(v)
\]
the jump. They are unique up to 
 the simultaneous replacement
$(\zeta^+,\zeta^-,\mathbf n_\zeta)
    \mapsto
    (\zeta^-,\zeta^+,-\mathbf n_\zeta)$
which leaves $\mathbf n_\zeta\otimes[\zeta]$ unchanged.
\end{definition}
\begin{remark}[The jump part is well defined]
    The BV trace and structure theorems ensure that $J_\zeta$ is countably $(d-1)$-rectifiable and that
the traces $\zeta^{\pm }$ and normal $\mathbf n_\zeta$ can be chosen measurably, so $\mathrm D^{\mathsf j}\zeta$ below is well defined;  see
\cite[Theorems~3.77--3.78 and Proposition~3.92]
{ambrosio2000functions}.
\end{remark}

The derivative of $\zeta\in \BVloc$  splits into three mutually singular parts,
\[
\mathrm{D}\zeta = \mathrm{D}^{\sf{a}}\zeta + \mathrm{D}^{\sf{j}}\zeta + \mathrm{D}^{\sf c}\zeta\in\mc M_{\sf loc}(E_1;L(E_1;E_2))
\]
where the  absolutely continuous part  $\mathrm{D}^{\sf a}\zeta=\zeta'\mc L$ involves the   $\mc{L}$-almost everywhere Jacobian $\zeta'$ of $\zeta$, the  jump part $\mathrm{D}^{\sf j}\zeta$    is concentrated on the jump set $J_\zeta$,  
and the Cantor part $\mathrm{D}^{\sf c}\zeta$ is the remainder.   
The functions we focus on in this work have no Cantor part and are called  special functions of locally bounded variation, denoted $\SBVloc$. Note that $\SBVloc$ is a subspace of $\BVloc$.

\begin{definition}[$\SBVloc$ functions]
\label{def:SBVloc}   
For every bounded Borel set $A\subset E_1$, define the Jacobian part and jump part of $\Dif \zeta$, of a $\zeta\in\BVloc(E_1;E_2)$ by 
\[  
\Dif^{\sf a}\zeta (A):=\int_A \zeta'(v)\mc L(\dif v), \quad \Dif^{\sf j}\zeta (A):=
\int_{ A\cap J_\zeta}
\mathbf n_\zeta(v)\otimes[\zeta](v)\,
\mc H(\dif v),\]
and the Cantor part $\Dif^{\sf c}\zeta:= \Dif \zeta-\Dif^{\sf a}\zeta - \Dif^{\sf j}\zeta$. 
Then $\zeta$ is a special function of locally bounded variation, written $\zeta \in \SBVloc(E_1;E_2)$,   if  $\mathrm{D}^{\sf c}\zeta=0$; equivalently,  
$ \mathrm{D}\zeta(A)
= \Dif^{\sf a}\zeta (A) +    \Dif^{\sf j}\zeta (A)
$
for every bounded Borel set $A\subset E_1$.

\end{definition}

We provide a discussion on  $\SBVloc$  in \Cref{section:sbvloc}; a nice reference for the $\SBVloc$ space is \cite{alberti1997note}.

\section{The Vector Approximate Message Passing (VAMP) Algorithm}\label{section:vamp}
 VAMP   applies to optimization problems of the general form
\begin{align}\label{opt:xzU}
\min_{x,z \in E^n} F_1(x) + F_2(z)\quad \text{subject to}\quad  z=U\tran x 
\end{align}
over the  variables $x$ and $z$, where $U\in\Real^{n\times n}$ is   orthogonal  and $F_1,F_2:E^n\to(-\infty, +\infty]$ are row-wise  separable, i.e.,   $F_\ell(x) =\sum_{i=1}^n F_{\ell i}(x_i)$  for $\ell\in \{1,2\}$, where $F_{\ell i}:E\to (-\infty, +\infty]$  is applied to the $i^{\rm th}$ row $x_i$ of $x$. 
VAMP alternates between two steps, one for   $F_1$ and the other for $F_2$, by applying its proximal operator. Between the two steps $(r_t,q_t)$, \textit{corrected} versions of the current estimate $(x_t, z_t)$ are passed from one proximal operator to the other via the orthogonal matrix $U$.

  \Cref{section:examples} discusses how different instances of problem \eqref{opt:theta_beta_Phi} can   be rewritten as problem \eqref{opt:xzU}.  The rest of this section describes VAMP, given in \Cref{algo:vamp}, for finding stationary points of  \eqref{opt:xzU}.

  \subsection{Denoisers}
VAMP uses $F_1$ and $F_2$ through their proximal operators, 
$g:E^n\to E^n$ and $h:E^n\to E^n$, defined as 
\begin{subequations}
\begin{align}
    g(r,\gamma) &:=\prox_{F_1}(r,\gamma) 
    \in\underset{x\in E^n}{\arg\min}\, \tfrac12\norm{x-r}_\gamma^2 + F_1(x), \qquad r\in E^n, \gamma\in \mc{E}_+ \label{def:g_prox}\\
   h(q,\delta) &:=\prox_{F_2}(q,\delta) \in\underset{z\in E^n}{\arg\min} \, \tfrac12\norm{z-q}_\delta^2 + F_2(z),\qquad q\in E^n, \delta\in\mc{E}_+.\label{def:h_prox}
\end{align}
\end{subequations}
VAMP alternates between the two denoisers
with a rotation by $U$ or $U^\top$ in between (\Cref{fig:vamp_se}). 
Since  $F_1$ and $F_2$ are row-wise separable,
the denoisers also act row-wise:
\begin{align*}
[g(r,\gamma)]_i=g_i(r_i,\gamma)=\prox_{F_{1i}}(r_i,\gamma), \qquad [h(q,\delta)]_i=h_i(q_i,\delta)=\prox_{F_{2i}}(q_i,\delta), \quad i\in[n].
\end{align*}
The row-wise functions $F_{1i}$ and $F_{2i}$ differ across the index  $i$ only through known side information $w_i\in E_w$ and $s_i\in E_s$, such as an observation $y_i$ or a singular value $s_i$.  Writing $w=(w_i)^n_{i=1} \in E_w^n$ and $s=(s_i)_{i=1}^n\in E_s^n$, we make this precise as follows.
\begin{assumption}[Problem structure]\label{assumption:structure}
Assume $F_2:E^n\to(\infty, +\infty]$ is closed proper convex.
Let $g:=\prox_{F_1}$ and $h:=\prox_{F_2}$ as in \eqref{def:g_prox}--\eqref{def:h_prox}. There exist maps $G: E\times \mc{E}_+\times E_w\to E$, $H: E\times \mc{E}_+\times E_s\to E$, and vectors $w\in E_w^n,$ and $s\in E_s^n$ such that   
\begin{align*}
    &g_i(v,\gamma) = G(v,\gamma,w_i)\quad \text{for all }v\in E,\gamma\in \mc E_+\,,\\
    &h_i(v,\delta) = H(v,\delta,s_i)\quad \text{for all }v\in E,\delta\in \mc E_+\,,
\end{align*}
and $G(\cdot,\gamma,u_w)\in\SBVloc$ for all $\gamma\in\mc E_+,u_w\in E_w$. 
\end{assumption}

In VAMP,  after each denoising step it forms an extrinsic estimate ($\widetilde r_t$ or $\widetilde q_t$), weighted by an Onsager term ($\alpha_t + c_t$ or $\beta_t$), which removes the part of the output that is correlated with the noise in the denoiser  input. As a result, in the large-system limit each denoiser input   behaves   like a linear transformation of the signal  plus independent Gaussian noise, whose precision matrix  is tracked by $\gamma_t$ or $\delta_t$.  The two half-steps are repeated, updating the precision matrices $\gamma_t$ and $\delta_t$ that weight the proximal operators, until the iterates stop changing.  VAMP returns the denoiser outputs at the fixed point as its estimates $(\hat x, \hat z)$. \Cref{sec:fixed_point_VAMP}  shows that these estimates  satisfy the KKT~conditions~of~\eqref{opt:xzU}.

\begin{figure}[!tp]
\centering
\begin{tabular}{p{0.46\textwidth}| p{0.46\textwidth}}
\hline
\vspace{-1.5ex}
\captionof{algorithm}{\label{algo:vamp} 
\textbf{VAMP}}
& 
\vspace{-1.5ex}
\captionof{algorithm}{\label{algo:se} 
\textbf{State Evolution}}\\
\hline
\vspace{-1.5ex}
\begin{tikzpicture}
[
>=latex,
block/.style={draw,ultra thick,minimum width=1cm,minimum height=0.7cm},
circ/.style={draw,circle,minimum width=.5cm},
tinycircle/.style={
circle,draw,fill=none,
inner sep=0pt,outer sep=0pt,
minimum size=2mm,line width=0.2pt,
},
var/.style={coordinate}
]
\node[anchor=north east] at (3.7,2.2) {input side};
\node[anchor=south west] at (-3.7,-2.2) {output side};

\node[var] (q) at (-2,1.6) {};
\node[var] (eta) at (2.8,1) {};
\node[var] (nu) at (-2.8,-1) {};
\node[var] (gamma) at (2.8,-1) {};
\node[var] (delta) at (-2.8,1) {};
\node[var] (gamma2) at (.5,-1) {};
\node[var] (delta2) at (-.5,1) {};
\fill (gamma2) circle[radius=1pt];
\fill (delta2) circle[radius=1pt];
\node[block] (G) at (3,0) {$g$};
\node[tinycircle] (o1) at (2,1.6) {};
\node[tinycircle] (o2) at (-2,-1.6) {};
\node[var] (r) at (2,-1.6) {};
\node[block] (H) at (-3,0) {$h$};
\node[block] (VT) at (0,1.6) {$U\tran$};
\node[block] (V) at (0,-1.6) {$U$};
\node[tinycircle] (Vscalar) at (V|-nu) {\ssmall \textbf{-}};
\node[tinycircle] (VTscalar) at (V|-eta) {\ssmall \textbf{-}};

\draw[->,thick] ([xshift=2mm]G.north) -- node[midway,right] {$x$}  (3.2,1.6)
--  (o1.east);
\draw[->,thick] (o1.west) -- node[midway,above] {$\widetilde{r}$} (VT.east);
\draw[->,thick] (r) -- (o1.south);
\draw[->,thick] (VT.west) -- node[midway,above] {$q$} (q) -- (-3.2,1.6)
-- ([xshift=-2mm]H.north);
\draw[->,thick] ([xshift=-2mm]H.south) -- node[midway,left] {$z$}  (-3.2,-1.6)
-- (o2.west);

\draw[->,thick] (o2.east) -- node[midway,below] {$\widetilde{q}$} (V.west);
\draw[->,thick] (q) -- (o2.north);
\draw[->,thick] (V.east) -- node[midway,below] {$r$}(r) -- (3.2,-1.6) -- ([xshift=2mm]G.south);

\draw[->,ultra thin,dashed] ([xshift=-2mm]G.north) -- node[midway,left] {$\eta$}  (eta)
--  (o1.south east);
\draw[->,ultra thin,dashed] ([xshift=2mm]H.south) -- node[midway,right] {$\nu$}  (nu)
-- (o2.north west);

\draw[->,ultra thin,dashed] (VTscalar.west) -- node[midway,below,xshift=5mm] {$\delta$} (delta) -- ([xshift=2mm]H.north);

\draw[->,ultra thin,dashed] (Vscalar.east) -- node[midway,above,xshift=-5mm] {$\gamma$} (gamma) -- ([xshift=-2mm]G.south);

\draw[->,ultra thin,dashed] (nu) -- (Vscalar.west);
\draw[->,ultra thin,dashed] (eta) -- (VTscalar.east);

\draw[->,ultra thin,dashed] (delta2) -- (Vscalar.north);
\draw[->,ultra thin,dashed] (gamma2) -- (VTscalar.south);

\draw[->,ultra thin,dashed]  ([xshift=5mm]o2.north east |- delta2) -- ([xshift=5mm,yshift=-2mm]o2.north east |- nu) -- (o2.north east);
\draw[->,ultra thin,dashed]  ([xshift=-5mm]o1.south west |- gamma2) -- ([xshift=-5mm,yshift=2mm]o1.south west |- eta) -- (o1.south west);
\end{tikzpicture}
\vspace{-1.5ex}
& 
\vspace{-1.5ex}
\begin{tikzpicture}
[
>=latex,
block/.style={draw,ultra thick,minimum width=1cm,minimum height=0.7cm},
normal/.style={draw,circle,ultra thick,minimum width=.5cm,minimum height=0.5cm},
circ/.style={draw,circle,minimum width=.5cm,inner sep=1pt,outer sep=0pt},
tinycircle/.style={
circle,draw,fill=none,
inner sep=0pt,outer sep=0pt,
minimum size=2mm,line width=0.2pt,
},
var/.style={coordinate}
]

\node[anchor=north east] at (3.7,2.2) {input side};
\node[anchor=south west] at (-3.7,-2.2) {output side};

\node[var] (q) at (-2,1.6) {};
\node[var] (qtil) at (-2,-1.6) {};
\node[var] (eta) at (2.8,1) {};
\node[var] (nu) at (-2.8,-1) {};
\node[var] (gamma) at (2.8,-1) {};
\node[var] (delta) at (-2.8,1) {};
\node[var] (gamma2) at (.5,-1) {};
\node[var] (delta2) at (-.5,1) {};

\node[block] (G) at (3,0) {$G$};
\node[tinycircle] (o1) at (2,1.6) {};
\node[tinycircle] (o2) at (-2,-1.6) {};
\node[var] (r) at (2,-1.6) {};
\node[var] (rtil) at (2,1.6) {};
\node[block] (H) at (-3,0) {$H$};
\node[normal] (VT) at (0,1.6) {$\mc N$};
\node[normal] (V) at (0,-1.6) {$\mc N$};
\node[tinycircle] (Vscalar) at (V|-nu) {\ssmall\textbf{-}};
\node[tinycircle] (VTscalar) at (V|-eta) {\ssmall\textbf{-}};

\draw[->,thick] ([xshift=2mm]G.north) -- node[midway,right] {$X$}  (3.2,1.6)
--  (o1.east);
\draw[->,thick] (o1.west) -- node[midway,above] {$\widetilde{R}$} (VT.east);
\draw[->,thick] (r) -- (o1.south);
\draw[->,thick] (VT.west) -- node[midway,above] {$Q$} (q) -- (-3.2,1.6)
-- ([xshift=-2mm]H.north);
\draw[->,thick] ([xshift=-2mm]H.south) -- node[midway,left] {$Z$}  (-3.2,-1.6)
-- (o2.west);

\draw[->,thick] (o2.east) -- node[midway,below] {$\widetilde{Q}$} (V.west);
\draw[->,thick] (q) -- (o2.north);
\draw[->,thick] (V.east) -- node[midway,below] {$R$}(r) -- (3.2,-1.6) -- ([xshift=2mm]G.south);

\draw[->,ultra thin,dashed] ([xshift=-2mm]G.north) -- node[midway,left] {$\bar\eta$}  (eta)
--  (o1.south east);
\draw[->,ultra thin,dashed] ([xshift=2mm]H.south) -- node[midway,right] {$\bar\nu$}  (nu)
-- (o2.north west);

\draw[->,ultra thin,dashed] (VTscalar.west) -- node[midway,below,xshift=5mm] {$\bar\delta$} (delta) -- ([xshift=2mm]H.north);

\draw[->,ultra thin,dashed] (Vscalar.east) -- node[midway,above,xshift=-5mm] {$\bar\gamma$} (gamma) -- ([xshift=-2mm]G.south);

\draw[->,ultra thin,dashed] (nu) -- (Vscalar.west);
\draw[->,ultra thin,dashed] (eta) -- (VTscalar.east);

\draw[->,ultra thin,dashed] (delta2) -- (Vscalar.north);
\draw[->,ultra thin,dashed] (gamma2) -- (VTscalar.south);

\draw[->,ultra thin,dashed]  ([xshift=5mm]o2.north east |- delta2) -- ([xshift=5mm,yshift=-2mm]o2.north east |- nu) -- (o2.north east);
\draw[->,ultra thin,dashed]  ([xshift=-5mm]o1.south west |- gamma2) -- ([xshift=-5mm,yshift=2mm]o1.south west |- eta) -- (o1.south west);
\end{tikzpicture} 
\vspace{-1.5ex}
\\
\hline
\vspace{-1.5ex}
\input{algo/vamp} 
& \vspace{-1.5ex}
\input{algo/se}\\
\hline
\end{tabular}
\caption{Schematic of quantities in VAMP and its State Evolution.  The objects denoted alongside solid lines belong to $E^n$ in VAMP, and $L^2(E)$ in State Evolution ($E$-valued random variables with finite variance). $\gamma,\delta,\eta,\nu\in \mc E_+$ are precision matrices. The $\mc N$ circles hide a few steps that add correlated Gaussian innovation after calculating the $2^{\rm nd}$ moments.
Both diagrams are symmetric across diagonals. 
One can initialize by providing either $(r_0,\gamma_0)$ as stated below, or alternatively, $(q_0,\delta_0)$. In practice the latter can be a more natural initialization for some problems.  
}\label{fig:vamp_se}
\end{figure}

\subsection{Fixed point of VAMP}\label{sec:fixed_point_VAMP}
Since  $F_2$ is   convex,  $h$ is a non-expansive operator in the $\|\cdot\|_\delta$ norm. Since $F_1$ may be  non-convex,    $g$ can be discontinuous,  and the usual subdifferential no longer applies to $F_1$. To state the stationarity conditions of \eqref{opt:xzU}, we instead use the more general Fr\'echet subdifferential of $F_1$: for $x \in \operatorname{dom} F_1$,~define
\begin{align*}
    \hat\partial F_1(x) := \set{u\in E^n:  \underset{h\downarrow 0}{\lim\inf}\inf_{\norm{v}=1}\round{\frac{F_1(x+hv)-F_1(x)}{h}-\inner{u,v}}\geq 0} .
\end{align*}
For $F_2$ we use the convex subdifferential $\partial F_2(z) := \{u \in E^n : F_2(z') \ge F_2(z) + \langle u, z'-z\rangle \text{ for all } z' \in E^n\}$. The two notions coincide when the function is convex. 

\begin{restatable}[Fixed point of VAMP]{proposition}{propfixedpoint}\label{proposition:fixed points}
Let  $(\hat x,\hat z,\hat r,\hat q, \hat\gamma, \hat \delta )\in E^n\times E^n\times E^n \times\mc E_+$ be a fixed point of \Cref{algo:vamp}. Then   $\hat z=U\tran\hat x,$ and 
\begin{align*}
(\hat r-\hat x)\hat\gamma\in \hat\partial F_1(\hat x) \qquad (\hat x-\hat r)\hat\gamma \in U\, \partial F_2(\hat z)\,.
\end{align*}
Consequently,  $\hat x$ is a stationary point of  \eqref{opt:xzU}:
\begin{align*}
\zero_E \in \hat\partial F_1(\hat x)+ U\, \partial F_2(\hat z).
\end{align*}
\end{restatable}
The proof is provided in \Cref{section:other proofs}. The RHS  is the Minkowski sum $\{a + b : a \in \hat\partial F_1(\hat x),\ b \in U\,\partial F_2( \hat z)\}$.  This proposition tells us that $(\hat x,\hat z)$ satisfies the  KKT conditions of  \eqref{opt:xzU}. If $F_1$ is also convex, both subdifferentials are convex subdifferentials, and $\hat x$ is a global minimiser of \eqref{opt:xzU}. 

\subsection{State Evolution (SE)}
We consider a sequence of problems indexed by $n$ and run VAMP  (\Cref{algo:vamp})   on each problem. 
Its State Evolution (\Cref{algo:se})  is a recursion over a few $E$-valued random variables and deterministic operators that describes the VAMP iterates as  $n\to\infty$.  The VAMP iterates are random because the data are random: the signal $x_0\in E^n$, which enters through the side information $w\in E_w^n$ and $s\in E_s^n$; the initialization $r_0\in E^n$; and the orthogonal matrix $U$, which we take to be Haar distributed, e.g., when $\Psi$ is rotationally invariant (see \Cref{theorem:state evolution} for the precise condition). Hence every quantity in \Cref{algo:vamp} is random, including $\gamma_t,\delta_t \in \mc{E}_+$ and the Onsager terms $\alpha_t+c_t$ and $\beta_t$, which are computed from averages over the $n$ rows.

Each array in VAMP, such as $x_t, r_t \in E^n$, $w\in E_{w}^{n}$, $s\in E_s^n$,  corresponds to  a random variable in the SE, written in capital letters, such as $X_t, R_t\in E$, $W\in E_w$ and $S\in E_s$, whose law is the limit of the empirical distribution of the rows of the VAMP arrays (\Cref{def:weak complete convergence}). %
 Each VAMP parameter, such as $\gamma_t$, $\delta_t$ or  $\alpha_t + c_t$, converges to a deterministic limit (\Cref{def:complete convergence}), written 
  with a bar ($\bar\gamma_t$, $\bar\delta_t$) or as $  A_t +  C_t$.   
The averages over rows in \Cref{algo:vamp} become expectations in   \Cref{algo:se}. 
VAMP uses denoisers $g_i$ and $h_i$, while  SE uses    the same maps $G$ and $H$ that define $g_i$ and $h_i$. 

Since  $U$ is Haar distributed,  each denoiser input behaves in the limit  like a scaled signal plus independent Gaussian noise, e.g., $R_{t-1} = X_0\Sigma_0^{-1}\rho_{t-1} + N_{t-1}$  with $N_{t-1}\independent (X_0,W) $ (lines~\ref{line:Nt}--\ref{line:Rt}). The SE tracks the scaling $\rho_{t-1}$ and the noise covariance  across iterations. 
This correspondence between VAMP and SE   relies on the Onsager terms: they must remove the part of each denoiser output that is correlated with the noise in its input. \Cref{sec:correction_onsager} shows how to construct them when the denoiser is discontinuous.

\subsection{\brickred{New:} Correction to the Onsager term}\label{sec:correction_onsager}
In \Cref{algo:vamp}, the Onsager term $\alpha_t+c_t$ determines $\eta_t$ (line \ref{line:eta_t}), and hence the  extrinsic estimate~$\wt r_t$   (line~\ref{line:tilde_rt}). It should be chosen so that $\wt r_t$ is uncorrelated with the Gaussian noise in the denoiser input $r_{t-1}$. This condition is the cleanest to state using  the SE in \Cref{algo:se}. Note that  $\wt r_t$~and~$r_{t-1}$ correspond to    $\wt R_t$  (line~\ref{line:tilde_Rt}) and     $R_{t-1} = X_0\Sigma_0^{-1}\rho_{t-1} + N_{t-1}$   (line~\ref{line:Rt}), and the Gaussian  noise in $r_{t-1}$ corresponds to  $N_{t-1}\sim \mc{N}(0_E, \tau_{t-1})$ independent of $(X_0, W)$.  Using these and    \Cref{remark:symmetry_prox} below, $\wt R_t$ is uncorrelated with $N_{t-1}$ when  the SE Onsager term  equals $\tau_{t-1}^{-1}\operatorname{Cov}(N_{t-1}, X_t)$, the normalized covariance between the noise and the denoiser output~$X_t$, where $X_t = G(R_{t-1}, \bar\gamma_{t-1}, W)$. 

For a Lipschitz denoiser, Stein’s lemma gives this covariance as the average Jacobian. For a discontinuous denoiser we need a  distributional version  of the Stein's lemma, which averages the full derivative,  including its jumps, against a Gaussian density.
To state the lemma, we introduce the following notation. 
For  $\zeta\in\SBVloc(E;E')$,  
and  a Gaussian with  mean $v_0\in E,$ and covariance $\tau\in\mc{E}_+$, we  define 
\begin{subequations}\label{def:onsager_gaussian}
\begin{align}
&\mc A(\zeta,v_0,\tau):= 
\int_E \varphi_\tau(v-v_0)\zeta'(v) \mc{L}(\dif v) \label{def:jacobian_contribution_gaussian},\\
&\mc C(\zeta,v_0,\tau) := 
\int_{J_{\zeta}} \varphi_{\tau}(v-v_0)~ \mathbf{n}_\zeta(v)\otimes [\zeta](v) \mc H(\dif v).\label{def:jump_contribution_gaussian}
\end{align}
\end{subequations}
Both lie in $L(E; E')$. By \Cref{def:SBVloc}, $\mc{A}(\zeta, v_0, \tau)+\mc{C}(\zeta, v_0, \tau) = \int_E \varphi_\tau(v-v_0) \mathrm{D}\zeta(\dif v)$.

\begin{restatable}[Stein's lemma for $\SBVloc$]{lemma}{steinlemma}\label{lemma:stein} Let $\zeta\in \SBVloc(E;E)$ 
be a proximal operator for a proper, lower semicontinuous function $E\mapsto \Real$ that  is bounded below. Let $R=X+N$ where $X$ is an $E$-valued random variable independent of $(M,N)$ which are jointly Gaussian with $N\sim\mc N(\zero_E,\tau)$ for some $\tau\in\mc E_+$. 
Further, suppose $\Exp\norm{\zeta(R)}^2<\infty$. Then, with $\mc{A}$ and $\mc C$ as in  \eqref{def:jacobian_contribution_gaussian}--\eqref{def:jump_contribution_gaussian}, 
\begin{align*}
\Cov(M,\zeta(R))=\Cov(M,N)\round{\Exp\zeta'(R) +\Exp \mc C(\zeta,X, \tau )} = \Cov(M,N) \Exp\left[ \mc{A}(\zeta, X, \tau) +   \mc{C}(\zeta, X, \tau)\right]\,.
\end{align*}
\end{restatable}
We prove this lemma in \Cref{section:proof stein}. 
\begin{remark}[Random denoisers]\label{remark:random_denoiser}
\Cref{lemma:stein}  also holds   when $\zeta$ is an $\SBVloc(E;E')$-valued random variable, provided  $(M,N)\independent (X,\zeta)$ and appropriate measurability of $\zeta.$ In this case, the jump sets may  be random, and the expectation in both terms is over $(X, \zeta)$. 
\end{remark}

\paragraph{Onsager term in SE:} 
Applying \Cref{lemma:stein} and \Cref{remark:random_denoiser} with $\zeta = G(\cdot, \bar\gamma_{t-1}, W)$ and with $X_0\Sigma_0^{-1}\rho_{t-1}$ in place of $X_0$   gives
$$\operatorname{Cov}(N_{t-1}, X_t) = \tau_{t-1}(A_t \brickred{\,+ \,C_t}),$$
where
\begin{subequations}\label{eq:SE_At_Ct_def}
\begin{align}
    A_t &:=  \Exp G'(R_{t-1},\bar \gamma_{t-1}, W)=\Exp \mc A\big(G(\cdot,\bar \gamma_{t-1}, W),  X_0\Sigma_0^{-1}\rho_{t-1},  \tau_{t-1}\big), \label{eq:SE_At_def}\\ 
    \brickred{C_t} \,&\,\brickred{:= \Exp\mc C\big(G(\cdot, \bar\gamma_{t-1}, W), X_0\Sigma_0^{-1}\rho_{t-1}, \tau_{t-1}\big)}.\label{eq:SE_Ct_def}
\end{align}
\end{subequations}
These are the quantities on lines~\ref{line:At}--\ref{line:Ct} of \Cref{algo:se}, so the SE uses exactly the required Onsager term. $A_t $ is the usual Onsager term (the average Jacobian)  in the AMP literature, and $C_t$ is the correction for the jumps.

\paragraph{Onsager term in VAMP:} VAMP must estimate   $A_t +  C_t$   from data. As in \Cref{algo:vamp},  the denoiser for the  $i^{\rm th}$ row is $g_i(\cdot, \gamma) = G(\cdot, \gamma, w_i)$, where $w_i$ is known, and the pairs $\big((r_{t-1})_i, w_i\big)$ behave like samples of $(R_{t-1}, W)$.

The jump term  $C_t$ integrates the conditional density of $R_{t-1}$ given $W$, denoted by $f_{R_{t-1}\mid W}$, over the jump set of $G(\cdot, \bar\gamma_{t-1}, W)$, so it cannot be estimated by an  analogous average of a function of $\big((r_{t-1})_i, w_i\big)$ over the rows. The density $f_{R_{t-1}\mid W}$ is also unknown since  it depends on $\tau_{t-1}$ and on the joint  law of $(X_0, W)$. Thus, $C_t$ cannot be   computed directly. 
We therefore     replace the unknown density of $R_{t-1}$ by a Gaussian kernel density estimate (KDE) with bandwidth $b > 0$. 
  For $t \ge 1$ and $b>0$, define
\begin{subequations}\label{eq:kde_onsager}
\begin{align}
    \alpha_{t}(b) &:= \frac{1}{n}\sum_{i=1}^n \mc A(g_i(\cdot,\gamma_{t-1}),(r_{t-1})_i, b^2\gamma_{t-1}\inv) \label{eq:alpha_t(b)},\\ 
    \brickred{c_{t}(b) } &\brickred{:= \frac{1}{n}\sum_{i=1}^n \mc C  \big(g_{i}(\cdot, \gamma_{t -1}), (r_{t-1 })_i, b^2\gamma_{t -1}\inv\big)}.\label{eq:c_t(b)}
\end{align}
\end{subequations}
These are obtained from \eqref{eq:SE_At_def}--\eqref{eq:SE_Ct_def} by replacing the Gaussian with mean $X_0\Sigma_0^{-1}\rho_{t-1}$ and covariance $\tau_{t-1}$ by a kernel centred at each row $(r_{t-1})_i$ with covariance $b^2\gamma_{t-1}^{-1}$, and the expectation over $(X_0, W)$ by an average over the rows. 
Our Onsager term at iteration $t$ is then 
\begin{align}
\alpha_{t}  \brickred{\,+\,c_{t}}\,,\quad  \text{where}\quad 
\alpha_t:=\alpha_t(b_t) \quad \text{and}\quad \brickred{c_t:=c_t(b_t)} \label{eq:def_alpha_t_c_t}
\end{align}
for a bandwidth $b_t>0$. 
Here, $\alpha_{t}$ is a Gaussian-smoothed version of the  usual Onsager term, and $c_{t}$  is the correction for the jumps.

\paragraph{Consistency:} \Cref{lemma:consistency of onsager terms} below shows that if   $b_t=b_t^{(n)}$ decreases to zero slowly enough as $n\to \infty$,  then  $ \alpha_{t} +c_{t} \cto A_{t} + C_{t}$. 
This consistency result  is the key new step in proving our main result \Cref{theorem:state evolution} in \Cref{sec:main_result}, which establishes  the validity of the  State Evolution.

\begin{remark}[Smoothing  the Jacobian term]\label{remark:smooth_onsager} The Gaussian smoothing  of the Jacobian term in \eqref{eq:alpha_t(b)} is not always necessary. Suppose that, in certain settings,   $G'$ satisfies the same almost everywhere joint continuity as $H'$ in \Cref{assumption:regularity},  and that 
$G'$ is  bounded uniformly in \((v,w)\) and locally uniformly in~\(\gamma\). 
Then under the hypotheses of \Cref{lemma:consistency of onsager terms}, both the plain average $\frac1n \sum_{i=1}^n g_i'\big((r_{t-1})_i, \gamma_{t-1}\big)$ and its  smoothed version $\alpha_t(b_t)$  converge completely to $A_t$, for any $b_t\to 0$. 
  For the bandwidth $b_t$ given by \Cref{lemma:consistency of onsager terms}, it follows that  $c_t(b_t) \cto C_t$. Hence VAMP may use the plain average in place of $\alpha_t$, and its Onsager term  remains consistent, i.e., $\frac1n\sum_{i=1}^n g_i'\big((r_{t-1})_i, \gamma_{t-1}\big) + c_t(b_t)\cto A_t+C_t$.

If $G'$ is unbounded, as can happen for proximal operators of non-convex functions, convergence of the empirical distribution of the rows  no longer  implies convergence of the plain average. The Gaussian smoothing in  $\alpha_t$ avoids this problem, giving  $\alpha_t + c_t\cto A_t+C_t$. 
\end{remark}

\begin{remark}[Symmetry of the Onsager terms]
 \label{remark:symmetry_prox} By \Cref{proposition:asplund}, for every $\gamma\in \mc{E}_+$, $g_i(\cdot, \gamma)\gamma$ is almost everywhere the gradient of a convex function, so $(\mathrm{D}g_i)\gamma$ is a symmetric positive semidefinite measure. By \eqref{def:onsager_gaussian}, $(\alpha_t + c_t)\gamma_{t-1}$ and $(A_t + C_t)\bar\gamma_{t-1}$ are Gaussian-weighted integrals of such measures, and are therefore also symmetric positive semidefinite. 
Hence $\eta_t = \gamma_{t-1}[(\alpha_t + c_t)^*]^{-1} = (\alpha_t + c_t)^{-1}\gamma_{t-1}$ is symmetric, and so is  $\bar\eta_t$. Likewise, for every $\delta\in \mc{E}_+$, $(\mathrm{D}h_i)\delta $ is a symmetric positive definite measure, so  $\nu_t = \delta_t (\beta_t^*)^{-1} = \beta_t^{-1}\delta_t$ and $\bar\nu_t$ are both symmetric.
\end{remark}

\section{Applications of  VAMP to Analyze Non-Convex M-Estimators}
\label{section:examples}

In this section we  consider two non-convex M-estimation problems in the form of \eqref{opt:theta_beta_Phi}: group sparse regression, where the non-convexity comes from an $\ell_0$-type penalty  (\Cref{sec:group_sparse_example}), and robust regression with the Cauchy loss, where the non-convexity comes from the loss (\Cref{sec:cauchy_example}). We show how each  
can be written in the form of \eqref{opt:xzU} so that VAMP (\Cref{algo:vamp}) can be applied, and how its State Evolution (\Cref{algo:se}) then characterizes the VAMP fixed points, which are stationary points of \eqref{opt:xzU} by \Cref{proposition:fixed points}. 

We use 
the full SVD of the design matrix $\Psi\in \Real^{m\times p}$:   $\Psi=V_{\sf L}\operatorname{diag}(\sigma)V_{\sf R}^\top$, where $\diag(\sigma)\in \Real^{m\times p}$ has the singular values of  $ \Psi$ as the first $\min(m,p)$ entries on its diagonal and $\sigma_i:=0$ for $i>\min(m,p)$. 
 
\subsection{Penalized least squares with a non-convex regularization}\label{sec:group_sparse_example}
Consider $y\in \Real^{m\times d}$ of the form $y=\Psi\theta_0+\xi$, where  $\theta_0\in \Real^{p\times d}$ is the unknown signal, and  the noise $\xi\in \Real^{m\times d}$ has  i.i.d.\ entries from $\mc N(0,\eps)$. The penalized least-squares estimator $\widehat\theta$ based on $y$ and $ \Psi$ is  defined as
\begin{align}\label{eq:least_squares_objective}
\widehat\theta \in \underset{\theta \in \Real^{p\times d}}{\arg\min}\,\tfrac12\norm{y-\Psi \theta}^2 + \sum_{i=1}^p\pen_i(\theta_i)
\end{align}
where $ \sum_{i=1}^p\pen_i(\theta_i)$ is some regularization term. 
To apply VAMP from \Cref{algo:vamp}, we set $E=\Real^{d}$,  $(n,x,U)\gets(p,\theta,V_{\sf R})$. 
So the objective in \eqref{eq:least_squares_objective} becomes
\begin{align*}
    \min_{x, z\in \Real^{p\times d}} F_1(x)+F_2(z) \quad \text{subject to}\quad  z=U\tran x ,
\end{align*}
where  $F_1(x)= \sum_{i=1}^p\pen_i(\theta_i)$ and $F_2(z) = \tfrac{1}{2}\| V_{\sf L}^\top y-\diag(\sigma) z\|^2.$ 
Note here  $F_2$ is convex and $F_1$ can be non-convex, as required by \Cref{assumption:structure}. Write $\wt y:=V_{\sf L}\tran y \in \Real^{m\times d}$;   if $m<p$ we set $\tilde y_i := 0_d$ for $m<i\le p$, and if $m>p$ the rows $\tilde y_{p+1},\dots,\tilde y_m$ only contribute a constant to $F_2$ and are dropped. Hence, the proximal operator of $F_2$ is row-wise separable, where for $i\in [p]$, 
\begin{align}\label{eq:h_sparse_group}
h_i(q_i,\delta)\!=\!\prox_{F_{2i}} (q_i, \delta) \in\underset{z_i\in\Real^{ d}}{\arg\min} \, \tfrac12\|z_i-q_i\|_\delta^2+\tfrac12\|\wt y_i-\sigma_i z_i\|^2 \!= \!\left(\sigma_i\wt y_i+q_i\delta\right)\round{\sigma_i^2 I_{d}+\delta}\inv\!.
\end{align}
Note for $\sigma_i=0$, $h_i(q_i,\delta)=q_i$. 
Here the side information  $s_i$ in  $h_i(\cdot, \delta)=H(\cdot, \delta, s_i)$ is  $s_i=(\sigma_i,\wt y_i)$. Next, we focus on a penalty function $F_1$ often used in  group sparse regression.

\subsubsection{Group sparse regression}\label{sec:group_sparse_regression_g}
A  problem of particular interest is group sparse (row-sparse) regression with the non-convex  penalty $$\pen_j(x_j)=
\lambda \mathbbm{1}\!\set{x_j\neq 0_{d}}$$ which counts the nonzero rows of $x_j$; the  $d=1$ case recovers standard sparse regression.
The proximal operator of $F_1(x)$, also row-wise separable, is then the hard-thresholding operator:
\begin{align*}
    g_i(r_i, \gamma)=\prox_{F_{1i}}(r_i, \gamma) \in  \underset{x_i\in\Real^{d}}{\arg\min} \,\tfrac12\norm{x_i -r_i}_\gamma^2 + \lambda \mathbbm{1}\{x_i\ne 0_{d}\}=r_i\mathbbm{1}\set{\norm{r_i}_\gamma>\sqrt{2\lambda}} .
\end{align*}
This expression does not depend on $i$, so  the side information $w_i$ in $ g_i(\cdot, \gamma)= G(\cdot, \gamma, w_i)$ is trivial and we can set $w_i=\emptyset$. 

 \subsubsection{Onsager term in VAMP}

Given $g(v,\gamma)=v\mathbbm{1}\big\{\norm{v}_\gamma>\sqrt{2\lambda}\big\}$,  we have 
 $g'(v,\gamma)=I_d 
\indicator\big\{\norm{v}_\gamma >\sqrt{2\lambda}\big\}$. Following the notation from \Cref{def:jump_set},   the jump set  $J_{g(\cdot, \gamma)}$, 
  the   jump normal $\mathbf{n}_{g(\cdot, \gamma)}$    and the jump $[g(\cdot, \gamma)]$ are
\begin{align*}
  J_{g(\cdot, \gamma)}:=\set{v\in\Real^{ d}: \norm{v}_\gamma=\sqrt{2\lambda}},\quad \mathbf{n}_{g(\cdot, \gamma)}(v)= \frac{v\gamma}{\norm{v\gamma}} \;\; \text{and}\;\; [g(\cdot, \gamma)](v)=v,\;\;\text{for all }v\in J_{g(\cdot, \gamma)}.
\end{align*}
Note the jump set  $J_{g(\cdot, \gamma)}$ is an ellipsoid.  Hence by \eqref{def:jacobian_contribution_gaussian},   the Jacobian term equals \begin{align}
    \mc A\big(g(\cdot,\gamma),v_0,\tau\big)=\left[\int_{\|v\|_\gamma > \sqrt{2\lambda}}\varphi_\tau(v-v_0)  \dif v\right] I_d = \mathbb{P}&\left(\|v_0+N\|_\gamma>\sqrt{2\lambda}\right)I_d \nonumber\\
    &\quad \text{with}\quad   N\sim \mc{N}(0_d, \tau), \label{eq:hard_threshold_mc_A}
\end{align}
which is the probability that the input exceeds the threshold $\sqrt{2\lambda}$ in the $\norm{\cdot}_\gamma$ norm.
For the jump term for $d=2$, parametrizing  the ellipse $J_{g(\cdot,\gamma)}$ by $v(\theta)=\sqrt{2\lambda} u(\theta)\gamma\isqrtm$ for $\theta\in[0,2\pi)$, where $u(\theta)=(\cos\theta,\sin\theta)$ gives   $\mc H(\dif v)/\|v\gamma\|=\dif \theta/\sqrt{\det\gamma}$,  so the jump term defined in \eqref{def:jump_contribution_gaussian} becomes  \begin{align}
\mc C(g(\cdot,\gamma),v_0,\tau)
&=\int_{J_{g(\cdot, \gamma)}}\!\!
\!\varphi_\tau(v-v_0)
\frac{v\gamma}{\norm{v\gamma}}\otimes v\,
\mc H(\dif v) \nonumber\\
&=\tfrac{2\lambda}{\sqrt{\det\gamma}}\,
\gamma\sqrtm\!
\sbrac{\int_0^{2\pi}
\!\varphi_\tau\left(\sqrt{2\lambda} u(\theta)\gamma^{\isqrtm}-v_0\right)\,
u(\theta)\otimes u(\theta)\,\dif\theta}
\gamma\isqrtm. \label{eq:mathcal_C_hard_threshold_general}
\end{align}
where we have used $d=2$ in the last line.
The Gaussian kernel used in  \eqref{eq:alpha_t(b)}--\eqref{eq:c_t(b)} has covariance $\tau=b^2\gamma^{-1}$, which has the same shape as the ellipse. The Jacobian term becomes 
\begin{align}\label{eq:mathcal_A_hard_threshold}
    \mc A\big(g(\cdot,\gamma),v_0,b^2\gamma^{-1}\big)=\mathbb{P}\left(\big\|v_0 \gamma^{\sqrtm} + b\Xi_0\big\|> \sqrt{2\lambda}\right)I_d \;\; \text{with}\;\;  \Xi_0\sim \mc{N}(0, I_d). 
\end{align}
Let
$\kappa_0:=\sqrt{2\lambda \|v_0\|_\gamma^2}/b^2$. For $k\in \mathbb{Z}$, using $\int_0^{2\pi}e^{\kappa_0\cos\theta}\cos (k\theta)\dif\theta=2\pi \mathrm{B}_k(\kappa_0)$, where $\mathrm{B}_k$ is the modified Bessel function of the first kind \cite[Eq.~9.6.19]{Abramowitz1964handbook}, the jump term   has the closed form
\begin{align}
    \label{eq:mathcal_C_hard_threshold}
\mc C\big(g(\cdot,\gamma),v_0,b^2\gamma^{-1}\big)=\frac{\lambda}{b^2}\exp\Big(-\tfrac{2\lambda+\|v_0\|_\gamma^2}{2b^2}\Big)\left[\big(\mathrm{B}_0(\kappa_0)-\mathrm{B}_2(\kappa_0)\big)I_d+2\mathrm{B}_2(\kappa_0)\,\frac{(v_0\gamma)\otimes v_0}{\|v_0\|_\gamma^2}\right].
\end{align}
When $v_0=0_d$, 
$\|v_0\|_\gamma^2=0$ since $\gamma\in\mc \Real_+^{d\times d}$,
so the ratio $(v_0\gamma)\otimes v_0/\|v_0\|_\gamma^2$ is undefined, and we set the   term $2\mathrm{B}_2(\kappa_0) (v_0\gamma)\otimes v_0/{\|v_0\|_\gamma^2}$ to zero. This is its limit as $v_0\to0_d$, because  $(v_0\gamma)\otimes v_0/{\|v_0\|_\gamma^2}$ is bounded and $\mathrm{B}_2(\kappa_0)\to0$. Together with $\mathrm{B}_0(0)=1$, the jump term then reduces to $\mc C\big(g(\cdot,\gamma),0_d,b^2\gamma^{-1}\big)=(\lambda/{b^2})e^{-\lambda/b^2}I_d$.
 Moreover, note that for general covariance $\tau$,
$\mc{C}$ has no closed form.    Nevertheless, it remains a one-dimensional integral over $\theta$ that is cheap to evaluate numerically.

Finally by  \eqref{eq:alpha_t(b)}--\eqref{eq:c_t(b)} and \eqref{eq:def_alpha_t_c_t}, $\alpha_t$ and $c_t$ are the averages of \eqref{eq:mathcal_A_hard_threshold} and \eqref{eq:mathcal_C_hard_threshold} over the rows, with $v_0=(r_{t-1})_i$,  $b=b_t$ and $\gamma=\gamma_{t-1}$. Since $g'$ is bounded, \Cref{remark:smooth_onsager} also allows $\alpha_t$ to be replaced by the plain average $\frac1n \sum_{i=1}^n g_i'\big((r_{t-1})_i, \gamma_{t-1}\big)$.

\subsubsection{State Evolution  for an overparameterized orthogonal design}\label{sec:SE_simplified_orthogonal_hard_threshold}
We now make the State Evolution in \Cref{algo:se} explicit   for a specific overparameterized orthogonal design. Let $a:=p/m>1$ be the aspect ratio and consider the design $\Psi$ that satisfies $\Psi\Psi^\top=I_m$.  This means the singular values   $\sigma_i\in \{0,1\}$ where   a  $1/a$  fraction  of them equal 1 and the rest are 0. 
\begin{remark}[Underparameterized regime]
   The underparameterized case where $a=p/m\le 1$, i.e., when   $\Psi\tran\Psi=I_p$, is much simpler.
Since $\|y-\Psi\theta\|^2=\|\Psi^\top y-\theta\|^2+\text{constant}$, problem \eqref{eq:least_squares_objective} separates over rows and has the closed-form solution
$$\widehat\theta_i=(\Psi^\top y)_i\,\mathbbm{1}\big\{\|(\Psi^\top y)_i\|>\sqrt{2\lambda}\big\},$$
i.e. hard thresholding of $\Psi^\top y$ at level $\sqrt{2\lambda}$. 
The strong law of large numbers therefore gives the asymptotic MSE directly:
$$\frac1p\|\widehat\theta-\theta_0\|^2\to\Exp\Big\|(X_0+\Xi) \mathbbm{1}\big\{\|X_0+\Xi\|>\sqrt{2\lambda}\big\}-X_0\Big\|^2\quad\text{with}\quad  \Xi\sim\mc N(0,\varepsilon I_d),$$
rendering the  VAMP analysis unnecessary. 
We thus focus on the overparameterized case $a>1$ in the rest of this section.
\end{remark}

The side information of $G$ is trivial, and  that of $H$ is $S=(\Sigma, \Sigma Z_0+\Xi)$,  the limit of $(\sigma_i,\tilde y_i)$, where $\Sigma\in\{0,1\}$ with $\mathbb P(\Sigma=1)=1/a$,  and $\Sigma$, $Z_0$, $\Xi$ are independent.
 Let $X_0$ have second moment $\Sigma_0$. The State Evolution (\Cref{algo:se}) reduces to a recursion for $(\bar\gamma_t,\tau_{tt})$. We start on the output side with $q_0=0$ and any $\bar\delta_0\in\mc E_+$. Then $Q_0-Z_0=-Z_0$, and the output-side computation below, applied with $Q_0=0$,  gives the inputs required by the SE:  
 \begin{align}\label{eq:initialise_hard_threshold}
  \rho_0=\Sigma_0,\quad    \tau_{00}=(a-1)\Sigma_0+a\eps I_d,\quad \bar\gamma_0^{-1}=aI_d+(a-1)\bar\delta_0^{-1}\succ aI_d
 \end{align}so that $0\prec\bar\gamma_0\prec \frac{1}{a} I_d$.

\textbf{Input side:} At iteration $t$, by lines~\ref{line:Nt}--\ref{line:Rt}  with $\rho_{t-1}=\Sigma_0$,  the input to $g$ is $R_{t-1}=X_0+N_{t-1}$, with $N_{t-1}\sim\mc N(0,\tau_{t-1, t-1})$ independent of $X_0$. The denoiser output is  $X_t=G(R_{t-1},\bar\gamma_{t-1})$. 
Two quantities summarise how well $G$ performs on this input. The first is its mean squared error matrix $
\Exp\big[(X_t-X_0)^{\otimes2}\big],$
 and the second is the Onsager term $A_t+C_t$ of line~\ref{line:At+Ct}. With $\mc{A}$ and $\mc C$ defined as in \eqref{eq:hard_threshold_mc_A} and \eqref{eq:mathcal_C_hard_threshold_general}, and $ g_i(\cdot, \gamma)= G(\cdot, \gamma, \emptyset)$ as shown in \Cref{sec:group_sparse_regression_g}, we have 
 \begin{align*}
A_t  =\Exp \mc A\big(G(\cdot,\bar\gamma_{t-1}, \emptyset),X_0,\tau_{t-1,t-1}\big) ,
\quad 
C_t  = \Exp  \mc{C}\big(G(\cdot,\bar\gamma_{t-1},\emptyset), X_0, \tau_{t-1,t-1}\big) . %
 \end{align*}
Note both terms are evaluated at  covariances that are in general not proportional to $\bar\gamma_{t-1}^{-1}$, so the simplified expressions in \eqref{eq:mathcal_A_hard_threshold}--\eqref{eq:mathcal_C_hard_threshold} do not apply. 
 By \Cref{lemma:stein}, it is the normalized correlation between the input noise and the output,
$$A_t+C_t=\tau_{t-1, t-1}^{-1}\operatorname{Cov}(N_{t-1},X_t)=\tau_{t-1, t-1}^{-1} \Exp\big[N_{t-1}\otimes X_t\big],$$where the second equality holds because $N_{t-1}$ is centered.
Lines~\ref{line:At+Ct}--\ref{line:delta_bar} then give  $\bar\eta_t$ and $\bar\delta_t$,  and line~\ref{line:tilde_Rt} forms the extrinsic estimate $\widetilde R_t$ with   error $$\widetilde R_t-X_0=\big[(X_t-X_0)\,\bar\eta_t-(R_{t-1}-X_0)\,\bar\gamma_{t-1}\big]\,\bar\delta_t^{-1}=\big[(X_t-X_0)\,\bar\eta_t-N_{t-1}\bar\gamma_{t-1}\big]\,\bar\delta_t^{-1}.$$
When we take its second moment, the cross terms $\Exp[((X_t-X_0)\bar\eta_t) \otimes  (N_{t-1} \bar\gamma_{t-1})]$ and  $\Exp[(N_{t-1} \bar\gamma_{t-1})\otimes ((X_t-X_0)\bar\eta_t)]$ involve $\Exp N_{t-1}\otimes X_t =\tau_{t-1, t-1}(A_t+C_t)$, and by line~\ref{line:At+Ct}, reduce to   $\bar\gamma_{t-1}\tau_{t-1, t-1}\bar\gamma_{t-1}$, giving  
$$\Exp (\widetilde R_t-X_0)^{\otimes2} =\bar\delta_t^{-1}\big(\bar\eta_t\Exp (X_t-X_0)^{\otimes 2}\bar\eta_t-\bar\gamma_{t-1}\tau_{t-1, t-1}\bar\gamma_{t-1}\big)\bar\delta_t^{-1}.$$

\textbf{Output side:}
The output step sees the extrinsic estimate after rotation. Lines \ref{line:pi_t}--\ref{line:Q_t} build $Q_t=Z_0\Sigma_0^{-1}\pi_t+M_t$ where $M_t\sim \mc{N}(0, \mu_{tt})$ with the same error second moment as $\widetilde R_t$:
$$\Exp (Q_t-Z_0)^{\otimes2} =\mu_{tt}+(\Sigma_0^{-1}\pi_t-I_d)^*\Sigma_0(\Sigma_0^{-1}\pi_t-I_d)=\Exp (\widetilde R_t-X_0)^{\otimes2} .$$
 Note $S$, $Z_0$, $\Xi$, $M_t$ are independent. Using $H(\cdot, \delta, s_i)=h_i(\cdot, \delta)$ from \eqref{eq:h_sparse_group} combined with line~\ref{line:Z_t} gives
 \begin{align}\label{eq:Zt_Z0}
     Z_t-Z_0=\begin{cases}Q_t-Z_0, & \Sigma=0,\\ \big(\Xi+(Q_t-Z_0)\bar\delta_t\big)(I+\bar\delta_t)^{-1}, & \Sigma=1.\end{cases}
 \end{align}
The Jacobian of $H$ is $I_d$ when $\Sigma=0$ and $\bar\delta_t(I_d+\bar\delta_t)^{-1}$ when $\Sigma\!=\!1$, so line~\ref{line:B_t} gives $ B_t=I_d-\tfrac1a(I_d+\bar\delta_t)^{-1}$. This is symmetric and also invertible 
for $a\ge1$.  Lines~\ref{line:bar_nu_t}--\ref{line:bar_gamma_t} then simplify to 
\begin{align}
\bar\gamma_t=\bar\delta_t  B_t^{-1}(I_d-  B_t)=\tfrac1a \bar\delta_t   B_t^{-1}(I_d+\bar\delta_t)^{-1}=  \tfrac1a \bar\delta_t\big[(I_d+\bar\delta_t)-\tfrac1aI_d\big]^{-1}=\bar\delta_t\big[(a-1)I_d+a\bar\delta_t\big]^{-1},
\end{align}
that is, 
$$\bar\gamma_t^{-1}=  (a-1)\,\bar\delta_t^{-1}+ aI_d.$$
Line~\ref{line:tilde_Qt} gives $\widetilde Q_t-Z_0=\big[(Z_t-Z_0)\bar\nu_t-(Q_t-Z_0)\bar\delta_t\big]\bar\gamma_t^{-1}$.
Using $\bar\delta_t\bar\gamma_t^{-1}=a(I_d+\bar\delta_t)-I_d$ and $\bar\nu_t\bar\gamma_t^{-1}=a(I_d+\bar\delta_t)$, both of which follow from the update above, and applying \eqref{eq:Zt_Z0} gives
$$\widetilde Q_t-Z_0=\begin{cases}Q_t-Z_0, & \Sigma=0,\\ a\,\Xi-(a-1)(Q_t-Z_0), &  \Sigma=1.\end{cases}$$
Averaging over $\Sigma$, the terms in $Q_t-Z_0$ cancel in $\Exp[Z_0\otimes(\widetilde Q_t-Z_0)]$, so $\rho_t=\Sigma_0$ (line~\ref{line:rho_t}). Line~\ref{line:Rt}  then gives $R_t=X_0+N_t$, which is the form assumed on the input side. Line~\ref{line:tau_tt}  gives
$$\tau_{tt}=\Exp (\widetilde Q_t-Z_0)^{\otimes2} =a\varepsilon I_d+(a-1)\Exp  (Q_t-Z_0)^{\otimes2} .$$
For a general $X_0$, $A_t+C_t$ and $\Exp (X_t-X_0)^{\otimes2}$ can be evaluated by Monte Carlo. The MSE of  the VAMP output $x_t$  for each fixed $t$ is  
\begin{align}\label{eq:MSE_Bern_Gaussian}
  \frac{1}{p} \|x_t - x_0\|^2\to  \trace \big[\Exp  (X_t - X_0)^{\otimes 2} \big] = \Exp  \Big\| R_{t-1}\mathbbm{1}\Big\{ \|R_{t-1}\|_{\overline{\gamma}_{t-1}} > \sqrt{2\lambda}\Big\}  - X_0\Big\|^{ 2} ,
\end{align}
the MSE of the final estimate as $t\to \infty$, which constitutes $\widehat{\theta}$, is  the RHS in the limit of $t\to \infty$. If the state-evolution MSE converges as \(t\to\infty\), its limit gives the asymptotic MSE of the VAMP iterates, with \(p\to\infty\) taken first.

\paragraph{SE specialized to Bernoulli-Gaussian signal:}   \Cref{section:expts_hard_threshold} assumes that 
 the rows of $\theta_0$ are i.i.d.\ with a Bernoulli-Gaussian distribution:
$$X_0\sim(1-\sparse)\,\delta_{0_{ d}}+\sparse\, \mc N\left(0_d, \sparse^{-1} I_d\right),$$
where $\sparse\in(0,1]$ is the probability that each row  is nonzero,  and the scaling ensures $\Sigma_0=\Exp[X_0^{\otimes 2}] = I_d$.
Equivalently this means $X_0= B_{X_0} G_{X_0}$  where $B_{X_0}\sim \text{Bern}(\sparse)$ and $G_{X_0}\sim \mc{N}\left(0_d, \sparse^{-1} I_d\right)$. In this setting, the SE derived above can be further simplified as follows.

Initializing with $\rho_0=\Sigma_0$  as in \eqref{eq:initialise_hard_threshold} gives  
$R_{t-1}=X_0+N_{t-1}$, where $N_{t-1}\sim\mc N(0,\tau_{t-1, t-1}) \indep X_0$. 
Conditioned on $B_{X_0}$, $R_{t-1}$ is a centered Gaussian with covariance  
\begin{align*}
   \tau_{R,0}:= \tau_{t-1, t-1} \quad \text{when}\; B_{X_0}=0\quad \text{or}\quad \tau_{R,1}:=\tau_{t-1,t-1} + \sparse^{-1}I_d\quad \text{when}\; B_{X_0}=1.
\end{align*}
The tower property gives
\begin{align*}
    A_t + C_t &= (1-\sparse )\Exp \big[(\mc{A}+\mc{C})(G(\cdot, \bar\gamma_{t-1},\emptyset), X_0, \tau_{R,0}) \mid B_{X_0}=0\big] \\
    &\qquad + \sparse \Exp\big[(\mc{A}+\mc{C})(G(\cdot, \bar\gamma_{t-1}, \emptyset), X_0, \tau_{R,0}) \mid B_{X_0}=1\big]\\
    & = (1-\sparse )(\mc{A}+\mc C)\big(G(\cdot, \bar\gamma_{t-1}, \emptyset), 0_d, \tau_{R,0}\big)   +\sparse (\mc{A}+\mc C)\big(G(\cdot, \bar\gamma_{t-1}, \emptyset), 0_d, \tau_{R,1}\big)\end{align*} 
    where the second equality uses that, by \eqref{eq:hard_threshold_mc_A}--\eqref{eq:mathcal_C_hard_threshold_general}, $\mc{A}$ and $\mc{C}$ depend on $X_0$ only through the function  $v\mapsto \varphi_{\tau}(v - X_0)$ on  $\Real^d$ and $\Exp\big[\varphi_{\tau}(v- X_0) \mid B_{X_0}=1\big] = \varphi_{\tau +  \sparse^{-1} I_d}(v)$ for all $v\in \Real^d$. By~\eqref{eq:hard_threshold_mc_A}, the Jacobian term at $v_0=0_d$ takes the form $$\mc{A}\big(G(\cdot,  \gamma, \emptyset), 0_d, \tau \big) =\mathbb{P} \left(\|N\|_{\gamma} > \sqrt{2\lambda}\right) I_d \quad \text{with}
    \quad N\sim \mc N(0_d, \tau).$$
    For $d = 2$, the jump term  at $v_0 = 0_d$ also has a closed form.  Let $0 < q_1 \le q_2$ be the eigenvalues of $\gamma^{\isqrtm}\tau^{-1}\gamma^{\isqrtm}$, and let $e \in \mathbb{R}^d$ be a unit eigenvector for $q_1$. Measuring $\theta$ from $e$, the Gaussian exponent in \eqref{eq:mathcal_C_hard_threshold_general} becomes
    \begin{align*}
        -\tfrac12  v(\theta) \tau^{-1}v(\theta)^\top = -\lambda  u(\theta) \gamma^{\isqrtm}\tau^{-1}\gamma^{\isqrtm} u(\theta)^\top=-\lambda(q_1\cos^2\theta + q_2\sin^2\theta) &= -\tfrac{\lambda}{2}(q_1+q_2) + \kappa\cos (2\theta ),\\
        &\text{where}\quad \kappa=\tfrac{\lambda}{2}(q_2-q_1)\ge0\,.
    \end{align*}
For $k\in \mathbb{Z}$, using the Bessel identity  $\int_0^{2\pi}e^{\kappa \cos\theta}\cos (k\theta)\dif\theta=2\pi \mathrm{B}_k(\kappa )$ like above, we obtain
\begin{align*}
   \mc C(G(\cdot,  \gamma, \emptyset),0_d,\tau) = 
    \frac{\lambda e^{-\lambda(q_1+q_2)/2}}
         {\sqrt{\det\gamma\,\det\tau}}\,
    \gamma^{\sqrtm}
    \sbrac{
        \round{\mathrm{B}_0(\kappa)-\mathrm{B}_1(\kappa)}I_d
        +2\mathrm{B}_1(\kappa)\,e\otimes e
    }
    \gamma^{\isqrtm}.
\end{align*}
When $q_1=q_2$, we may choose $e$ arbitrarily among unit vectors,
because $\mathrm{B}_1(0)=0$. 

Finally, write   $\chi(u)
    = (1+u)e^{-u}$ and define for $\tau\in \Real_\succ^{d\times d}$,   $$k_\lambda(\tau,\gamma)
    =
    \Exp_{R\sim \mc N(0_d,\tau)}
    \left[
        R\otimes R
        \mathbbm{1}\big\{\norm{R}_\gamma>\sqrt{2\lambda}\big\} 
    \right]= 
    \frac1\pi
    \int_0^{2\pi}
        \big(u(\theta)\tau^{\sqrtm}\big)^{\otimes 2}
        \chi\!\left(
            \tfrac{\lambda}
            { \big\|u(\theta)\tau^{\sqrtm}\big\|^2_\gamma}
        \right)
    \dif \theta.$$
For $d=2$, we then have 
$$\Exp (X_t - X_0)^{\otimes 2}  = I_d + (1-\sparse) k_\lambda(\tau_{R,0},\bar\gamma_{t-1}) + \sparse k_\lambda(\tau_{R,1},\bar\gamma_{t-1}) - \tau_{R,1}^{-1}k_\lambda(\tau_{R,1}, \bar\gamma_{t-1}) - k_\lambda(\tau_{R,1}, \bar\gamma_{t-1}) \tau_{R,1}^{-1},$$
and by \eqref{eq:MSE_Bern_Gaussian}, the asymptotic MSE is $\trace[\Exp  (X_t - X_0)^{\otimes 2} ] $ as $t\to \infty$.

 \paragraph{Isotropic initialization.} In this Bernoulli-Gaussian setting, 
if we further initialize $\bar\gamma_0 \propto I_d$, i.e., $\bar\delta_0\propto I_d$ in \eqref{eq:initialise_hard_threshold}, then $\bar\gamma_t$, $\bar\delta_t$, $\tau_{tt}$, $A_t+C_t$ and $\Exp(X_t-X_0)^{\otimes 2}$ are multiples of $I_d$ for all $t\ge 0$. Indeed, for $\bar\gamma_{t-1}\propto I_d$ the denoiser $G(\cdot,\bar\gamma_{t-1},\emptyset)$ commutes with rotations, and for $\tau_{t-1,t-1}\propto I_d$ both $X_0$ and $N_{t-1}$ are rotation invariant. Hence $A_t+C_t$ and $\Exp(X_t-X_0)^{\otimes2}$ commute with every orthogonal matrix and are therefore multiples of $I_d$;  the remaining updates preserve this. All these matrices commute, and for $d=2$, writing $\chi(uI_d):=\chi(u)I_d$, we have $k_\lambda(\tau,\bar\gamma)=\tau\,\chi\big(\lambda(\bar\gamma\tau)^{-1}\big)$ for $\tau, \bar\gamma\propto I_d$, so
\begin{align*}
A_t &= (1-\sparse)\, e^{-\lambda(\bar\gamma_{t-1}\tau_{R,0})^{-1}}
        + \sparse\, e^{-\lambda(\bar\gamma_{t-1}\tau_{R,1})^{-1}},\\
  C_t &= (1-\sparse)\, \lambda(\bar\gamma_{t-1}\tau_{R,0})^{-1} e^{-\lambda(\bar\gamma_{t-1}\tau_{R,0})^{-1}}
        + \sparse\, \lambda(\bar\gamma_{t-1}\tau_{R,1})^{-1} e^{-\lambda(\bar\gamma_{t-1}\tau_{R,1})^{-1}},\\
  A_t+C_t &= (1-\sparse)\,\chi\big(\lambda(\bar\gamma_{t-1}\tau_{R,0})^{-1}\big) + \sparse\,\chi\big(\lambda(\bar\gamma_{t-1}\tau_{R,1})^{-1}\big),\\
  \Exp(X_t-X_0)^{\otimes2} &= I_d + (1-\sparse)\,\tau_{R,0}\,\chi\big(\lambda(\bar\gamma_{t-1}\tau_{R,0})^{-1}\big) + (\sparse\tau_{R,0}-I_d)\,\chi\big(\lambda(\bar\gamma_{t-1}\tau_{R,1})^{-1}\big).
\end{align*}
Lines~\ref{line:At+Ct}--\ref{line:delta_bar} of \Cref{algo:se} give $\bar\delta_t = \bar\gamma_{t-1}\big(I_d-(A_t+C_t)\big)(A_t+C_t)^{-1}$, which is positive definite since $0<\chi(u)<1$ for $u>0$ implies $0\prec A_t+C_t\prec I_d$, and the SE reduces to
\begin{align*}
  \bar\gamma_t^{-1} = aI_d + (a-1)\,\bar\delta_t^{-1},\qquad
  \tau_{tt} = a\varepsilon I_d + (a-1)\,\bar\delta_t^{-2}\bar\gamma_{t-1}^{2}\Big((A_t+C_t)^{-2}\,\Exp(X_t-X_0)^{\otimes2} - \tau_{t-1,t-1}\Big).
\end{align*}

\subsubsection{Numerical experiments}
\label{section:expts_hard_threshold}

\begin{figure}[h]
  \centering
\includegraphics{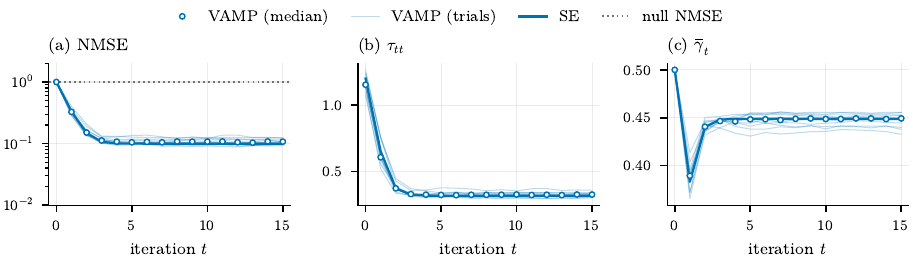}
   \vspace{-0.2cm}
  \caption{VAMP and its SE for group hard thresholding ($d=2$, $a=2$, $\lambda=2$,
    $p_{\mathsf{nz}}=0.1$, $\varepsilon=0.1$, $m=1024$). Since the SE is isotropic, each panel
    shows the scalar multiplying $I_d$: (a)~the NMSE
    $ \trace \mathbb{E}(X_t-X_0)^{\otimes2}/ \trace\Sigma_0$, (b)~$\frac{1}{d}\trace \tau_{tt}$, 
    (c)~$\frac{1}{d}\trace \overline{\gamma}_t$.
    The dotted line is the null NMSE, i.e., the NMSE~of~$\widehat{\theta}=0$.}
  \label{fig:ht-iterations}
  \includegraphics{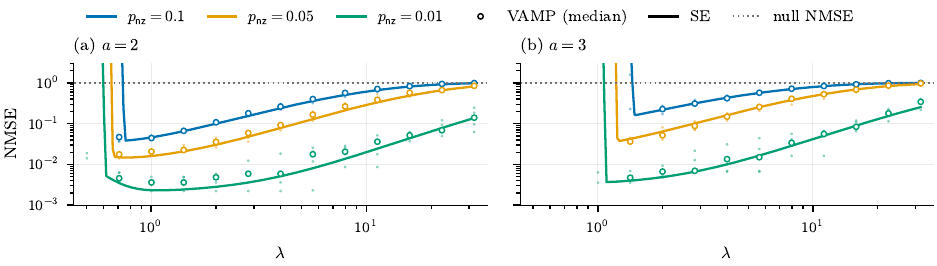}
    \vspace{-0.2cm}
  \caption{NMSE of group hard thresholding after $T=15$ iterations as a function
    of $\lambda$, for aspect ratios (a)~$a=2$ and (b)~$a=3$ and three sparsity
    levels $p_{\mathsf{nz}}$ ($d=2$, $\varepsilon=0.1$, $m=1024$).
    The $\lambda=2$ point of
    the $p_{\mathsf{nz}}=0.1$ curve in~(a) is the setting of
 \Cref{fig:ht-iterations} at $t=15$.
 }
  \label{fig:ht-tune}
\end{figure}

In \Cref{fig:ht-iterations} and \Cref{fig:ht-tune}, we adopt the orthogonal design   $\Psi\Psi^\top=I_m$ with $m=1024$ and    the Bernoulli-Gaussian prior of \Cref{sec:SE_simplified_orthogonal_hard_threshold}.   We 
initialize with $q_0=0$ and $\bar\delta_0\to\infty$, so that by \eqref{eq:initialise_hard_threshold}  $\tau_{00} = (a-1+a\eps)\, I_d,
\bar\gamma_0 = \bar\delta_0\big((a-1)I_d + a\bar\delta_0\big)^{-1} \to \tfrac1a I_d.$  
 The SE then stays isotropic, and we run VAMP  over multiple  trials. 
 
 \Cref{fig:ht-iterations} shows that the SE closely tracks VAMP at every iteration, including the initial transient phase in which  $\bar\gamma_t$ falls before it settles. The normalized mean squared error (NMSE), computed as  $\trace\Exp(X_t-X_0)^{\otimes2}/ \trace\Sigma_0$,  drops from $1$, the MSE of $\widehat\theta=0$, by around two orders of magnitude to its fixed point, which by \eqref{eq:MSE_Bern_Gaussian} is the asymptotic value of $\frac{1}{p}\frac{1}{d}\|\hat\theta-\theta_0\|^2$ as $p\to\infty$ and then $t\to\infty$.

\Cref{fig:ht-tune} sweeps $\lambda$ for three sparsity levels and two aspect ratios, 
comparing the SE after $T=15$ iterations with VAMP. The SE matches the VAMP 
median across the whole range of $\lambda$.
Below a critical $\lambda$ the NMSE diverges and exceeds the null NMSE. 
Over an intermediate range it 
lies below the null NMSE. For 
large $\lambda$, hard thresholding removes all signal and the NMSE rises to $1$. 
Sparser signals give a lower NMSE throughout.

\begin{figure}[h]
  \centering
\includegraphics{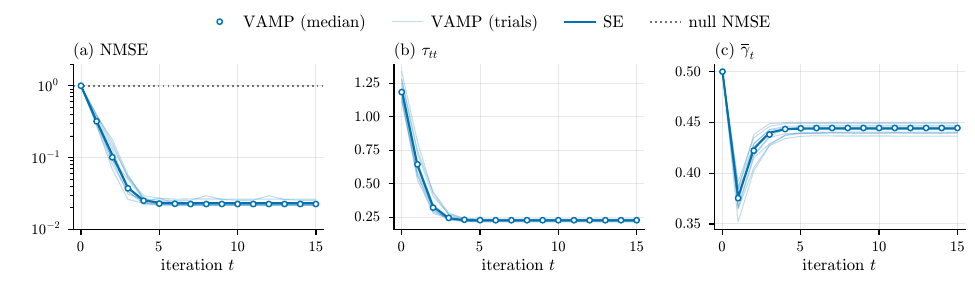}
\includegraphics{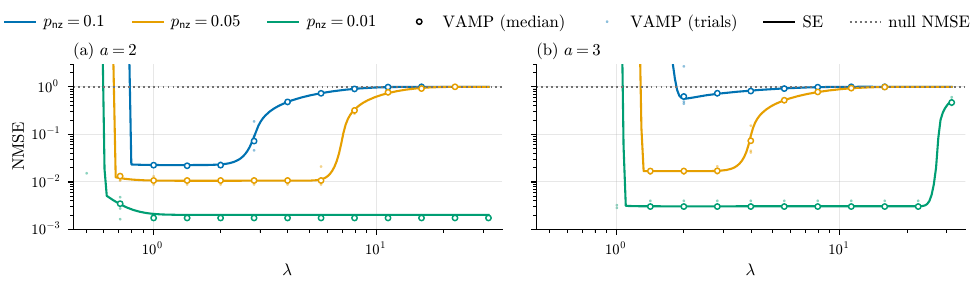}
\caption{
This figure repeats the settings of \Cref{fig:ht-iterations,fig:ht-tune} but with $X_0$ distributed as a Bernoulli-Rademacher taking values from the set $\set{(0,0), \frac{1}{\sqrt{1-p_{\mathsf{nz}}}}(\pm1,\pm1)}$ and $\mathbb{P}(X_0\neq (0,0))=p_{\mathsf{nz}}$. 
}\label{fig:bernoulli_rademacher}
\end{figure}

\Cref{fig:bernoulli_rademacher} analogously  presents the results for the Bernoulli-Rademacher signal prior.

\subsection{Ridge-regularized regression with a non-convex loss}
\label{sec:cauchy_example}
While \Cref{sec:group_sparse_example} studies problems with  non-convexity  in the regularization term, this section studies problems with non-convexity   in the loss function. We observe $y\in\Real^{m\times d}$ with rows
\begin{align*}
    y_i=\phi((\beta_0)_i;\,\xi_i)\qquad \beta_0=\Psi\theta_0, 
\end{align*}
where  $\theta_0\in\Real^{p\times d}$ is the unknown signal,      $\Psi\in\Real^{m\times p}$ is the   design, $\xi_i\in \Real^{1\times d}$ is noise, and $\phi$ is the observation model.  \Cref{sec:cauchy_regression} takes $\phi((\beta_0)_i;\,\xi_i) = (\beta_0)_i + \xi_i$ with Cauchy noise. For losses $\ell_i : \Real^{d}\to \Real$, for instance $\ell_i(\beta_i) = -\log p(y_i \mid \beta_i)$ or a robust surrogate, and for $\Lambda \in \Real_\succ^{d\times d}$, we consider the ridge-regularized M-estimator 
\begin{align}\label{eq:M_estimator_ridge_regularised}
\widehat \theta \in \underset{\theta\in\Real^{p\times d}}{\arg\min} \;\sum_{i=1}^m \ell_i(\beta_i) + \tfrac12\norm{\theta}_\Lambda^2,\qquad\text{subject to }\beta=\Psi\theta,
\end{align}
where $\norm{\theta}_\Lambda^2
=\trace(\theta\Lambda\theta^\top ) 
= \sum_{i=1}^p \|\theta_i\|_\Lambda^2$ is a weighted ridge penalty. 

This objective can be reduced to \eqref{opt:xzU} as follows. Let  $\Psi=V_{\sf L}\diag(\sigma)V_{\sf R}\tran$ be the full SVD from \Cref{section:examples}. Write $\theta = \theta_\parallel + \theta_\perp$, where the columns of $\theta_\parallel$ lie in $\operatorname{range}(\Psi^\top)=\text{range}(\Psi\pinv)$ and those of $\theta_\perp$ in $\operatorname{null}(\Psi)$. The loss part of \eqref{eq:M_estimator_ridge_regularised} depends on $\theta$ only through $\Psi\theta = \Psi\theta_\parallel$, while $\|\theta\|_\Lambda^2 = \|\theta_\parallel\|_\Lambda^2 + \|\theta_\perp\|_\Lambda^2$, so every minimizer has $\theta_\perp \!=\! 0_d$. Equivalently, at the minimizer, $\theta=V_{\sf R}\diag(\sigma)\pinv z$ for some $ z\in\Real^{m\times d}$ with $z_i = 0_d$ whenever $\sigma_i = 0$. Then $$\Psi\theta = V_{\sf L} z\quad  \text{and}\quad  \|\theta\|_\Lambda^2 =\norm{V_{\sf R}\tran  \theta}^2_\Lambda=\norm{\diag(\sigma)\pinv   z}_\Lambda^2= \sum_{\sigma_i > 0}\|\sigma_i^{-1} z_i\|_\Lambda^2\,.$$  
To apply VAMP from \Cref{algo:vamp}, we therefore set $E = \mathbb{R}^d$ and $(n, x, U) \leftarrow (m, \beta, V_{\sf L})$, so that \eqref{eq:M_estimator_ridge_regularised} becomes
$$\min_{x, z \in \Real^{m\times d}}\; F_1(x) + F_2(z) \quad \text{subject to}\quad z = U^\top x,$$
where $F_1(x) = \sum_{i=1}^m \ell_i(x_i)$ and $F_2(z) = \sum_{i=1}^m F_{2i}(z_i)$ with
$$F_{2i}(z_i) = \begin{cases} \tfrac12\|\sigma_i^{-1} z_i\|_\Lambda^2, & \sigma_i > 0, \\ \iota_{\{0_d\}}(z_i), & \sigma_i = 0, \end{cases}\qquad \text{where}\quad \iota_{C}(t)=\begin{cases}
    0 & t\in C\\
    +\infty & t\notin C
\end{cases}$$
 is the indicator function on a set $C$.

The estimate is recovered as $\widehat\theta = V_{\sf R} \diag (\sigma)^\dagger \widehat z$. Here $F_2$ is convex, while $F_1$ can be non-convex, as allowed by \Cref{assumption:structure}. The proximal operator of $F_2$ is row-wise separable: for $i \in [m]$,
$$h_i(q_i, \delta) = \operatorname{prox}_{F_{2i}}(q_i, \delta) \in \underset{z_i \in \Real^{d}}{\arg\min}\; \tfrac12\|z_i - q_i\|_\delta^2 + F_{2i}(z_i) = \sigma_i^2\, q_i\,\delta\big(\sigma_i^2\delta + \Lambda\big)^{-1},$$
which equals $0_d$ when $\sigma_i = 0$ because $\Lambda \succ 0$. The side information in $h_i(\cdot,\delta) = H(\cdot,\delta,s_i)$ is $s_i = \sigma_i$, so   $S$ has the limiting distribution of the singular values.
The next section studies   the Cauchy loss as $F_1(x) = \sum_{i=1}^m \ell_i(x_i)$.

\subsubsection{Cauchy regression}\label{sec:cauchy_regression}
A problem of particular interest is robust regression under heavy-tailed noise. Take the additive model $y_i = (\beta_0)_i + \xi_i$, where the noise  $\xi_i\in \Real^d$ is i.i.d.\ multivariate Cauchy with zero mean and shape matrix  $\varepsilon^{-1} I_d$, written   as $\xi_i \sim \mathrm{Cauchy}(0_d, \varepsilon^{-1} I_d)$. Define the Cauchy loss 
$$ \ell_\cauchy(v) := \tfrac{1}{2\varepsilon}\log\big(1 + \varepsilon\|v\|^2\big),\qquad v\in \Real^d.$$
Then $\xi_i$ has density   on $\Real^d$ proportional to $  (1 + \varepsilon\|v\|^2)^{-(d+1)/2} = \exp(-(d+1)\eps \ell_\cauchy(v))$.  We therefore set the loss function to 
$$\ell_i(x_i): = \ell_{\cauchy}(y_i - x_i), $$
which is a rescaled negative log-likelihood: $-\log p(y_i \,|\, x_i) = (d+1)\,\varepsilon\,\ell_i(x_i) + \text{constant}$, and the factor $(d+1)\varepsilon$ can be absorbed into $\Lambda$ in \eqref{eq:M_estimator_ridge_regularised}. The normalization $\frac{1}{2\varepsilon}$ is chosen so that $\ell_\cauchy(v) \to \tfrac12\|v\|^2$ as $\varepsilon \to 0$, that is, the Cauchy loss reduces to least squares and no $y_i$ is treated as an outlier. This loss is non-convex. 
The proximal operator of $F_1$ is row-wise separable and takes the form
\begin{align*}
    g_i(r_i,\gamma)=\prox_{F_{1i}}(r_i, \gamma)\in \underset{x_i\in \Real^d}{\arg\min} \;\tfrac12\norm{x_i-r_i}_\gamma^2 + \ell_{\cauchy}(y_i-x_i),\qquad\text{for }i\in[m].
\end{align*}
We can express $g_i$ through $\prox_{\ell_\cauchy}$: substitute $\tilde x_i = y_i - x_i$, then $\|x_i - r_i\|_\gamma = \|\tilde x_i - (y_i - r_i)\|_\gamma$, so
\begin{align*}
g_i(r_i,\gamma) = y_i - \underset{ \tilde x_i\in \Real^d}{\arg\min}\;\tfrac12\|\tilde x_i - (y_i - r_i)\|_\gamma^2 + \ell_\cauchy(\tilde x_i ) 
= y_i + \prox_{\ell_\cauchy}(r_i - y_i, \gamma),    
\end{align*}
where the last step holds because $\ell_\cauchy(-v)=\ell_\cauchy(v)$ for all $v\in \Real^d$.
So the side information in $g_i(\cdot,\gamma) = G(\cdot,\gamma,w_i)$ is $w_i = y_i$, and $((\beta_0)_i, w_i)$ converges to $(X_0, W)$ with $W = X_0 + \Xi$, where $\Xi \sim \mathrm{Cauchy}(0_d, \varepsilon^{-1}I_d)$ is independent of $X_0$.

\subsubsection{Onsager term in VAMP}
\label{section:correction cauchy}
This section  derives $\prox_{\ell_\cauchy}$ and the   jump correction needed in the Onsager term: first for general $\gamma \in \mathbb{R}^{d\times d}_\succ$, then explicitly  for $d = 2$ and $\gamma = \gamma_0 I_2$. 

Since $\nabla\ell_{\cauchy}(x) = \radius x$ with $\radius:= 1/(1+\varepsilon\|x\|^2) \in (0,1]$, any minimizer of $\ell_{\cauchy}(x) + \tfrac12\|x - v\|_\gamma^2$ satisfies $\radius x + (x - v)\gamma = 0$. It therefore has the form
\begin{align*}
u_\radius(v):=v\gamma(\gamma+\radius I_d)\inv
\end{align*}
where $\radius=\radius(v)>0$ is a solution to the implicit relation with $v$,
\begin{align*}
1+\eps
\norm{v\gamma(\gamma+\radius I)^{-1}}^2=\frac1{\radius},\qquad \radius\in(0,1]
\end{align*}
which achieves minimum proximal objective function.
The solution $\radius(v)$ is unique if $\gamma\succeq\frac18 I_E$. When that is violated, there can be multiple solutions $\radius(v)$ that solve the equation above for some inputs $v$. When there are multiple solutions $\radius$, we must pick the candidate $u_\radius(v)$ which achieves the global minimum of the proximal objective $\ell(u_\radius)+\half\norm{u_\radius-v}_\gamma^2$. We call the resulting solution $u(v)=\prox_\ell(v,\gamma)$. When two distinct roots $\radius$ achieve the global minimum via $u_\radius$, the proximal map is set-valued. At the jump set, we choose a measurable selection so that $u$ denotes a function of $v$, but that choice does not change the jump correction which we seek.

The above equation defining $\radius=\radius(v)$ is a polynomial in $\radius$, after clearing the denominators. Distinct roots $\radius, \radius'$ can lead to distinct global minimizers $u_\radius, u_{\radius'}$ for the proximal objective function. When two roots $\radius_-<\radius_+$ of the polynomial have the same global minimum proximal objective value, there is a jump because the proximal map selects the global optimum. In fact the proximal operator can take both $u_\pm=u_{\radius_\pm}$. The discontinuity occurs across the switching set
\begin{align*}
    \set{v:\ell(u_-)+\tfrac12\norm{u_--v}_\gamma^2 = \ell(u_+)+\tfrac12\norm{u_+-v}_\gamma^2}.
\end{align*}
In what follows, let $E=\Real^2$, because it brings out the necessary objects, the polynomial above becomes a $5^{\rm th}$ order polynomial for a general $\gamma\in\mc E_+=\Real^{2\times 2}_\succ$. Let us simplify for the case $\gamma=\gamma_0 I_2$, for $\gamma_0\in\Real_+$, for which we get a $3^{\rm rd}$ order polynomial. Note that in this isotropic case, $u(v)$ is aligned with $v$, and only its magnitude depends on the magnitude of $v.$

For $\gamma_0<1/8$, the equation can have multiple roots $\radius$ for some inputs $v$. Let $\radius_-<\radius_+$ be two distinct roots of the cubic polynomial, that attain the same value of the proximal objective function, then they satisfy
\begin{align*}
\radius_-\radius_+\bigl(1-2\gamma_0-\radius_--\radius_+\bigr)&=\gamma_0^2,\\
\log\frac{\radius_+}{\radius_-}
&=\frac{1}{\gamma_0}(\radius_+-\radius_-)(1-\radius_--\radius_+),\text{ and}\\
0<\radius_-<1-2\gamma_0&-\radius_--\radius_+<\radius_+<1.
\end{align*}
and they depend only on $\gamma_0$. Furthermore
$u$ is discontinuous across the set
\begin{align*}
J_{\prox_\ell(\cdot,\gamma_0 I)}=\set{v:\norm{v} = r_*}\qquad\text{where }r_*= 
\frac{\gamma_0+\radius_-}{\gamma_0\sqrt{\eps}} 
\sqrt{\frac{1-\radius_-}{\radius_-}} = \frac{\gamma_0+\radius_+}{\gamma_0\sqrt{\eps}} 
\sqrt{\frac{1-\radius_+}{\radius_+}}.
\end{align*}
Outside the circle, the proximal minimizer corresponds to the smallest
root of the cubic, whereas inside it corresponds to the largest root.
Their limiting values as $\norm{v}\to r_*$ are $\radius_-$ and
$\radius_+$, respectively. The constants $\radius_-$ and $\radius_+$ are their respective limiting values at the jump set. Writing $u_{\mathrm{out}}$ and
$u_{\mathrm{in}}$ for the corresponding traces, we obtain
\begin{align*}
\mathbf n(v)=\frac{v}{r_*},\qquad
[u](v)=u_{\mathrm{out}}(v)-u_{\mathrm{in}}(v)
=\kappa_*\mathbf n(v)
\quad\text{where } \kappa_* = \frac{\gamma_0r_*(\radius_+-\radius_-)}
{(\gamma_0+\radius_-)(\gamma_0+\radius_+)}>0
\end{align*}
which together yields
\begin{align*}
\mc C(\prox_\ell(\cdot,\gamma_0 I),v_0,\eps_0 I)&=\kappa_*r_*\int_0^{2\pi}
\varphi_{\eps_0 I}(r_* u(\theta)-v_0)\,
 u(\theta) \otimes u(\theta) \,d\theta \in \mc E_+\\
&=\frac{\kappa_*r_*}{2\eps_0}
e^{-\frac{r_*^2+\norm{v_0}^2}{2\eps_0}}\sbrac{
(\mathrm{B}_0(\radius)-\mathrm{B}_2(\radius))I_2
+2\mathrm{B}_2(\radius)\,e\otimes e}
\end{align*}
where $\radius=r_*\norm{v_0}/\eps_0$ and $e=v_0/\norm{v_0}.$ When $v_0=0$, choose $e$ arbitrarily on the unit circle.
Since $\mathrm B_2(0)=0$, the expression is independent of this choice.
Define
\[
d_i=\|(r_t)_i-y_i\|_\gamma,\qquad
e_i=\sqrt{\gamma_0}\frac{(r_t)_i-y_i}{d_i}\in E,\qquad
w_i=e^{-\frac{\gamma_0r_*^2+d_i^2}{2b_{t+1}^2}}.
\]
Then the correction term is 
\[
c_{t+1}
=
\frac{\kappa_*r_*\gamma_0}{nb_{t+1}^2}
\sum_{i=1}^{n}w_i
\left[
\frac{\mathrm{B}_0\round{\tfrac{r_*d_i\sqrt{\gamma_0}}{b_{t+1}^2}}-\mathrm{B}_2\round{\tfrac{r_*d_i\sqrt{\gamma_0}}{b_{t+1}^2}}}2 I_E
+\mathrm{B}_2\round{\tfrac{r_*d_i\sqrt{\gamma_0}}{b_{t+1}^2}}\,e_i\otimes e_i
\right]\in L(E;E)
\]
Here \(\mathrm{B}_0,\mathrm{B}_2\) are modified Bessel functions of the first kind, and \(r_*,\kappa_*\) are the denoiser’s jump radius and magnitude. For $d_i=0,$ we can define $e_i$ arbitrarily because the corresponding rank-1 term vanishes.

\subsubsection{State Evolution for an underparameterized orthogonal design}

We now make the State Evolution in \Cref{algo:se} explicit for a specific orthogonal design. Let $a:=p/m<1$ be the aspect ratio and consider the design matrix $\Psi$ that satisfies $\Psi\tran\Psi=I_p$. This means the singular values $\sigma_i\in\set{0,1}$. 
\begin{remark}[Overparameterized regime]
The overparameterized case where $a=p/m\ge 1$, i.e., when   $\Psi\Psi\tran=I_m$, is much simpler.
The problem has a closed form solution given by,
\begin{align*}
\widehat\beta_i=y_i-\prox_{\ell_\cauchy}(y_i,\Lambda).&\qquad\text{and }\wh\theta=\psi\tran\wh\beta
\end{align*}
The strong law of large numbers therefore gives the asymptotic prediction MSE directly:
\begin{align*}
\frac1m\|\widehat\beta-\beta_0\|^2&\to\Exp\norm{\Xi-\prox_{\ell_\cauchy}(X_0+\Xi,\Lambda)}^2\quad\text{with}\quad  \Xi\sim\text{Cauchy}(0_2,\varepsilon\inv I_d),
\end{align*}
rendering the  VAMP analysis unnecessary. One can argue that even though $\Xi$ is heavy-tailed, the shrinkage provided by the proximal operator allows the above expectation to hold if $X_0$ has a finite second moment. We thus focus on the underparameterized case $a<1$ in the rest of this section.
\end{remark}

Suppose that $\theta_0^{(p)},\sigma^{(p)}\cweak (\Theta_0,\Sigma_p)$ such that $\Theta_0$ and $\Sigma_p$ are independent. Then $z_0^{(m)}=\diag(\sigma^{(p)})\theta_0^{(p)}$, such that $z_0^{(m)},\sigma^{(m)}\cweak (Z_0,\Sigma)$, with a jointly law such that such that $Z_0|\Sigma=0=0$, and $Z_0|\Sigma=1=\Theta_0$. 
\begin{align*}
    \Sigma_0=\Cov(Z_0)=\Prob(\Sigma=1)\Cov[Z-0|\Sigma=1]+0 = a\Cov(\Theta_0)
\end{align*}
Finally, $x_0^{(m)}=Uz_0^{(m)}\cweak X_0$ where $U=[\Psi|\Psi_\perp]$ where $\Psi_\perp$ is the orthonormal basis for $\text{null}(\Psi)^\perp$. Consequently, if we assume that $x_0^{(m)}=Uz_0^{(m)}$, then we have that $X_0\sim\mc N(0,\Sigma_0)$.

We will work with $E=\Real^2$. We assume the input side ground truth converges to $X_0$ with $2^{\rm nd}$ moment $\Sigma_0$. The State evolution reduces to a recursion for $(\wb\gamma_t,\tau_{tt})$. 
The input side information is $(y_i,\xi_i)$ with the limit $W=(X_0+\Xi,\Xi)$ and the output side information is $(\sigma_i)$ with limit $S=\Sigma$ such that $\Prob(\Sigma=1)=a$, $\Xi\sim\text{Cauchy}(0,\eps\inv I_2)$. 

The following result holds under the assumption that $U$ is Haar distributed over $m\times m$ orthogonal matrices, for the model stated above.

\begin{proposition}\label{proposition:cauchy_se_simplified}
Suppose 
\Cref{algo:se} is initialized with the following $E$-valued random variables.
Let $Z_0=\Sigma\Theta_0$ where
$\Theta_0\independent\Sigma\sim\mathrm{Bern}(a)$ and $\Exp Z_0^{\otimes2}=\Sigma_0$. 
Furthermore $\Xi\independent X_0\sim\mc N(0,\Sigma_0)$. Define
\begin{align*}
V_t&\sim\mathcal N(0,\Omega_t),\qquad V_t\independent\Xi,\\
J_t&=\Omega_t^{-1}\mathbb E[V_t\otimes (\prox_{\ell_{\cauchy}}(\Xi+V_t,\bar\gamma_t)-\Xi)],
\\
F_t&=\mathbb E[(\prox_{\ell_{\cauchy}}(\Xi+V_t,\bar\gamma_t)-\Xi)^{\otimes2}].
\end{align*}
Then the following iteration in \((\Omega_t,\bar\gamma_t)\in\mc E_+\times \mc E_+\) describes the reduced form of \Cref{algo:se} initialized with
\(
\Omega_0=\tau_{00}
+(\rho_0-\Sigma_0)^*\Sigma_0^{-1}(\rho_0-\Sigma_0).
\), where $\Sigma_0,\rho_0$ are specified in line 1 of \Cref{algo:se}
\begin{subequations}
\label{eq:reduced_se_cauchy}
\begin{align}
\bar\gamma_{t+1}
&=\frac{\Lambda+(1-a)\bar\gamma_t
\bigl((J_t^*)^{-1}-I\bigr)}a,\\[3pt]
\Omega_{t+1}
&=\bar\gamma_{t+1}^{-1}
\left[
\frac{\Lambda\Sigma_0\Lambda}{a^2}
+\frac{1-a}{a}\bar\gamma_t
\left(
(J_t^*)^{-1}F_tJ_t^{-1}-\Omega_t
\right)\bar\gamma_t
\right]\bar\gamma_{t+1}^{-1}.
\end{align}
\end{subequations}
where $\Exp (R_t-X_0)^{\otimes 2}=\Omega_t$ and $
\Exp (X_t-X_0)^{\otimes 2}=F_t$. Furthermore, if $\rho_0=0$ and $\Sigma_0,M,\Lambda,\wb\gamma_0,\tau_{00}$ are scaled versions of $I_d$, then so are $\wb\gamma_t$ and $\Omega_t$ for $t\ge 1.$
\end{proposition}

\subsubsection{Numerical experiments}

We take $d=2$, an orthogonal design with $\Psi^\top\Psi=I_p$, $m=4096$ and
$p=2048$ ($a=1/2$), signal $X_0\sim\mathcal N(0,\Sigma_0)$ with
$\Sigma_0=16\,I_2$, and ridge penalty $\Lambda=\lambda I_2$. The noise is
$\xi_i\sim\mathrm{Cauchy}(0_2,\kappa I_2)$, and we fit the loss $\ell_{\mathrm{cau}}$ with
$\varepsilon=1$ for every $\kappa$.
VAMP starts from $\overline{\gamma}_0=I_2/4$, $\rho_0=0$ and $\tau_{00}=0$.
We report $\operatorname{tr}\mathbb{E}(X_t-X_0)^{\otimes 2}/\operatorname{tr}\Sigma_0$,
the NMSE  $\|\widehat\beta-\beta_0\|^2/\|\beta_0\|^2$ in the
limit; it equals~1 for $\widehat\beta=0$.
The SE is isotropic, so every matrix in \Cref{proposition:cauchy_se_simplified} is a scalar
multiple of $I_2$, and the figures show these scalars.

\begin{figure}[h]
  \centering
  \includegraphics{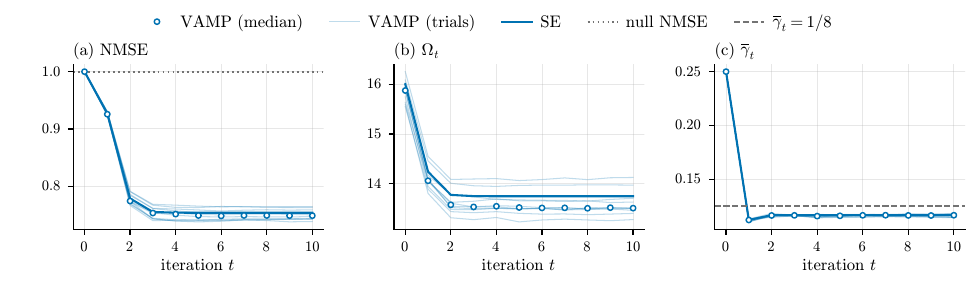}
  \vspace{-0.8cm}
  \caption{VAMP and its SE for ridge-regularized Cauchy regression ($d=2$,
    $a=1/2$, $\Lambda=0.05\,I_2$, $\Sigma_0=16\,I_2$, noise
    $\xi_i\sim\mathrm{Cauchy}(0_2,16 I_2)$, loss $\ell_{\mathrm{cau}}$
    with $\varepsilon=1$, 
    $p=2048$, $\overline{\gamma}_0=I_2/4$, $\rho_0=0$, $\tau_{00}=0$). Since the SE is
    isotropic, each panel shows the scalar multiplying $I_d$: (a)~the NMSE
    $\operatorname{tr}\mathbb{E}(X_t-X_0)^{\otimes2}/\operatorname{tr}\Sigma_0$, with
    $x_0=0$ so that it starts at~1, (b)~$\Omega_t$,  (c)~$\overline{\gamma}_t$.
    The dotted line is the NMSE
    of $\widehat{\beta}=0$; below the dashed line $\overline{\gamma}_t=I_d/8$ the
    proximal operator of $\ell_{\mathrm{cau}}$ is discontinuous.}
  \label{fig:cauchy-iterations}
  \includegraphics{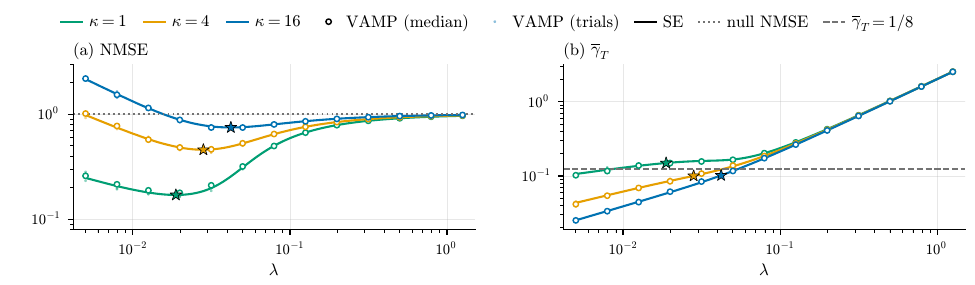}
    \vspace{-0.8cm}
  \caption{(a)~NMSE and (b)~$\overline{\gamma}_T$ after $T=10$ iterations as a
    function of $\lambda$ ($\Lambda=\lambda I_2$), for three noise shapes $\kappa I_2$; all
    other parameters as in \Cref{fig:cauchy-iterations}.
    Stars mark the  optimal $\lambda$, at which  the
    proximal operator is continuous for $\kappa=1$ ($\overline{\gamma}_T\approx0.15$)
    and discontinuous for $\kappa=4$ and $\kappa=16$  ($\overline{\gamma}_T\approx0.10<1/8$). 
    The $\lambda=0.05$ point of the
    $\kappa=16$ curve is the setting of \Cref{fig:cauchy-iterations} at
    $t=10$.
    }
  \label{fig:cauchy-tune}
\end{figure}

Figure~\ref{fig:cauchy-iterations} fixes $\kappa=16$ and $\lambda=0.05$. The SE
tracks VAMP at every iteration. After the first iteration $\overline{\gamma}_t$
falls below $1/8$ and stays there, so the proximal operator of
$\ell_{\mathrm{cau}}$ is discontinuous at every later iteration and the jump
correction is active. The SE still predicts the NMSE, $\Omega_t$ and
$\overline{\gamma}_t$ closely, which converge within a few iterations.

Figure~\ref{fig:cauchy-tune} sweeps $\lambda$ for $\kappa\in\{1,4,16\}$ and
compares the SE after $T=10$ iterations with VAMP. For each $\kappa$ the NMSE is
U-shaped in $\lambda$ and approaches the null value~1 as $\lambda$ grows; for
$\kappa=16$ and very small $\lambda$ it is worse than $\widehat\beta=0$. For the
lightest-tailed noise, $\kappa=1$, the optimal $\lambda\approx0.02$ gives
$\overline{\gamma}_T\approx0.15>1/8$, so the proximal operator is continuous
there. For $\kappa=4$ and $\kappa=16$, the optimal $\lambda\approx0.03$ and
$\approx0.04$ give $\overline{\gamma}_T\approx0.10<1/8$. With heavier-tailed
noise the optimal regularization therefore lies in the regime where the
proximal operator is discontinuous: our analysis covers this regime, but
existing results for Lipschitz denoisers do not.

\section{Large-System Limit of VAMP}
\label{section:state evolution}
We introduce  a few concepts to formally state our result in \Cref{theorem:state evolution}.

\begin{definition}[Complete convergence]\label{def:complete convergence}
A sequence of $E$-valued random variables $\set{X_n}_{n\in\Natural}$, is said to converge to a deterministic limit $x\in E$ completely, denoted $X_n\cto x$, if $Y_n\asto x$ for every sequence
$\set{Y_n}_{n\in\Natural}$ defined on a common probability
space and satisfying $Y_n\overset{d}=X_n$ for every $n$. 
\end{definition}
We also denote the above property as $\clim_{n\to\infty} X_n= x$. 

\begin{definition}[Weak complete convergence]\label{def:weak complete convergence}
A sequence of random vectors $\set{x^{(n)}}_{n\in\Natural}$, with $x^{(n)}\in E^{n}$ is said to converge to an $E$-valued random variable $X$, weakly completely, if
\begin{align*}
\frac1n\sum_{i=1}^n \psi(x^{(n)}_i) \cto \Exp \psi(X)\quad\text{for all }\psi\in C_b(E;E'),\qquad\equiv\qquad\textbf{Notation: } x^{(n)}\cweak X,
\end{align*}
where $C_b$ is the space of bounded continuous functions. 
\end{definition}
Let $F_n:E\to[0,1]$ be the empirical cumulative distribution function obtained from $x^{(n)}$, i.e., $F_n(t)=\frac1n\sum_{i=1}^n\indicator\set{x^{(n)}_i\le t}$, and let $F$ be the cumulative distribution function of $X$.
\begin{remark}\label{remark:weak convergence}
 Since $x^{(n)}$ is random, so is $F_n$. The above notion of convergence is equivalent to $F_n\cto F$ in the weak topology.
\end{remark}

\begin{definition}[Pseudo-Lipschitz test functions] A function $\psi:E\to E'$ is said to be pseudo-Lipschitz of order $k$ if for some $0<L<\infty$ we have
\begin{align*}
    \norm{\psi(u)-\psi(v)}\leq L(1+\norm{u}^{k-1}+\norm{v}^{k-1})\norm{u-v}.
\end{align*}
\end{definition}
We denote the above by $\psi\in\textrm{PL}(k;E,E')$. Note that $\textrm{PL}(1)$ is the space of Lipschitz functions.

\begin{definition}[PL($k$) convergence]\label{def:pl convergence}
A sequence of random vectors $\set{x^{(n)}}_{n\in\Natural}$, with $x^{(n)}\in E^{n}$ is said to converge to an $E$-valued random variable $X$, with finite $k^{\rm th}$ moment, in the PL($k$) sense, if
\begin{align*}
    \frac1n\sum_{i=1}^n \psi(x^{(n)}_i) \cto \Exp \psi(X)\quad\text{for all }\psi\in\textrm{PL}(k;E,E'),\qquad\equiv\qquad
    \textbf{Notation: } x^{(n)}\cplconv{k} X.
\end{align*}
\end{definition}
\begin{remark}[\!\texorpdfstring{\cite[Corollary B.21(b)]{feng2022unifying}}{Feng}]\label{remark:equivalence PLconv moments}
The above notion of convergence is equivalent to 
\begin{enumerate}
\item $x^{(n)}\cweak X$ and $\frac1n\sum_{i=1}^n\norm{x^{(n)}_i}^k\cto \Exp\norm{X}^k<\infty.$ This is a special case of \Cref{proposition:moments durrett}.
\item $W_k(F_n,F)\cto 0$, where $W_k$ is the Wasserstein-$k$ metric.
\end{enumerate}
\end{remark}
\begin{restatable}[Moment based  characterization of PL$(k)$ convergence]{proposition}{propPLk}
\label{proposition:plk equivalence}
Suppose $x^{(n)}\cweak X$ and $\frac1n\sum_{i=1}^n\norm{x^{(n)}_i}^k\cto \Exp \norm{X}^k<\infty$. Then $x\plconv{k} X$.
\end{restatable}
We provide the proof of this proposition in Appendix \ref{section:useful probability}.

\begin{definition}[Truly denoising]\label{def:truly denoising} We say  $\zeta:E\to E$ is \textit{truly denoising}, if for independent $E$-valued random variables $X_0$ and $N$ such that $\Exp\norm{X_0}^2<\infty$, and $N\sim\mc N(0,\tau)$ for some $\tau\in\mc E_+$, we have
\begin{align}\label{eq:truly denoising}
    \Cov(\zeta(X_0+N)-X_0)\prec \tau.
\end{align}
\end{definition}

\subsection{Assumptions}
We consider a sequence of problems indexed by $n\in\Natural$ solved using \Cref{algo:vamp} for each $n$.

\begin{assumption}[Regularity of proximal operators]\label{assumption:regularity}
For every $\gamma,\delta\in\mathcal E_+$, $u_w\in E_w$, and $u_s\in E_s$, there exist Lebesgue-null sets $J^G_{\gamma,u_w},J^H_{\delta,u_s}\subset E$ such that
$G$ is jointly continuous at $(v,\gamma,u_w)$ whenever $v\notin J^G_{\gamma,u_w}$ and $H,H'$ are jointly continuous at $(v,\delta,u_s)$ whenever $v\notin J^{H}_{\delta,u_s}$.
\end{assumption}

\begin{assumption}[Ground truth, perturbations, and initialization] \label{assumption:ground truth}
We assume that there exist $E$-valued random variables $X_0,Z_0,R_0$, an $E_w$-valued random variable $W$, and an $E_s$-valued random variable $S$, such that the following weak convergences holds,
\begin{align*}    
&(x_0^{(n)},r_0^{(n)},w^{(n)}) \cweak (X_0,R_0,W).\\
&(z_0^{(n)},s^{(n)}) \cweak (Z_0,S).
\end{align*}
Furthermore, for some $k_x,k_z\ge 2$, we have
\begin{subequations}
\begin{align}
&\tfrac1n\sum_{i=1}^n\norm{(x_0^{(n)})_i}^{k_x} \cto \Exp \norm{X_0}^{k_x}<\infty,
\label{eq:x0 empirical moment}\\
&\tfrac1n\sum_{i=1}^n\norm{(r_0^{(n)})_i}^{k_x} \cto \Exp \norm{R_0}^{k_x}<\infty,
\label{eq:r0 empirical moment},\qquad \gamma_0^{(n)}\cto\bar\gamma_0\\
&\tfrac1n\sum_{i=1}^n\norm{(z_{0}^{(n)})_i}^{k_z} \cto \Exp \norm{Z_0}^{k_z}<\infty, \text{ and}
\label{eq:z0 empirical moment}\\
&\tfrac1n\sum_{i=1}^n (x_0^{(n)})_i\otimes (x_0^{(n)})_i=\tfrac1n\sum_{i=1}^n (z_0^{(n)})_i\otimes (z_0^{(n)})_i \quad \text{ for all }n\in\Natural.
\end{align}
\end{subequations}
The law of $R_0|(X_0,W)$ is $\mc N(X_0\Sigma_0\inv\rho_0,\tau_{00})$ for some $\rho_1\in L(E;E)$ and $\tau_{00}\in\mc E_+$ specified in \Cref{algo:se}, written as lines \ref{line:def-N0} and \ref{line:def-R0}.
\end{assumption}

\begin{assumption}[Domain of $G$]\label{assumption:domain}
For $P_W$-almost every $w$,
$G(\cdot,\gamma,w)$ is a proximal selection of $F_w$ such that the affine hull of the domain of $F_w$ is $E$.
\end{assumption}

\begin{assumption}[Growth rates of proximal operators]\label{assumption:growth rates}
Suppose in addition to \Cref{assumption:ground truth}, for some $k_w,k_s\ge 0$, we have the following empirical convergences
\begin{subequations}
\begin{align}
    &\tfrac1n\sum_{i=1}^n\round{1+\norm{w_i^{(n)}}}^{k_w} \cto \Exp\round{1+\norm{W}}^{k_w}<\infty,\text{ and} \label{eq:assumption_W}\\
    &\tfrac1n\sum_{i=1}^n\round{1+\norm{s_i^{(n)}}}^{k_s} \cto \Exp\round{1+\norm{S}}^{k_s}<\infty
\end{align}
\end{subequations}
and the correspondingly the following growth rate bounds on the proximal operators. For every compact $K\subset \mc E_+$, and for all $\gamma,\delta\in K$, and all $w\in E_w$, $s\in E_s$, and $v\in E$,
\begin{subequations}
\begin{align}
&\norm{G(v,\gamma,w)}\leq C_{K,1}
(1+\norm{v}+(1+\norm{w})^{k_w/k_x}),\label{eq:G_growth_rate}\\
&\norm{H(v,\delta,s)}\leq C_{K,2}
(1+\norm{v}+(1+\norm{s})^{k_s/k_z}),\label{eq:H_growth_rate}
\end{align}
\end{subequations}
\end{assumption}
\begin{remark} $k_w$ (or $k_s)$$=0$ are allowed, i.e., no moments of $W$ (or $S$) are assumed to be finite.
\end{remark}

\begin{assumption}[Shrinkage of $G$]
\label{assumption:shrinkage}
For every $\gamma,\tau\in\mc E_+$ and $\rho\in L(E;E)$, let
$R=X_0\rho+N$, where $N\sim\mc N(0,\tau)$ is independent of $(X_0,W)$.
Then,
\begin{align}
& \Exp[\Cov(G(R,\gamma,W)| X_0,W)] \prec \tau.\label{eq:G shrinkage}
\end{align}
A sufficient condition for the above condition is that $G$ is truly denoising, as   in \Cref{def:truly denoising}.
\end{assumption}

\begin{assumption}[Non-degeneracy of proximal operators under gaussian noise]
\label{assumption:non-degeneracy}
For \(\gamma,\delta,\tau,\mu\in\mathcal E_+\) and \(K_x,K_z\in L(E;E)\), let
\(R=X_0K_x+N\) and \( Q=Z_0K_z+M,\)
where \(N\sim\mathcal N(0,\tau)\) is independent of \((X_0,W)\) and \(M\sim\mathcal N(0,\mu)\) is independent of \((Z_0,S)\). Assume
\begin{subequations}
\begin{align}
&\Exp\Cov\sbrac{(R,G(R,\gamma,W))|X_0,W}\succ0,\quad\text{and}\label{eq:G non-degenerate}\\
&\Exp\Cov\sbrac{(Q,H(Q,\delta,S))|Z_0,S}\succ0.\label{eq:H non-degenerate}
\end{align}
\end{subequations}
\end{assumption}

\subsection{Main result}\label{sec:main_result}
The VAMP iterates progressively reveal information about $U^{(n)}$ through its action in the updates. For each $n$ define the following VAMP filtration $\mc H_t^{(n)}\subseteq\mc H_0^{(n)}\vee\sigma(U^{(n)})$ and a corresponding State Evolution filtration $\mc H_t\subseteq \mc H_0\vee\sigma(M_j,N_j;j\ge 1)$ where $M$ and $N$ are jointly gaussian random variables.
\begin{align*}
\mc H_0^{(n)}
&:=\sigma\!\round{x_0^{(n)},z_0^{(n)},w^{(n)},s^{(n)},r_0^{(n)},\gamma_0^{(n)}},
& \mc H_0 &:=\sigma\!\round{X_0,Z_0,W,S,R_0}\\
\mc H_{t-\nicefrac12}^{(n)}
&:=\mc H_{t-1}^{(n)}
\vee\sigma\round{q_{t}^{(n)}},
&\mc H_{t-\nicefrac12} &:= \mc H_{t-1}\vee\sigma (Q_t)
\quad t\ge1, \\
\mc H_{t}^{(n)}
&:=\mc H_{t-\nicefrac12}^{(n)}
\vee\sigma\round{r_{t}^{(n)}},
&\mc H_{t}&:= \mc H_{t-\nicefrac12}\vee\sigma (R_t)
\quad t\ge1.
\end{align*}
where for $\sigma$-algebras $\mc F,\mc G$, we denote by $\mc F\vee\mc G=\sigma(\mc F\cup\mc G)$. Consequently we have, for $t\ge0$, the inclusion $\mc H_t^{(n)}\subseteq \mc H_{t+\nicefrac12}^{(n)}\subseteq\mc H_{t+1}^{(n)}$ and similarly $\mc H_t\subseteq \mc H_{t+\nicefrac12}\subseteq\mc H_{t+1}$.

The State Evolution \Cref{algo:se} defines a sequence of random variables with a dependency structure. These variables describe the asymptotic statistical behaviour of the iterates of the Vector Approximate Message Passing in \Cref{algo:vamp}.

To state our result precisely, recall the notation $x^{(n)}\plconv{k}X$ in \Cref{def:pl convergence} and $x^{(n)}\cweak X$ in \Cref{def:weak complete convergence}. For ease of notation we define \(q_0^{(n)}=0\in E^n,\) and \(Q_0=0\in E
\).

\begin{theorem}[VAMP asymptotics]\label{theorem:state evolution}
Consider a sequence of $\set{x^{(n)}_0,U^{(n)},z_0^{(n)}}_{n\in\Natural}$, with $U^{(n)}\in\Real^{n\times n}$ such that $U^{(n)}|\mc H_0^{(n)}$
is conditionally Haar distributed on $n\times n$ orthogonal matrices that satisfy $z^{(n)}_0={U^{(n)}}\tran x^{(n)}_0$.
For each $n\in\Natural$, suppose \Cref{algo:vamp,algo:se} are run with  proximal operators $g,h$ satisfy \Cref{assumption:regularity,assumption:ground truth,assumption:domain,assumption:growth rates,assumption:shrinkage,assumption:non-degeneracy}. 
Then, for every fixed
$t\ge1$ there exist deterministic bandwidth sequences
$b_1^{(n)},\ldots,b_t^{(n)}$, each $\downarrow 0$ as $n\to\infty$,
for which the following holds with $k_x,k_z\ge2$ and $k_w,k_s\ge 0$.

The random variables in 
\Cref{algo:se} admit a coupling such that 
$(N_1,\ldots,N_t)\sim\mc N(0,\tau_t)\independent (X_0,W,N_0)$ and
$(M_1,\ldots,M_t)\sim\mc N(0,\mu_t)\independent (Z_0,S)$ with $\tau_t,\mu_t\in\mc E^t_+$.
For this coupling,
\begin{align}
&\round{w^{(n)},\set{x_j^{(n)}\!,r_j^{(n)}}_{j=0}^t}
\cweak \round{W,\set{X_j,R_j}_{j=0}^t},\quad\text{ and }
&\set{x_j^{(n)}\!,r_j^{(n)}}_{j=0}^t
\plconv{k_x} \set{X_j,R_j}_{j=0}^t,\label{eq:plkx}\\
&\round{s^{(n)},\set{z_j^{(n)},q_j^{(n)}}_{j=0}^t}
\cweak \round{S,\set{Z_j,Q_j}_{j=0}^t},\quad\text{ and }
&\set{z_j^{(n)},q_j^{(n)}}_{j=0}^t
\plconv{k_z} \set{Z_j,Q_j}_{j=0}^t.\label{eq:plkz}
\end{align}
Furthermore, for every $1\le j\le t$,
\begin{align}
(\alpha_j^{(n)}+c_j^{(n)},\eta_j^{(n)},\delta_j^{(n)})
&\cto (A_j+C_j,\bar\eta_j,\bar\delta_j), \quad\text{and}\label{eq:alpha_convergence}\\
(\beta_j^{(n)},\nu_j^{(n)},\gamma_j^{(n)})
&\cto (B_j,\bar\nu_j,\bar\gamma_j), \label{eq:beta_convergence}
\end{align} where $\wb\delta_j,\wb\gamma_j\succ0$.
\end{theorem}
The following corollary holds immediately by definition of the $\PL(k)$ convergence as in \Cref{def:pl convergence}.
\begin{corollary} For any $\psi\in \mathrm{PL}(k,E\times E;\Real)$, we have
    $\frac1n\sum_{i=1}^n \psi(x_t^{(n)},x_0) \cto \Exp \psi(X_t,X_0)$.
\end{corollary}

\begin{remark}[Information revelation] Note that for $t\geq1,$ $q_t^{(n)}\in\mc H^{(n)}_{t-\nicefrac12}$, and $\delta_t^{(n)}\in\mc H_{t-1}^{(n)}$, whereby $z_t^{(n)},\nu_t^{(n)},\beta_t^{(n)},\wt q_t^{(n)},\gamma_t^{(n)}\in\mc H^{(n)}_{t-\nicefrac12}$ since they are deterministic functions of $(q_t^{(n)},\delta_t^{(n)},s^{(n)})$. Similarly $r_t^{(n)}\in\mc H_t^{(n)}$, and $\gamma_t^{(n)}\in\mc H^{(n)}_{t-\nicefrac12}$, whereby $x_t^{(n)},\eta_t^{(n)},\alpha_t^{(n)},c_t^{(n)},\wt r_t^{(n)},\delta_t^{(n)}\in\mc H^{(n)}_{t-1}$ since they are deterministic functions of $(r_{t-1}^{(n)},\gamma_{t-1}^{(n)},w^{(n)})$. 
Each iteration reveals the action of $U^{(n)}$ on at most $2d$ additional directions. Thus, for fixed $t$, the observations constrain its action on a subspace of dimension $O(td)$, leaving a Haar-distributed orthogonal transformation between the corresponding orthogonal complements.
\end{remark}

\begin{remark}[Choice of bandwidth sequence]
\label{remark:bn_deterministic}
Note that the bandwidth sequence $\set{b_t^{(n)}}_{n\in\Natural}$ depends on the law of $\left((r_{t-1}^{(n)})_i, (b_t^{(n)})^2(\gamma_{t-1}^{(n)})^{-1}\right)$. 
The choice of the sequence depends on the rate of convergence to the State Evolution. We would need to show a guarantee similar to \cite{cademartori2024non} and obtain a concentration bound of the form
\begin{align*}
\Prob\round{\abs{\frac1n\sum_{i=1}^n\psi\Big((x_0^{(n)})_i,(r_t^{(n)})_i,(x_{t}^{(n)})_i\Big)-\Exp \psi(X_0,R_t,X_t) }>\eps}\leq \textsf{C}_\psi\textsf{C}'\textsf{C}_t\exp(-\textsf{c}'\textsf{c}_t n\eps^2)\,,
\end{align*}
where $\textsf{c},\textsf{c}',\textsf{C},\textsf{C}'$ are universal constants and $\textsf{C}_t=\textsf{C}^{2t}(t!)^{16}$ and $\textsf{c}_t=\textsf{c}^{-2t}(t!)^{-26}$.
Then a sufficient  condition on the bandwidth $b^{(n)}$ is
\(\round{{(\log n+\log(1+\textsf{C}_t))}/{(\textsf{c}_t n)}}^{1/(2d+2)}=o(b^{(n)})\), and \(b^{(n)}=o(1)\).
For fixed $t$, this simplifies to
\(b^{(n)}=\omega\round{\round{{\log n}/{n}}^{1/(2d+2)}}\), and \(b^{(n)}=o(1)\). Since we do not derive a concentration inequality in this paper, we leave the choice of the decay rate of the length-scale sequence for future work. In our experiments we use $b_n=n^{-\nicefrac16}.$
\end{remark}

\section{Proof of \texorpdfstring{\Cref{theorem:state evolution}}{state evolution}}
\label{section:proof sketch}

We use the same outline as the proof of the state evolution of VAMP given in \cite{rangan2019vector} which is largely based on the same induction argument as \cite{bayati2011dynamics} that analyzes AMP using induction via the Gaussian conditioning technique  \cite{bolthausen2014iterative}. We point out steps that need modification. We refer the reader to a recap of the conditional distribution of Haar random matrices under linear constraints in \cref{section:haar conditioning}.

The induction identifies the empirical limits of $r_t^{(n)}$ and $q_t^{(n)}$ through the conditional Gaussian representations $R_t$ and $Q_t$ in \Cref{algo:se}. The Haar conditioning argument conditions on $\mc H_{t-\nicefrac12}^{(n)}$ and $\mc H_{t-1}^{(n)}$, respectively. The induction also tracks the joint empirical behaviour across iterations and the inner products appearing in the Haar decomposition.

We will need the following continuous mapping theorem proved in \Cref{section:useful probability}.
\begin{restatable}[Continuous mapping theorem for empirical convergence]{proposition}{propCMT}\label{proposition:continuous mapping}
Let $E_x,E_y$ be finite-dimensional Euclidean spaces, and
suppose $x^{(n)}\cweak X$. Let $h:E_x\to E_y$ be measurable, and
let $D_h\subset E_x$ denote the set of discontinuity points of $h$. Suppose
$\Prob(X\in D_h)=0$. Let $y^{(n)}\in E_y^n$ such that $y^{(n)}_i = h(x^{(n)}_i)$.
Then, $(x^{(n)},y^{(n)})\cweak (X,h(X)).$
\end{restatable}

\begin{restatable}[Slutsky's lemma for empirical convergence]{proposition}{propslutsky}\label{proposition:slutsky}
Suppose $x^{(n)}\cweak X\in E$ and $V_n\cto V\in L(E;E')$, then $(x^{(n)},x^{(n)}V_n)\cweak (X,XV)$. Furthermore, if $x^{(n)}\plconv{k} X\in E$ for some $k\ge1$, then $(x^{(n)},x^{(n)}V_n)\plconv{k} (X,XV)$.
\end{restatable}

We will also need the following version of the dominated convergence theorem which is a corollary to \Cref{proposition:moments durrett} detailed in \Cref{section:useful probability}. A proof of this corollary is given in \Cref{section:useful probability}.
\begin{restatable}[Empirical dominated convergence]{corollary}{corollaryEmpiricalConvergence}\label{corollary:empirical dominated convergence}
Let $\mu_n$ be a sequence of random empirical measures converging weakly completely
to $\mu$. Suppose $m\ge0$ is continuous and $\mu_n m\cto\mu m<\infty$.
If $f$ is measurable, continuous $\mu$-almost everywhere, and
$\abs{f}\le m$, then $\mu_n f\cto\mu f$.
\end{restatable}
In other words, if $x^{(n)}\cweak X$, and $\frac1n\sum_{i=1}^n m(x_i^{(n)})\cto \Exp m(X)$, and $f$ is measurable, $P(X\in D_f)=0$, and $\abs{f}\le m$, then $\frac1n\sum_{i=1}^n f(x_i^{(n)})\cto \Exp f(X)$. For $m(t)=\norm{t}^k$, this can be used to show $x^{(n)}\plconv{k} X$.

We will also need a lemma to show the consistency of the Onsager term.

\subsection{Consistency of the KDE approximated  Onsager term with the correction}
In \Cref{algo:vamp}, we use kernel density estimation (KDE) to add the correction to the Onsager term. The following lemma states that this correction is consistent with the corresponding State Evolution term if the bandwidth $b^{(n)}$ is chosen to be small enough. 
Let $\alpha_t(b), c_t(b)$ for a length-scale $b$ be defined as  in \eqref{eq:kde_onsager}, and $A_t, C_t$ be defined as in \eqref{eq:SE_At_Ct_def}.
\begin{restatable}[Consistency of the KDE approximation for the Onsager term]{lemma}{lemmaconsistencyonsager}
\label{lemma:consistency of onsager terms}
Fix $t\ge 0$, and suppose
\begin{align}
  &\gamma_t^{(n)}
  \cto \bar\gamma_t\in\mc E_+, \label{eq:hypothesis_onsager_gamma}\\
  &(x_0^{(n)},r_t^{(n)},w^{(n)})
  \cweak (X_0,R_t,W), \label{eq:hypothesis_onsager_weak}\\
  &\frac1n\sum_{i=1}^n
  \|(r_t^{(n)})_i\|^{k_x}
  \cto \Exp\|R_t\|^{k_x}<\infty,\label{eq:hypothesis_onsager_r}
\end{align}
where $k_x > 1$, $k_w\ge0$, and $\cweak$ denoting convergence as in \Cref{def:weak complete convergence}, and further assume
\begin{align*}
  R_t=X_0\Sigma_0\inv\rho_t+N_t,\qquad
  N_t\sim\mc N(0_E,\tau_{tt}),\quad
  \tau_{tt}\in\mc E_+,\quad
  N_t\independent(X_0,W).
\end{align*}
If $G,X_0,W$ satisfy \Cref{assumption:regularity,assumption:growth rates,assumption:ground truth},
then $A_{t+1}$ and $C_{t+1}$ are finite, and there exists
a deterministic sequence $b^{(n)}_t\downarrow0$ such that
\begin{align}
  \alpha_{t+1}^{(n)}(b^{(n)}_{t+1})+c_{t+1}^{(n)}(b^{(n)}_{t+1})
  \cto A_{t+1}+C_{t+1}.
\end{align}
\end{restatable}
We provide the proof of this lemma  in \Cref{app:proof_consistency_onsager}. 
See \Cref{remark:smooth_onsager} for a discussion on how to handle $G$ with bounded Jacobians $G'.$

\subsection{Proof by induction}

We record the history in the following matrices.
For $t\ge1$, stack the following vectors columnwise
\begin{align*}
\mathbf{R}\n_{t-1} &= 
\begin{bmatrix}
x_0\n & \round{\wt r_j\n-x_0\n\Sigma_0\inv\pi_j}_{1\le j < t}
&  \round{r_j\n-x_0\n\Sigma_0\inv\rho_j}_{1\le j < t}
\end{bmatrix}\\
\mathbf{R}\n_{t-\nicehalf} &= 
\begin{bmatrix}
x_0\n & \round{\wt r_j\n-x_0\n\Sigma_0\inv\pi_j}_{1\le j < t+1}
&  \round{r_j\n-x_0\n\Sigma_0\inv\rho_j}_{1\le j < t}
\end{bmatrix}\\
\mathbf{Q}\n_{t-1} &= 
\begin{bmatrix}
z_0\n & \round{q_j\n-z_0\n\Sigma_0\inv\pi_j}_{1\le j < t}
&  \round{\wt q_j\n-z_0\n\Sigma_0\inv\rho_j}_{1\le j < t}
\end{bmatrix}\\
\mathbf{Q}\n_{t-\nicehalf} &= 
\begin{bmatrix}
z_0\n & \round{q_j\n-z_0\n\Sigma_0\inv\pi_j}_{1\le j < t+1}
&  \round{\wt q_j\n-z_0\n\Sigma_0\inv\rho_j}_{1\le j < t}
\end{bmatrix}
\end{align*}
At \(t-\tfrac12\), the \(\pi_j\) blocks of both matrices include \(j=t\), while the \(\rho_j\) blocks still run through \(j=t-1\).
Observe that $\mathbf{Q}\n_{t-1},\mathbf{R}\n_{t-1}\in\mc H_{t-1}\n$, and $\mathbf{Q}\n_{t-\nicehalf},\mathbf{R}\n_{t-\nicehalf}\in\mc H_{t-\nicehalf}\n$. The event $U\n \mathbf{Q}\n_{t-1}=\mathbf{R}\n_{t-1}$ lies in $\mc H_{t-1}\n$ whereas the event $U\n \mathbf{Q}\n_{t-\nicehalf}=\mathbf{R}\n_{t-\nicehalf}$ lies in $\mc H_{t-\nicehalf}\n$.

We paraphrase \cite[Lemmas 5-8]{rangan2019vector}  below in our notation without a proof.
\begin{lemma}[Empirical limit of a conditionally Haar rotation]
\label{lemma:haar empirical limit}
Let $E$ be a real Euclidean space of dimension $d$, and fix $m\ge1$. For each $n$, let
$A\n,B\n\in L(E^m;\Real^n),
v\n,\tilde v\n\in E^n,
h\n,\tilde h\n\in(E^m)^n$
be $\mc H\n$-measurable. Suppose that
$(A\n)^*A\n=(B\n)^*B\n$ almost surely
and that $U\n|\mc H\n$ is Haar distributed
on the orthogonal matrices satisfying $U\n B\n=A\n$. Let
\begin{align*}
p\n
&:=B\n(A\n)^\dagger v\n,&
V_n
&:=\tfrac1n(v\n)^*
  (I_n-A\n(A\n)^\dagger)v\n,\\
\tilde p\n
&:=A\n(B\n)^\dagger\tilde v\n,&
\tilde V_n
&:=\tfrac1n(\tilde v\n)^*
  (I_n-B\n(B\n)^\dagger)\tilde v\n.
\end{align*}
Let $H,\tilde H$ be $E^m$-valued and $P,\tilde P $ be $E$-valued random variables.
\begin{enumerate}[label=(\alph*)]
\item\label{part:U_tran} If $(h\n,p\n)\cweak(H,P)$, $V_n\cto V\in\mc E_\ge $, and $G\sim\mc N(0,V)\independent (H,P)$, then $(U\n)\tran v\n-p\n\plconv{k} G$ for every fixed $k<\infty$ and $(h\n,p\n,
     (U\n)\tran v\n-p\n,
     (U\n)\tran v\n)
\cweak(H,P,G,P+G).$
\item\label{part:U} If 
$(\tilde h\n,\tilde p\n)\cweak(\tilde H,\tilde P)$, $\tilde{V}_n\cto \tilde{V}\in\mc E_\ge$, and $\tilde G\sim\mc N(0,\tilde V)\independent (\tilde H,\tilde P)$, then $U\n \wt v\n-\wt p\n\plconv{k} \wt G$ for every fixed $k<\infty$ and
$(\tilde h\n,\tilde p\n,
     U\n\tilde v\n-\tilde p\n,
     U\n\tilde v\n)
\cweak(\tilde H,\tilde P,\tilde G,\tilde P+\tilde G).$
\end{enumerate}
\end{lemma}

\subsubsection{Identifying the full block covariance matrix} 
\Cref{theorem:state evolution} claims $(N_1,\ldots,N_t)\sim\mc N(0,\tau_t)$ and $(M_1,\ldots,M_t)\sim\mc N(0,\mu_t)$ without giving precise expressions. In addition to proving the Theorem statement, we will also identify these block covariance matrices, for $1\le j,j'\le t$,
\begin{align}
\mu_{jj'}
&=\Exp[\widetilde R_j\otimes\widetilde R_{j'}]
  -\pi_j^*\Sigma_0^{-1}\pi_{j'}, \label{eq:mu_def}\\
\tau_{jj'}
&=\Exp[\widetilde Q_j\otimes\widetilde Q_{j'}]
  -\rho_j^*\Sigma_0^{-1}\rho_{j'}. \label{eq:tau_def}
\end{align}
To that end, we will also need to show the auxiliary covariance identities for $1\le j,j'\le t.$
\begin{align}
\Exp[N_0\otimes\widetilde R_{j}]&=0,
\label{eq:N0R_uncorrelated}\\
\Exp[M_j\otimes\widetilde Q_{j'}]&=0, 
\label{eq:MQ_uncorrelated}\\
\Exp[N_j\otimes\widetilde R_{j'}]&=0, 
 \label{eq:NR_uncorrelated}\\
\Exp[N_0\otimes N_j]&=0.
\label{eq:N0N_uncorrelated}\\
\Exp[R_j\otimes R_{j'}]&=\Exp[\wt Q_j\otimes \wt Q_{j'}].
\label{eq:RR_correlation}\\
\Exp[Q_j\otimes Q_{j'}]&=\Exp[\wt R_j\otimes \wt R_{j'}].
\label{eq:QQ_correlation}
\end{align}
The last two come from noticing that they are limits of the same quantities
\begin{align*}
\tfrac1n\sum_{i=1}^n (r\n_j)_i\otimes (r\n_{j'})_i = 
\frac{(r\n_j)^*r\n_{j'}}n=
\frac{(U\n r\n_j)^*(U\n r\n_{j'})}n=
\frac{(\wt q\n_j)^*\wt q\n_{j'}}n=\tfrac1n\sum_{i=1}^n (\wt q\n_j)_i\otimes (\wt q\n_{j'})_i,\\
\tfrac1n\sum_{i=1}^n (\wt r\n_j)_i\otimes (\wt r\n_{j'})_i = 
\frac{(\wt r\n_j)^*\wt r\n_{j'}}n=
\frac{(U\n \wt r\n_j)^*(U\n \wt r\n_{j'})}n=
\frac{(q\n_j)^*q\n_{j'}}n=\tfrac1n\sum_{i=1}^n (q\n_j)_i\otimes (q\n_{j'})_i.
\end{align*}

\subsubsection{Proof of the base case \texorpdfstring{$t=1$}{t=1}.}

We will show all claims in \Cref{theorem:state evolution} for $t=1$ below. We split proving \eqref{eq:plkx} in two. Part 1 will show $(w\n,x_0\n,r_0\n,x_1\n)\cweak (W,X_0,R_0,X_1)$, and $(x_0\n,r_0\n,x_1\n)\plconv{k_x}
(X_0,R_0,X_1)$, part 2 will show the remaining claimed convergences in \eqref{eq:plkx} for $t=1$.

We will prove the claim in the following order: \eqref{eq:plkx} part 1, \eqref{eq:alpha_convergence}, $\wb\delta_1\succ0$, $\mu_1\succ 0$, \eqref{eq:N0R_uncorrelated}, \eqref{eq:plkz} with \eqref{eq:mu_def}, \eqref{eq:beta_convergence}, $\wb\gamma_1\succ0$,  $\tau_1\succ 0$, \eqref{eq:MQ_uncorrelated},
\eqref{eq:plkx} part 2 with \eqref{eq:tau_def}, \eqref{eq:NR_uncorrelated}, and \eqref{eq:N0N_uncorrelated}.

\paragraph{Proof of \eqref{eq:plkx} part 1:} We will first show $(w\n,x_0\n,r_0\n,x_1\n)\cweak (W,X_0,R_0,X_1)$, and then show that the empirical $k_x^{\rm th}$-moment of $(x_0\n,r_0\n,x_1\n)$ converge to the $k_x^{\rm th}$ moment of $(X_0,R_0,X_1)$, and invoke \Cref{proposition:plk equivalence} to show $(x_0\n,r_0\n,x_1\n)\plconv{k_x}
(X_0,R_0,X_1)$.

Since $\tau_0\succ0$, the conditional law of $R_0|(X_0,W)$
has a postive everywhere density. Thus
$\Prob(R_0\in J^G_{\bar\gamma_0,W}\mid X_0,W)=0$ almost surely.
Applying the continuous mapping theorem in \Cref{proposition:continuous mapping} to the augmented rows
with $(a,r,w,\gamma)\mapsto(w,a,r,G(r,\gamma,w))$ gives the asserted
joint weak convergence, with $X_1=G(R_0,\bar\gamma_0,W)$.

To handle the varying $\gamma\n_0$ in defining $x_1\n$, first choose a compact neighbourhood $K\subset\mc E_+$ of $\bar\gamma_0$.
Let $A_n=\{\gamma_0\n\in K\}$ and define
\(
\widetilde\gamma_0\n
=
\begin{cases}
\gamma_0\n, & \text{on } A_n,\\
\bar\gamma_0,   & \text{on } A_n^c,
\end{cases}~
\)
whereby $\widetilde\gamma_0\n\in K$ and
$\widetilde\gamma_0\n\cto\bar\gamma_0$.
For any of the empirical averages below, write $T_n$ for the average
computed with $\gamma_0\n$ and $\widetilde T_n$ for the same average
computed with $\widetilde\gamma_0\n$. Since $T_n=\widetilde T_n$
on $A_n$, for every $\varepsilon>0$ and $\ell\in\Real$,
\[
    \Prob(\abs{T_n-\ell}>\varepsilon)
    \le \Prob(A_n^c)
       +\Prob(\abs{\widetilde T_n-\ell}>\varepsilon).
\]
Since $K$ is a neighbourhood of $\bar\gamma_0$ and
$\gamma_0\n\cto\bar\gamma_0$, we have $\sum_n\Prob(A_n^c)<\infty$.
Therefore it suffices to prove $\widetilde T_n\cto\ell$.

Let $\mu_n$ be the empirical law of
$(x_0\n,r_0\n,w\n)$ and let $\mu$ be the law of
$(X_0,R_0,W)$. Adjoining $\wt\gamma_0\n$ to each row gives the law
$\mu_n\otimes\delta_{\wt\gamma_0\n}$, which converges weakly completely
to $\mu\otimes\delta_{\bar\gamma_0}$ by continuity of
$(\nu,\gamma)\mapsto\nu\otimes\delta_\gamma$.

To show the convergence of the $k_x^{\rm th}$ moment: by \Cref{assumption:growth rates}, for every $\gamma\in K$,
\[
\norm{(a,r,G(r,\gamma,w))}^{k_x}
\le C_K\bigl(1+\norm{a}^{k_x}+\norm{r}^{k_x}
+(1+\norm{w})^{k_w}\bigr).
\]
The empirical average of the right-hand side converges completely to
its expectation by the moment assumptions \cref{eq:x0 empirical moment,eq:r0 empirical moment,eq:assumption_W}. The left-hand side is
continuous at $(X_0,R_0,W,\bar\gamma_0)$ almost surely by the same
regularity assumption. Thus \Cref{corollary:empirical dominated convergence}
applied to the augmented empirical laws $\mu_n\otimes\delta_{\wt\gamma_0\n}$ gives
\begin{align}\label{eq:kx_moment_convergence_proof_step}
    \frac1n\sum_{i=1}^n
    \norm{\bigl((x_0\n)_i,(r_0\n)_i,(x_1\n)_i\bigr)}^{k_x}
    \cto \Exp\norm{(X_0,R_0,X_1)}^{k_x}.
\end{align}
This completes the proof of \eqref{eq:plkx} part 1.$\qedsymbol$

\paragraph{Proof of \eqref{eq:alpha_convergence} and  $\wb\delta_1\succ0$:}
Write $D_1=A_1+C_1$. By \Cref{lemma:consistency of onsager terms},
$\alpha_1\n+c_1\n\cto D_1$ for the bandwidth sequence $b_1\n$
provided by that lemma. Since we have  \Cref{assumption:shrinkage}, by \Cref{lemma:proximal derivative} we obtain
\[
    0\prec D_1\bar\gamma_0\prec\bar\gamma_0.
\]
In particular, $D_1$ is invertible. With
$\bar\eta_1=\bar\gamma_0(D_1^*)^{-1}$, we obtain
\[
\bar\delta_1=\bar\eta_1-\bar\gamma_0
=\bar\gamma_0\round{(D_1\bar\gamma_0)^{-1}
-\bar\gamma_0^{-1}}\bar\gamma_0\succ0.
\]
By invertibility of the limiting constant $D_1,$ the map $(\alpha_1,c_1,\gamma_0)\mapsto (\eta_1,\delta_1)$ is continuous which gives
$(\alpha_1\n+c_1\n,\eta_1\n,\delta_1\n)\cto(A_1+C_1,\bar\eta_1,\bar\delta_1)$.$\qedsymbol$

\paragraph{Proof of $\mu_1\succ0$:}
Since $(x_0\n,r_0\n,x_1\n)\plconv{k_x} (X_0,R_0,X_1)$ from \eqref{eq:plkx} part 1, and $(\gamma_0\n,\eta_1\n,\delta_1\n)\cto(\wb\gamma_0,\wb\eta_1,\wb\delta_1)$ from \eqref{eq:alpha_convergence}, by Slutsky's lemma in \Cref{proposition:slutsky},
\((x_0\n,r_0\n,x_1\n,\widetilde r_1\n)
\plconv{k_x}(X_0,\widetilde R_1,X_1,\tilde R_1),\) where \(
\widetilde R_1:=(X_1\bar\eta_1-R_0\bar\gamma_0)\bar\delta_1^{-1}.
\)
Since $k_x\ge2$, the empirical second moments converge completely.
By definition $\pi_1:=\Exp[X_0\otimes\widetilde R_1]$. Now notice that
\begin{align*}
    T_x=\begin{bmatrix}{-\bar\gamma_0}\\{\bar\eta_1}\end{bmatrix}\bar\delta_1^{-1},
    \qquad \widetilde R_1=\begin{bmatrix}R_0&X_1\end{bmatrix}T_x.
\end{align*}
By \eqref{eq:mu_def}, conditioning on $(X_0,W)$ gives
\begin{align*}
\mu_1
&=\Exp[(\widetilde R_1-X_0\Sigma_0^{-1}\pi_1)^{\otimes2}]\succeq \Exp[\Cov(\widetilde R_1\mid X_0,W)]
=T_x^*\Exp[\Cov((R_0,X_1)\mid X_0,W)]T_x
\succ0.
\end{align*}
The first inequality holds because \(X_0\Sigma_0^{-1}\pi_1\) is fixed given \((X_0,W)\). The strict inequality follows from the non-degeneracy assumption in \cref{eq:G non-degenerate} and 
the fact that $T_x$ is full column rank since $\wb\gamma_0,\wb\eta_1\succ0$. 

\paragraph{Proof of \eqref{eq:N0R_uncorrelated}:} Note that 
\begin{align*}
\Exp N_0\otimes \tilde{R}_1
&=((\Exp N_0\otimes X_1)\eta_1-(\Exp N_0\otimes R_0)\wb\gamma_0)\wb\delta_1\inv
=(\tau_{0} D_1\wb\eta_1 - \tau_0 \wb\gamma_0)\wb\delta_1\inv\\
&=\tau_0(D_1\wb\gamma_0(D_1^*)\inv-\wb\gamma_0)\wb\delta_1\inv = \tau_0(\gamma_0D_1^*(D_1^*)\inv-\wb\gamma_0)\wb\delta_1\inv = 0
\end{align*}
where in the penultimate step we used the self-adjointness of $D_1\wb\gamma_0.$
$\qedsymbol$

\paragraph{Proof of \eqref{eq:plkz} with \eqref{eq:mu_def}.} Conditioned on $\mc H_0\n$, the constraint on $U\n$ is
$U\n\mathbf{Q}_0\n=\mathbf{R}_0\n$. We want to use \Cref{lemma:haar empirical limit}\ref{part:U_tran} with $A\n=\mathbf{R}\n_0$ and $B\n=\mathbf{Q}\n_0$ and $v\n\gets \tilde{r}\n_1$. Then $p\n\gets B\n(A\n)\pinv v\n$ whose limit is $P\gets Z_0\Sigma_0\inv \pi_1$, since 
\begin{align*}
z\n_0 \round{\frac1n \sum_{i=1}^n (x_0\n)_i\otimes (x_0\n)_i}\inv \round{\frac1n \sum_{i=1}^n (x_0\n)_i\otimes (\tilde r_1\n)}_i\cweak Z_0\Sigma_0\inv \pi_1
\end{align*} due to \Cref{proposition:slutsky} and $(x_0\n,\tilde{r}_1\n)\plconv{k_x}(X_0,\tilde R_1)$.
Further, observe that for the choice of $v\n$, we get that the limit of the resulting $V_n$ is $V\gets \mu_1$ as defined in \eqref{eq:mu_def} since
\begin{align*}
V_n&:=\tfrac1n(v\n)^*
(I_n-A\n((A\n)^* A\n)\inv (A\n)^*) v\n\\
&=\frac1n\sum_{i=1}^n (\tilde r\n_1)_i\otimes  (\tilde r\n_1)_i - \frac1{n^2}\sum_{i,j=1}^n (\tilde r\n_1)_i\otimes (x\n_0)_i (\tfrac1n(A\n)^* A\n)\inv_{ij} (x_0\n)_j\otimes (\tilde r\n_1)_j\\
&\cto \Exp [\tilde R_1\otimes \tilde R_1] - \pi_1^*\Sigma_0\inv \pi_1
\end{align*}
Put together, \Cref{lemma:haar empirical limit}\ref{part:U_tran} with $h\n\gets(s\n,z_0\n)$ and $H\gets(S,Z_0)$, yields $q_1\n=(U\n)^\top\widetilde r_1\n$ must satisfy
\(
(s\n,z_0\n,q_1\n)\cweak(S,Z_0,Q_1)\) where $Q_1=Z_0\Sigma_0^{-1}\pi_1+M_1$ and
$M_1\sim\mc N(0,\mu_1)\independent (Z_0,S)$. 

Since \(p^{(n)}=z_0^{(n)}C_n\) with \(C_n\cto\Sigma_0^{-1}\pi_1\), its empirical \(k_z\)-th moment converges completely. The same holds coordinatewise for \(e^{(n)}:=q_1\n-p\n\) by the normalized-Gaussian representation. Applying \Cref{corollary:empirical dominated convergence} to \((p^{(n)},e^{(n)})\) with the envelope \(C(\|p\|^{k_z}+\sum_j|e_j|^{k_z})\) gives \(\frac1n\sum_i\|(q_1^{(n)})_i\|^{k_z}\cto\mathbb E\|Q_1\|^{k_z}\)

Finally, to show \((z_0\n,q_1\n,z_1\n) \plconv{k_z}(Z_0,Q_1,Z_1)\): since $\mu_1\succ0$, the conditional law of $Q_1|(Z_0,S)$
has a positive everywhere density. Thus $H$ is jointly continuous at
$(Q_1,\bar\delta_1,S)$ almost surely.
Applying the argument of \eqref{eq:kx_moment_convergence_proof_step} with \(H\) in place of \(G\), using the growth bound \eqref{eq:H_growth_rate} and the empirical moment convergence of \(z_0^{(n)},q_1^{(n)},s^{(n)}\), gives the claimed \(\mathrm{PL}(k_z)\) convergence of \((z_0^{(n)},q_1^{(n)},z_1^{(n)})\)
$\qedsymbol$

\paragraph{Proof of \eqref{eq:beta_convergence}:}
Since $\mu_1\succ 0$, $Q_1|Z_0,S$ has a positive everywhere density, we have $P(Q_1\in J^{H'}_{\delta,S}|S)$ the discontinuity points of $H'$.
Then using \Cref{corollary:empirical dominated convergence} with the domination condition $H'\delta\preceq \delta$, gives 
$\beta_1\n\cto B_1$. Finally \cref{eq:H non-degenerate} allows concluding that $0\prec  B_1\wb\delta_1 \prec \wb\delta_1$. 

\paragraph{Proof of  and  $\wb\gamma_1\succ0$:} The convergence to $B_1$ along with a continuous mapping theorem, since $B_1\wb\delta_1\succ 0$ ensuring continuity, for complete convergence, yields $\nu_1\n\cto\bar\nu_1:=\bar\delta_1 (B_1^*)^{-1}$. Furthermore, $B_1\wb\delta_1\prec\delta_1$ yields $\wb\gamma_1:=\wb\nu_1-\wb\delta_1\succ 0$.

\paragraph{Proof of $\tau_1\succ0$:}
Let $T_z=\binom{-\bar\delta_1}{\bar\nu_1}\bar\gamma_1^{-1}$, so that
$\widetilde Q_1=[Q_1\;\;Z_1]T_z$, and let
$\rho_1=\Exp[Z_0\otimes\widetilde Q_1]$. Conditioning on $(Z_0,S)$ gives
\[
 \tau_1=\Exp[(\widetilde Q_1-Z_0\Sigma_0^{-1}\rho_1)^{\otimes2}]
 \succeq T_z^*\Exp[\Cov((Q_1,Z_1)\mid Z_0,S)]T_z\succ0
\]
by \eqref{eq:H non-degenerate} and the full column rank of $T_z$.

\paragraph{Proof of \eqref{eq:MQ_uncorrelated}:} Stein's lemma gives
$\Exp[M_1\otimes Z_1]=\mu_1B_1$. Hence
\[
 \Exp[M_1\otimes\widetilde Q_1]=\Exp M_1\otimes (Z_1\wb\nu_1-Q_1\wb\delta_1)\wb\gamma_1\inv
 =\mu_1(B_1\bar\nu_1-\bar\delta_1)\bar\gamma_1\inv=0,
\]
where $B_1\bar\nu_1=\bar\delta_1$ follows from the symmetry of
$B_1\bar\delta_1$ and the definition of $\nu_1$. 

\paragraph{Proof of \eqref{eq:plkx} part 2 with \eqref{eq:tau_def}:} These are proved exactly like \eqref{eq:plkz} above using \Cref{lemma:haar empirical limit}\ref{part:U}.
Now, the revealed constraints are
$U\n\mathbf{Q}_{\nicefrac12}\n=\mathbf{R}_{\nicefrac12}\n$. We note a few identities using $(z_0\n,q_1\n)\plconv{k_z} (Z_0,Q_1)$. First $\frac1n(z_0\n)^* q_1\n = \frac1n(x_0\n)^*(\wt r_1\n)$. Both converge completely to yield $\Exp Z_0\otimes Q_1=\Exp X_0\otimes \wt R_1=\pi_1$.
\begin{align*}
\tfrac1n(\Q_\nicehalf\n)^*\Q\n_\nicehalf&=\frac1n\sum_{i=1}^n\begin{bmatrix}
(z_0\n)_i\otimes (z_0\n)_i & (z_0\n)_i\otimes (q_1\n-z_0\n\Sigma_0\inv\pi_t)_i\\
(q_1\n-z_0\n\Sigma_0\inv\pi_t)_i\otimes (z_0\n)_i & (q_1\n-z_0\n\Sigma_0\inv\pi_t)_i\otimes (q_1\n-z_0\n\Sigma_0\inv\pi_t)_i
\end{bmatrix}\\
&\cto
\Exp\begin{bmatrix}
Z_0\otimes Z_0 &
Z_0\otimes (Q_1-Z_0\Sigma_0\inv\pi_1)\\
(Q_1-Z_0\Sigma_0\inv\pi_1) \otimes Z_0&
(Q_1-Z_0\Sigma_0\inv\pi_1)\otimes (Q_1-Z_0\Sigma_0\inv\pi_1)
\end{bmatrix}\\
&=\begin{bmatrix}
    \Sigma_0\\ & \Exp Q_1\otimes Q_1 - \pi_1^*\Sigma_0\inv\pi_1
\end{bmatrix} = \diag(\Sigma_0, \mu_1)
\end{align*}
where the last line uses the definition of $\pi_1$ from \Cref{algo:se} and $\Exp Q_1^{\otimes2}-\pi_1^*\Sigma_0\pi_1=\Exp M_1^{\otimes2}=\mu_1$ as defined in \eqref{eq:mu_def}. Similarly using $(z_0\n,q_1\n,\wt q_1\n)\plconv{k_z} (Z_0,Q_1,\wt Q_1)$
\begin{align*}
\tfrac1n (\Q_\nicehalf\n)^*\wt q_1\n = \frac1n\sum_{i=1}^n
\begin{bmatrix}
(z_0\n)_i\otimes (\wt q_1\n)_i \\
(q_1\n - z_0\n\Sigma_0\inv\pi_1)_i\otimes (\wt q_1\n)_i
\end{bmatrix}
\cto
\begin{bmatrix}
\Exp Z_0 \otimes \wt Q_1\\
\Exp Q_1\otimes \wt Q_1 - \pi_1^*\Sigma_0\inv\rho_1
\end{bmatrix} = \begin{bmatrix}
\rho_1\\
0_E
\end{bmatrix}
\end{align*}
where we used $\Exp Q_1\otimes \wt Q_1=\Exp (M_1+Z_0\Sigma_0\inv\pi_1)\otimes \wt Q=\Exp M_1\otimes \wt Q_1+\pi_1^*\Sigma_0\inv \Exp Z_0\otimes\wt Q_1=\pi_1^*\Sigma_0\inv\rho_1$. The first term vanished because of \eqref{eq:MQ_uncorrelated} proved in the previous paragraph.
Together, we get
\begin{align*}
(\Q\n_\nicehalf)\pinv \wt q_1\n=(\tfrac1n(\Q\n_\nicehalf)^*\Q\n_\nicehalf)\inv \tfrac1n(\Q\n_\nicehalf)^*\wt q_1\n \cto \begin{bmatrix}
\Sigma_0\inv\rho_1\\
0_E
\end{bmatrix}
\end{align*}
We wish to use \Cref{lemma:haar empirical limit}\ref{part:U} with $A\n=\mathbf{R}\n_\nicehalf$ and $B\n=\mathbf{Q}\n_\nicehalf$ and $\wt v\n\gets \tilde{q}\n_1$. Then $\wt p\n\gets A\n(B\n)\pinv \wt v\n$ whose limit is $\wt P\gets X_0\Sigma_0\inv \rho_1$, since 
\begin{align*}
\R_\nicehalf\n (\Q_\nicehalf\n)\pinv \wt q_1\n=
\begin{bmatrix}x\n_0 & \wt r_1\n - x_0\n\Sigma_0\inv\pi_1\end{bmatrix}(\Q_\nicehalf\n)\pinv \wt q_1\n \cweak X_0\Sigma_0\inv \rho_1
\end{align*} due to \Cref{proposition:slutsky}.
Further, observe that for the choice of $\wt v\n$, we get that the limit of the resulting $\wt V_n$ in the statement of \Cref{lemma:haar empirical limit}\ref{part:U} is $\wt V\gets \tau_1$ as defined in \eqref{eq:tau_def} since
\begin{align*}
\wt V_n
&=\tfrac1n(\wt q_1\n)^*
  \big(I_n-B\n(B\n)\pinv\big)\wt q_1\n
=\tfrac1n(\wt q_1\n)^*\wt q_1\n
  -\big(\tfrac1n(B\n)^*\wt q_1\n\big)^*
   \big((B\n)\pinv\wt q_1\n\big)\\
&\cto \Exp[\wt Q_1^{\otimes2}]
-\begin{bmatrix}\rho_1^* & 0\end{bmatrix}
\begin{bmatrix}\Sigma_0\inv\rho_1\\0\end{bmatrix}
=\Exp[\wt Q_1^{\otimes2}]-\rho_1^*\Sigma_0\inv\rho_1
=\tau_1.
\end{align*}
Put together, \Cref{lemma:haar empirical limit}\ref{part:U} with $\wt h\n\gets(w\n,x_0\n,r_0\n,x_1\n)$ and $\wt H\gets(W,X_0,R_0,X_1)$, yields $r_1\n=U\n\widetilde q_1\n$ must satisfy
\(
(w\n,x_0\n,r_0\n,x_1\n,r_1\n)\cweak(W,X_0,R_0,X_1,R_1)\) where $R_1=X_0\Sigma_0^{-1}\rho_1+N_1$ and
$N_1\sim\mc N(0,\tau_1)\independent (W,X_0,R_0,X_1).$ 

One can show the $k_{x}^{\rm th}$ moments converge like we did in \eqref{eq:kx_moment_convergence_proof_step} to conclude the proof of part 2.
$\qedsymbol$

\paragraph{Proof of \eqref{eq:NR_uncorrelated}:}
Notice that
\begin{align*}
\frac1n\sum_{i=1}^n (\tilde r\n_1)_i\otimes (r_1\n)_i=\frac1n(r\n_1)^*r\n_1=\frac1n(U\n q_1\n)^*U\n \tilde q_1\n=\frac1n\sum_{i=1}^n (q_1\n)_i\otimes (\tilde q_1\n) \cto \Exp \tilde Q_1\otimes Q_1
\end{align*}
where the last convergence holds due to $(q_1\n,\tilde{q}_1\n)\plconv{k_z}(Q_1,\tilde{Q}_1)$. Similarly LHS $\cto \Exp \tilde{R}_1\otimes R_1$ since $(r_1\n,\tilde r_1\n)\plconv{k_x}(R_1,\tilde R_1)$ and $k_x,k_z\ge 2$. Now
\begin{align}
    \Exp N_1\otimes \tilde R_1 &= \Exp (R_1-X_0\Sigma_0\inv\rho_1) \otimes \tilde R_1 = \Exp R_1\otimes \tilde{R}_1 - \rho_1^*\Sigma_0\inv\Exp X_0\otimes \tilde{R}_1\nonumber\\
    &=\Exp \tilde Q_1\otimes Q_1 - \rho_1^*\Sigma_0\inv \pi_1 \nonumber\\
    &= (\Exp \tilde{Q}_1 \otimes Z_0) \Sigma_0\inv \pi_1 + (\Exp M_1\otimes \tilde{Q}_1)^* - \rho_1^*\Sigma_0\inv \pi_1 = 0\label{eq:N1R1_uncorrelated}
\end{align}
since $(\Exp \tilde{Q}_1 \otimes Z_0)=\rho_1^*$ and $\Exp M_1\otimes \tilde{Q}_1=0$ from \eqref{eq:MQ_uncorrelated}. 

\paragraph{Proof of \eqref{eq:N0N_uncorrelated}:} Observe that $N_0=R_0-X_0\Sigma_0\inv\rho_0$, whereby it is a measurable function of $(R_0,X_0)$. Since $N_1\independent (W,X_0,R_0,X_1)$ we have $\Exp N_0\otimes N_1=\Exp\Exp[(R_0-X_0\Sigma_0\inv\rho_0)\otimes N_1|(R_0,X_0)]=\Exp(R_0-X_0\Sigma_0\inv\rho_0)\otimes \Exp[N_1|(R_0,X_0)]=\Exp(R_0-X_0\Sigma_0\inv\rho_0)\otimes \Exp[N_1]=0$.

\subsubsection{Proof of the induction step for \texorpdfstring{$t>1$}{t>1}}

For the induction step, we assume that the claim of \Cref{theorem:state evolution} holds for $j=1...t-1$, and using this induction hypothesis we wish to show the claim for $j=t.$

We had followed this order for the base case: \eqref{eq:plkx} part 1, \eqref{eq:alpha_convergence}, $\wb\delta_{t}\succ0$, $\boxed{\mu_t\succ0}$, \eqref{eq:N0R_uncorrelated}, 
\boxed{\eqref{eq:plkz}}, 
\eqref{eq:beta_convergence}, $\wb\gamma_t\succ0$, 
$\boxed{\tau_t\succ0}$, 
\eqref{eq:MQ_uncorrelated}, 
\boxed{\eqref{eq:plkx}~ \text{part 2}},
\eqref{eq:NR_uncorrelated}, \eqref{eq:N0N_uncorrelated} 

We will only show the parts which are boxed, the remaining steps follow the same procedure as in the base case, so we omit their proof.

Note that the Proof of \eqref{eq:plkx} part 1 crucially relies on $\tau_{t-1}\succ0$ so that $R_{t-1}|(W,X_0)$ has a positive every density so that the distributional version of Stein's lemma, as stated in \Cref{lemma:stein}, holds.

\paragraph{Proof of $\mu_t\succ0$:} Here we use the notation $N_{<t}$ as short for $N_1,\ldots,N_{t-1}$ and similarly $N_{<t-1}$. We also write
\begin{align*}
    \mu_{t}=\begin{bmatrix}
        \mu_{t-1} & \mu_{<t,t}\\
        \mu_{t,<t} & \mu_{tt}
    \end{bmatrix}
\end{align*}
and similarly for $\tau_{t-1}$.
The induction hypothesis tells us $\mu_{t-1},\tau_{t-1}\succ0$. We only need to show that the Schur complement $\mu_{tt}-\mu_{t,<t}\mu_{t-1}\inv \mu_{<t,t}\succ 0$ and invoke the Schur test to prove $\mu_t\succ 0$.

Since $\round{\set{x_j\n,r_j\n}_{j=0}^{t-1},x_t\n}\plconv{k_x} \round{\set{X_j,R_j}_{j=0}^{t-1},X_t}$ from \eqref{eq:plkx} part 1, and $(\gamma_{t-1}\n,\eta_t\n,\delta_t\n)\cto(\wb\gamma_{t-1},\wb\eta_t,\wb\delta_t)$ from \eqref{eq:alpha_convergence}, by Slutsky's lemma in \Cref{proposition:slutsky},
$\round{\set{x_j\n,r_j\n,\wt r_j\n}_{j=0}^{t-1},x_t\n,\wt r_t\n}\plconv{k_x} \round{\set{X_j,R_j,\wt R_j}_{j=0}^{t-1},X_t,\wt R_t}$ where \(
\widetilde R_t:=(X_t\bar\eta_t-R_{t-1}\bar\gamma_{t-1})\bar\delta_t\inv
=(X_t,R_{t-1})T_x\) where $T_x=\begin{bmatrix}
    \wb\eta_t\\-\wb\gamma_{t-1}
\end{bmatrix}\wb\delta_t\inv$.

For $1\le j\le t$, let $Y_j=\wt R_j-X_0\Sigma_0\inv\pi_j$ and observe that
$\mu_t = \Exp Y_{\le t}\otimes Y_{\le t}.$
Write
$P=\mu_{t-1}\inv\mu_{<t,t}\in L(E^{t-1};E)$ as a block column so that $Y_{<t}P$ is the best linear predictor of $Y_t$ then note that
\begin{align}\label{eq:mu_schur}
\mu_{tt}-\mu_{t,<t}\mu_{t-1}\inv\mu_{<t,t}
=\Exp(Y_t-Y_{<t}P)^{\otimes2}.
\end{align}
which we will show is positive. Note that conditioning on $(W,X_0,N_0,N_{<t-1})$ we obtain the following inequality by dropping the covariance-of-conditional-expectaiton term in law of total variance
\begin{align}\nonumber
\Exp(Y_t-Y_{<t}P)^{\otimes2}&\succeq\Exp\Cov\sbrac{(Y_t-Y_{<t}P)|W,X_0,N_0,N_{<t-1}}+\Cov\Exp\sbrac{Y_t-Y_{<t}P|W,X_0,N_0,N_{<t-1}}\nonumber\\
&\succeq\Exp\Cov\sbrac{(Y_t-Y_{<t}P)|W,X_0,N_0,N_{<t-1}}\nonumber\\
&=\Exp\Cov\sbrac{Y_t|W,X_0,N_0,N_{<t-1}}=\Exp\Cov\sbrac{\wt R_t|W,X_0,N_0,N_{<t-1}}\label{eq:mu_conditioned_expr}
\end{align}
since $Y_{<t}$ and $X_0\Sigma_0\inv\pi_t$ are functions of $(W,X_0,N_0,N_{<t-1}).$ Next, we will apply Gebelin's inequality in \Cref{lemma:Gebelein inequality} under conditioning over $(W,X_0)$ with $A\gets N_{t-1}$ and $B\gets (N_0,N_{<t-1})$, and $f_{W,X_0}(A)\gets \wt R_t$ and $g_{W,X_0}(B)\gets \Exp [\wt R_t|W,X_0,N_0,N_{<t-1}]$ almost surely.\footnote{For almost every \((w,x_0)\), the map \(a\mapsto f_{w,x_0}(a)\) is a fixed square-integrable function. Treat $f_{W,X_0},g_{W,X_0}$ as random functions, whose randomness only depends on $(W,X_0)$} Since $N_0,N_{<t}\independent (W,X_0)$, we have
\begin{align*}
\Sigma_a\isqrtm\rho\Sigma_b\isqrtm
=\begin{bmatrix}
0 & \tau_{t-1,t-1}\isqrtm
    \tau_{t-1,<t-1}\tau_{t-2}\isqrtm
\end{bmatrix},
\end{align*}
where we used that $N_0$ is independent of $(N_1,\ldots,N_{t-1})$.
We get the following from a conditional version of the  Gebelin's inequality
\begin{align*}
\Cov[g_{W,X_0}(B)|W,X_0]\preceq \norm{\Sigma_a\isqrtm\rho\Sigma_b\isqrtm}_{\sf op}\Cov[f_{W,X_0}(A)|W,X_0]\qquad(W,X_0) \text{ a.s.}
\end{align*}
Furthermore, note that the conditional law of total variance yields, almost surely,
\begin{align*}
    \Exp[ \Cov(f_{W,X_0}(A)|W,X_0,B)|W,X_0]&=\Cov[f_{W,X_0}(A)|W,X_0]-\Cov(\Exp[f_{W,X_0}(A)|W,X_0,B]|W,X_0)\\
    &=\Cov[f_{W,X_0}(A)|W,X_0]-\Cov[g_{W,X_0}(B)|W,X_0]\\
    &\succeq (1-\norm{\Sigma_a\isqrtm\rho\Sigma_b\isqrtm}_{\sf op})\Cov[f_{W,X_0}(A)|W,X_0]
\end{align*}
Since $\tau_{t-1}\succ 0$, by the Schur test, we have $\tau_{t-1,t-1}\succ \tau_{t-1,<t-1}\tau_{t-2}\inv\tau_{<t-1,t-1}$ for $t\ge2$, and for $t=2$ the inequality is interpreted as $\tau_{11}\succ 0$, whereby 
\begin{align*}
\rho^*\rho=\tau_{t-1,t-1}\isqrtm \tau_{t,<t-1}\tau_{t-2}\inv\tau_{<t-1,t}\tau_{t-1,t-1}\isqrtm\prec I.    
\end{align*}
For $t=2$, $N_{<t-1}$ is empty whereby $B=N_0$ and $\rho=0$ and the above inequality is still true. Thus $\norm{\rho}_{\sf op}<1$ for all $t\ge 2.$
Consequently, taking an expectation over $(W,X_0)$, for some $c>0,$ we have
\begin{align*}
    \Exp[\Cov[\wt R_t|W,X_0,N_0,N_{<t-1}]\succ c\cdot \Exp \Cov[\wt R_t|W,X_0]
\end{align*}
Finally, using the same argument as the base case, we have
\begin{align*}
    \Exp\Cov[\wt R_t|W,X_0]=\Exp\Cov[(X_t,R_{t-1})T_x|W,X_0]=T_x^*\Exp\Cov(X_t,R_{t-1}|W,X_0)T_x \succ 0
\end{align*}
since $T_x$ is full column rank and $\Exp\Cov\Exp(X_t,R_{t-1}|W,X_0)\succ 0$ by the non-degeneracy from \Cref{assumption:non-degeneracy}. Together \eqref{eq:mu_schur}, \eqref{eq:mu_conditioned_expr}, and the last two equations above prove the claim $\mu_t\succ0$ via the Schur test.

\paragraph{Proof of \eqref{eq:plkz}:} At the begining of the $t^{\rm th}$ iteration, the sigma algebra $\mc H_{t-1}\n$ contains the event $U\n\Q\n_{t-1}=\R_{t-1}$. Further, by showing \eqref{eq:plkx} part 1 for $j=t$, we get 
\begin{align*}
\round{w\n,\set{x_j\n,\wt r_j\n,r_j\n}_{j=0}^{t-1},x\n_t,\wt r\n_t}&\cweak \round{W,\set{X_j,\wt R_j,R_j}_{j=0}^{t-1},X_t,\wt R_t}\\
\round{\set{x_j\n,\wt r_j\n,r_j\n}_{j=0}^{t-1},x\n_t,\wt r\n_t}&\plconv{k_x} \round{\set{X_j,\wt R_j,R_j}_{j=0}^{t-1},X_t,\wt R_t}
\end{align*}

We wish to apply \Cref{lemma:haar empirical limit} with $v\n=\wt r_t$ and $A\n=\R\n_{t-1}$ and $B\n=\Q\n_{t-1}$. To that end, we note a few identities for $1\le j',j<t,$
\begin{align*}
&\frac1n\sum_{i=1}^n (x_0\n)_i\otimes (\wt r_j-x_0\Sigma_0\inv\pi_j\n)_i\cto\Exp X_0\otimes \wt R_j-\pi_j=0\\
&\frac1n\sum_{i=1}^n (x_0\n)_i\otimes (r_j-x_0\Sigma_0\inv\rho_j\n)_i\cto\Exp X_0\otimes R_j-\rho_j\overset{\rm (a)}=0\\
&\frac1n\sum_{i=1}^n (\wt r_j-x_0\Sigma_0\inv\pi_j\n)_i\otimes (r_{j'}-x_0\Sigma_0\inv\rho_{j'}\n)_i \cto \Exp (\wt R_j-X_0\Sigma_0\inv\pi_j)\otimes (R_{j'}-X_0\Sigma_0\inv\rho_{j'})\\
&\qquad\qquad=\Exp \wt R_j\otimes R_{j'}-\pi_j^*\Sigma_0\inv\rho_{j'}=\Exp \wt R_j\otimes N_{j'}+\Exp\wt R_j\otimes X_0\Sigma_0\inv \rho_{j'}-\pi_j^*\Sigma_0\inv\rho_{j'}\overset{\rm (b)}=0
\end{align*}
where in (a) we have used the fact that $\Exp X_0\otimes R_j=\Exp Z_0 \otimes Q_j=\rho_j$, and in (b) we used \eqref{eq:NR_uncorrelated} from the induction hypothesis. Similarly,
\begin{align*}
&\frac1n\sum_{i=1}^n (\wt r_j-x_0\Sigma_0\inv\pi_j\n)_i\otimes (\wt r_{j'}-x_0\Sigma_0\inv\pi_{j'}\n)_i \cto 
\Exp \wt R_j\otimes \wt R_{j'}-\pi_j^*\Sigma_0\inv\pi_{j'}=\mu_{jj'}\\
&\frac1n\sum_{i=1}^n (r_j-x_0\Sigma_0\inv\rho_j\n)_i\otimes (r_{j'}-x_0\Sigma_0\inv\rho_{j'}\n)_i \cto 
=\Exp \wt Q_j\otimes \wt Q_{j'}-\rho_j^*\Sigma_0\inv\rho_{j'}=\tau_{jj'}
\end{align*}
Together, these yield that
\begin{align*}
\tfrac1n(\R\n_{t-1})^*(\R\n_{t-1})=
\frac1n\sum_{j=1}^n\cto \diag(\Sigma_0,\mu_t,\tau_t)
\end{align*}
Note that by the distributional Stein's lemma, we have, for $0\le j<t,$
\begin{align*}
\Exp N_j\otimes \wt R_t=\Exp N_j\otimes N_{t-1}((A_t+C_t)\wb\eta_t-\wb\gamma_{t-1})\wb\delta_t\inv=0.
\end{align*}
Further,
\begin{align*}
\tfrac1n(\R\n_{t-1})^*\wt r\n_{t}\cto
\Exp \begin{bmatrix}
X_0\otimes \wt R_t\\
((\wt R_j-X_0\Sigma_0\inv\pi_j)\otimes \wt R_t)_{j=1}^{t-1}\\
((R_j-X_0\Sigma_0\inv\rho_j)\otimes \wt R_t)_{j=1}^{t-1}\\
\end{bmatrix}=\begin{bmatrix}
    \pi_t\\
    (\mu_{jt})_{j=1}^{t-1}\\
    (\Exp N_j\otimes \wt R_t)_{j=1}^{t-1}
\end{bmatrix}=\begin{bmatrix}
    \pi_t\\
    (\mu_{jt})_{j=1}^{t-1}\\
    0_{E^{t-1}}
\end{bmatrix}
\end{align*}
Thus $p\n\gets \Q_{t-1}\R_{t-1}\pinv \wt r\n_t$ and $P\gets Z_0\Sigma_0\inv\pi_t+\Exp[M_t|M_1,M_2,\ldots,M_{t-1}]$
\begin{align*}
p\n\gets \Q_{t-1}\R_{t-1}\pinv \wt r\n_t&\cweak Z_0\Sigma_0\inv\pi_t+\round{(Q_t-Z_0\Sigma_0\inv\pi_j)_{j=1}^{t-1}}^*\mu_{t-1}\inv\round{\mu_{jt}}_{j=1}^{t-1}\\
&=Z_0\Sigma_0\inv\pi_t+((M_{jt})_{j=1}^{t-1})^*\mu_{t-1}\inv (\mu_{jt})_{j=1}^{t-1}=Z_0\Sigma_0\inv\pi_t+\Exp[M_t|M_1,M_2,\ldots,M_{t-1}]\\
\R_{t-1}\R_{t-1}\pinv \wt r\n_t&\cweak X_0\Sigma_0\inv\pi_t+\round{(\wt R_t-X_0\Sigma_0\inv\pi_j)_{j=1}^{t-1}}^*\mu_{t-1}\inv\round{\mu_{jt}}_{j=1}^{t-1}
\end{align*}
Next, we identify the limit $V$ for this choice of $v\n$.
\begin{align*}
&\frac1n\sum_{i=1}^n v\n_i\otimes v\n_i\cto\Exp \wt R_t\otimes \wt R_t\\
&\frac1n\sum_{i=1}^n v\n_i\otimes ( \R\n_{t-1}(\R\n_{t-1})\pinv v\n)_i\cto \sum_{j=1}^{t-1}\Exp \wt R_t \otimes (\wt R_j-X_0\Sigma_0\inv \pi_j)_{j=1}^{t-1}\mu_{t-1}\inv(\mu_{jt})_{j=1}^{t-1}\\
&\quad=\pi_t\Sigma_0\inv\pi_t + (\mu_{tj})_{j=1}^{t-1}\mu_{t-1}\inv(\mu_{jt})_{j=1}^{t-1}
\end{align*}
which together gives us $V\gets \mu_{tt}-\mu_{t,<t}\mu_{t-1}\inv\mu_{<t,t}\succ 0$ by the Schur test since $\mu_t\succ0$. Note that $V=\Cov(M_t|M_1,M_2,\ldots,M_{t-1})$ based on how $(M_1,\ldots,M_{t-1})$ are defined.

Invoking \Cref{lemma:haar empirical limit}\ref{part:U_tran} with $h\n\gets \round{s\n,\set{z\n_j,q\n_j,\wt q\n_j}_{j=0}^{t-1}}$ we get 
\begin{align*}
    \round{s\n,\set{z\n_j,q\n_j,\wt q\n_j}_{j=0}^{t-1},q\n_t}\cweak \round{S,\set{Z_j,Q_j,\wt Q_j}_{j=0}^{t-1},Q_t}
\end{align*}
where $P+G$ in the Lemma $\gets Q_t:=Z_0\Sigma_0\inv\pi_t+M_t$. The proof for the convergence of the $k_x^{\rm th}$ moment is same as in the base case using the growth bound \eqref{eq:G_growth_rate}. Similarly the proof for the 

\paragraph{Proof of \eqref{eq:plkx} part 2:} This is symmetrically similar to the proof of \eqref{eq:plkz} proved earlier.

\paragraph{Proof of $\tau_t\succ0$:} This is symmetric to the proof of $\mu_t\succ 0$.

\section{Conclusion}
In this paper we extended Vector Approximate Message Passing (VAMP) for rotationally invariant designs to M-estimation in generalized linear models (GLMs) with discontinuous proximal operators, which arise when the loss or the regularizer is non-convex. For proximal operators in $\SBVloc$, we derived a correction to the Onsager terms that accounts for the jumps, using a distributional Stein's lemma, and computed the correction from a kernel density estimate (KDE) of the denoiser input, so that the corrected algorithm requires no knowledge of the signal prior. We proved that the State Evolution (SE) is valid for  the corrected VAMP. Numerical experiments show that VAMP and its SE match closely for group sparse regression with an $\ell_0$-type penalty and for robust regression with the Cauchy loss.

Future work includes extending the analysis to Generalized-VAMP (G-VAMP) \cite{schniter2016vector} and multi-layer G-VAMP \cite{pandit2019asymptotics}.  On the technical side, deriving concentration bounds for the SE would determine the admissible decay rate of the KDE bandwidth, which we leave open (\Cref{remark:bn_deterministic}).

\section*{Acknowledgments}SL was supported by   UK Research and Innovation (UKRI) under the
UK government’s Horizon Europe funding guarantee [grant number EP/Y028333/1]. TA was supported by the IIT Bombay C-MInDS Predoctoral Fellowship when the research was conducted. PP was supported in part by the AI2050 program at Schmidt Sciences, DST INSPIRE Faculty fellowship, and grants from SBI Foundation Hub at IIT Bombay, 
Amazon IIT-B AI ML Initiative, and the Nilekani~Research~Fund.

\bibliographystyle{unsrt}
\bibliography{aux/ref}

\clearpage
\appendix

\section{ Proof of the Stationarity of VAMP Fixed Points}
\label{section:other proofs}
We restate the lemma and provide its proof. 
\propfixedpoint*
\begin{proof}[Proof of \Cref{proposition:fixed points}]
Assuming the algorithm has reached convergence. We have variables $x,r,z,q \in E^n$ and $\gamma,\delta,\eta,\nu\in\mc E_+$. Note that according to lines \ref{line:delta} and \ref{line:gamma} we have
\begin{align*}
    \eta=\nu=\gamma+\delta
\end{align*}
Furthermore note that by lines \ref{line:q} and \ref{line:r},
\begin{align*}
    U\tran r\gamma-q\delta = (z-U\tran x)\eta-(q\delta-U\tran r\gamma)\implies z=U\tran x
\end{align*}
Finally,
\begin{align*}
    r\gamma=U(z\nu-q\delta)=Uz\gamma+U(z-q)\delta=x\gamma+U(z-q)\delta \implies -(x-r)\gamma=U(z-q)\delta
\end{align*}

Note that we have
\begin{align*}
    0\in \partial F_2(z) + (z-q)\delta \quad\iff\quad (x-r)\gamma\in U\partial F_2(U\tran x)
\end{align*}
This implies that for all $x'$ we have
\begin{align*}
    F_2(U\tran x)\leq F_2(U\tran x') -\inner{(x-r)\gamma,x'-x}\quad \forall\, x'\in E^n 
\end{align*}
By definition of $\prox_{F_1}$ we have,
\begin{align*}
    F_1(x)+\tfrac12\norm{x-r}_\gamma^2 \leq F_1(x')+\tfrac12\norm{x'-r}_\gamma^2\qquad\forall\,x'\in E^n 
\end{align*}
Adding these inequalities we have, for all $x'\in E^n$,
\begin{align*}
    F_1(x)+F_2(U\tran x) &\leq F_1(x')+F_2(U\tran x')+\tfrac12\norm{x'-r}_\gamma^2-\inner{(x-r)\gamma,x'-x}-\tfrac12\norm{x-r}_\gamma^2,\\
    &= F_1(x')+F_2(U\tran x')+\tfrac12\norm{x-x'}_\gamma^2
\end{align*}
Let $F_3(x)=F_1(x)+F_2(U\tran x)$. 
Due to the above result we have that
\begin{align*}
    \frac{F_3(x')-F_3(x)}{\norm{x'-x}} \geq -\tfrac12\norm{x'-x}\norm{\gamma}_{\op}
\end{align*}
Observe that $0\in\hat\partial F_3(x)$ by definition of the Fr\'echet sub-differential. This proves the claim.
\end{proof}

\section{The \texorpdfstring{$\SBVloc$}{SBVloc} Space}
\label{section:sbvloc}

Let $E$ and $E'$ be finite-dimensional real Euclidean spaces, and let
$d:=\dim(E)$. We write $\mc L$ for Lebesgue measure on $E$ and $\mc H$
for $(d-1)$-dimensional Hausdorff measure on $E$. We equip $L(E;E')$
with its Hilbert--Schmidt inner product and norm. As in the main text,
operators act on the right, so $v\delta:=\delta(v)\in E'$ for $v\in E$
and $\delta\in L(E;E')$.

For $\zeta\in L^1_{\mathrm{loc}}(E;E')$ and an open set $U\subseteq E$,
define the variation of $\zeta$ on $U$ by
\[
V(\zeta,U)
:=
\sup\set{
\int_U
\inner{\zeta(x),\operatorname{div}\phi(x)}_{E'}
\dif\mc L(x)~:~
\phi\in C_c^1(U;L(E;E')),~
\sup_{x\in U}\norm{\phi(x)}_{\sf HS}\leq 1}
\]
Here $C_c^1$ denotes the space of compactly supported, continuously
differentiable test fields. The divergence is taken columnwise. In
orthonormal bases $(e_i)$ of $E$ and $(e'_j)$ of $E'$, writing
$\phi_{ij}(v):=\inner{e_i\phi(v),e'_j}_{E'}$, we have
\[
(\operatorname{div}\phi)_j=\sum_{i=1}^d\partial_i\phi_{ij}.
\]
We write $\zeta\in\BVloc(E;E')$ if $V(\zeta,U)<\infty$ for every
relatively compact open set $U\subseteq E$. Its distributional derivative
$\Dif\zeta$ is the unique locally finite $L(E;E')$-valued Radon
measure satisfying
\[
\int_E \inner{\zeta(v),\operatorname{div}\phi(v)}_{E'}\,\mc L(\dif v)
=
-\int_E \inner{\phi(v),\Dif\zeta(\dif v)}_{L(E;E')},
\qquad \phi\in C_c^1(E;L(E;E')).
\]
Its total variation satisfies $|\Dif\zeta|(U)=V(\zeta,U)$ for
every open set $U\subseteq E$.

A point $v\in E$ belongs to the approximate jump set $J_\zeta\subset E$ if there
exist distinct traces $\zeta^+(v),\zeta^-(v)\in E'$ and a unit normal
$\mathbf n_\zeta(v)\in E$ such that, for each choice of sign,
\[
\lim_{r\downarrow0}
\frac{1}{\mc L(B_r^\pm(v,\mathbf n_\zeta(v)))}
\int_{B_r^\pm(v,\mathbf n_\zeta(v))}
\norm{\zeta(u)-\zeta^\pm(v)}\,\mc L(\dif u)=0,
\]
where $B_r^\pm(v,n):=\{u\in B_r(v):\ \pm\inner{u-v,n}>0\}$.
Write $[\zeta](v):=\zeta^+(v)-\zeta^-(v)$. For $n,h\in E$ and $a\in E'$,
our tensor convention gives $h(n\otimes a)=\inner{h,n}a$. Reversing
$\mathbf n_\zeta$ exchanges the two traces, so
$\mathbf n_\zeta\otimes[\zeta]$ is independent of the orientation.

For $\zeta\in\BVloc(E;E')$, the BV structure theorem \cite[Theorems~3.78 and~3.83 and Proposition~3.92]{ambrosio2000functions} gives the mutually
singular decomposition
\[
\Dif\zeta
=\Dif^{\sf a}\zeta+\Dif^{\sf j}\zeta+\Dif^{\sf c}\zeta.
\]
The density of the absolutely continuous part $\Dif^{\sf a}\zeta$
with respect to $\mc L$ is the approximate differential $\zeta'$.
The jump part $\Dif^{\sf j}\zeta$ is concentrated on $J_\zeta$,
which is countably $(d-1)$-rectifiable. The Cantor part satisfies
$|\Dif^{\sf c}\zeta|\perp\mc L$ and
$|\Dif^{\sf c}\zeta|(A)=0$ for every Borel set $A$ that is
$\sigma$-finite with respect to $\mc H$.

We define
\[
\SBVloc(E;E')
:=\{\zeta\in\BVloc(E;E'):\ \Dif^{\sf c}\zeta=0\}.
\]
Thus, for every $\zeta\in\SBVloc(E;E')$ and every relatively compact
Borel set $A\subseteq E$,
\[
\Dif\zeta(A)
=
\underbrace{\int_A \zeta'(v)\,\mc L(\dif v)}_{\Dif^{\sf a}\zeta(A)}
+
\underbrace{\int_{A\cap J_\zeta}
\mathbf n_\zeta(v)\otimes[\zeta](v)\,\mc H(\dif v)}_{\Dif^{\sf j}\zeta(A)}.
\]

\section{Proof of the Distributional Stein's Lemma}
\label{section:proof stein}

\steinlemma*
\begin{proof}[Proof of \Cref{lemma:stein}]
Let $J_\zeta$ denote the approximate jump set of $\zeta$.
We allow $\zeta$ to be an $\SBVloc(E;E)$-valued random variable, with a jointly measurable version,
and assume $(M,N)\independent(\zeta,X)$. Thus, conditional on
$\mc F:=\sigma(\zeta,X)$, both $\zeta$ and $X$ are fixed,
while $(M,N)|\mc F$ with $N\sim\mc N(\zero_E,\tau)$. 

We first show the case of $\Cov(N,\zeta(R))$ and then use Gaussian regression to transfer the result to $\Cov(M,\zeta(R))$
\begin{align*}
    \Cov(N,\zeta(R))
    &\overset{\rm (a)}=
    \Exp_{X,\zeta}\Exp\left[N\otimes\zeta(X+N)
     | \mc F\right]=    \Exp_{X,\zeta}\left[\int_E v\otimes\zeta(X+v)
    \varphi_\tau(v)\dif v\right]\\
    &\overset{\rm (b)}=
    -\tau\Exp_{X,\zeta}\left[\int_E \mathrm{D}\varphi_\tau(v)
    \otimes\zeta(X+v)\dif v\right]\\
    &\overset{\rm (c)}=
    \tau\Exp_{X,\zeta}\left[\int_E
    \varphi_\tau(u-X)\mathrm{D}\zeta(\dif u)\right]\\
    &\overset{\rm (d)}=
    \tau\Exp_{X,\zeta}\left[\int_E
    \varphi_\tau(u-X)\zeta'(u)\dif u\right]
    +\tau\Exp_{X,\zeta}\left[\int_{J_\zeta}
    \varphi_\tau(u-X)\mathbf n_\zeta(u)\otimes[\zeta](u)
    \mc H(\dif u)\right]\\
    &\overset{\rm (e)}=
    \tau\Exp_{X,\zeta}\mc A(\zeta,X,\tau)+\tau\Exp_{X,\zeta}\mc C(\zeta,X,\tau).
\end{align*}
Here, (a) follows from $\Exp[ N| \mc F]=\zero_E$, (b) uses
$\mathrm{D}\varphi_\tau(v)=-v\tau^{-1} \varphi_\tau(v)$;
(c) uses the substitution $u=X+v$ and distributional integration
by parts, justified below; (d) uses the $\SBVloc$ decomposition
of $\mathrm{D}\zeta$, which holds for every realization; and
(e) follows by definition
of $\mc C$.

To justify (c), first replace $\varphi_\tau(u-X)$ by
$\chi_m(u-X)\varphi_\tau(u-X)$, where
$\chi_m\in C_c^\infty(E)$,
$0\leq\chi_m\uparrow1$, and $\norm{\mathrm{D}\chi_m}_\infty\leq c/m$ (See \Cref{remark:smooth cutoff} for a construction).
These are admissible test functions for distributional integration
by parts. Since a proximal map agrees almost everywhere with the
gradient of a convex function, $\mathrm{D}\zeta \gamma$ is a positive-semidefinite
matrix-valued measure. Taking traces in the cutoff identity and
applying monotone convergence gives
\begin{align*}
    \Exp\int_E\varphi_\tau(u-X)|\mathrm{D}\zeta|(\dif u)
    \leq \norm{\tau^{-1}}_{\mathrm{op}}\,
    \Exp\big[\norm{N}\norm{\zeta(R)}\big]<\infty.
\end{align*}
The last inequality follows by Cauchy--Schwarz.
The cutoff derivative contributes a term bounded in expectation by
$(c/m)\Exp\norm{\zeta(R)}$, which tends to zero.
Thus the cutoff limit is valid, and both terms in (d) are integrable.

For $\Cov(M,\zeta(R))$, since \((M,N)\) is jointly Gaussian and independent of \(\mc F\), Gaussian regression gives \(\Exp[M-\Exp M| N,\mc F]=N\tau^{-1}\Cov(N,M)\). As \(\zeta(R)=\zeta(X+N)\) is measurable with respect to \(\sigma(N)\vee\mc F\), conditioning yields
\[
\begin{aligned}
\Cov(M,\zeta(R))
&=\Exp\Exp[(M-\Exp M)\otimes\zeta(R)|N,\mc F]= \Cov(M,N)\tau^{-1}\Cov(N,\zeta(R))\\
&=\Cov(M,N)\,\Exp\!\left[\mc A(\zeta,X,\tau)+\mc C(\zeta,X,\tau)\right],
\end{aligned}
\]
where the second equality follows from (e).
\end{proof}

\section{Proof of Consistency of the Corrected Onsager Term}\label{app:proof_consistency_onsager}
We restate the lemma and provide its proof. For Lipschitz denoisers $G$ with non-Lipschitz $G'$ was considered by a similar smoothing argument in \cite{berthier2020state}. 
\lemmaconsistencyonsager*
\begin{proof}[Proof of \Cref{lemma:consistency of onsager terms}]
Fix $t\ge 0 $ and suppress its occurrence in the notation.
Let
\[
  \mu_{\gamma,w}:=\Dif_vG(\cdot,\gamma,w).
\]

\paragraph{Gaussian integrability.}
Proximality implies that
$v\mapsto G(v,\gamma,w)\gamma$ is almost everywhere the
gradient of a finite convex function. Consequently,
$\mu_{\gamma,w}\gamma$ is a symmetric positive-semidefinite
matrix-valued measure. Thus, for $T\in\mc E_+$,
\begin{align}
  \int_E\varphi_T(v-u)\,|\mu_{\gamma,w}|(\dif v)
  &\le
  \|\gamma^{-1}\|_{\mathrm{op}}
  \int_E\varphi_T(v-u)\,
    \trace\bigl(\mu_{\gamma,w}(\dif v)\gamma\bigr)
  \nonumber\\
  &=
  \|\gamma^{-1}\|_{\mathrm{op}}\,
  \Exp_Z\left\langle
    ZT^{-1},G(u+Z,\gamma,w)\gamma
  \right\rangle \qquad Z\sim\mc N(0_E,T),
  \nonumber\\
  &\le
  C_{\gamma,T}\bigl(1+\|u\|+(1+\|w\|)^{k_w/k_x}\bigr).
  \label{eq:kde_gaussian_derivative_bound}
\end{align}
Here $|\mu_{\gamma,w}|$ denotes total variation, using the
Frobenius norm on matrices. The integration-by-parts identity in the second step
is justified first with smooth compact cutoffs.
The growth bound makes the boundary terms vanish; positivity
of the trace measure then permits removal of the cutoffs.
The last inequality follows from the growth bound \eqref{eq:G_growth_rate} with $K=\{\gamma\}$ and the
finite Gaussian moments of $Z$, where $C_{\gamma, T}:= \|\gamma^{-1}\|_{\op} \|\gamma\|_{\op}\|T^{-1}\|_{\op} C_{\{\gamma\}, 1} \big(\sqrt{\trace T } + \trace T\big)$, which depends only on $\gamma$ and $T$, not on $u$ or $w$.

\paragraph{Fixed-bandwidth consistency.}
For $b>0$ and $\gamma\in \mc{E}_+$,  let $B_b(\gamma):=b^2\gamma^{-1}$ and define $\mc{D}_b: E\times \mc{E}_+\times E_w\to L(E;E)$ by 
\[
  \mc D_b(r,\gamma,w)
  :=
  \int_E
  \varphi_{B_b(\gamma)}(v-r)\,
  \mu_{\gamma,w}(\dif v).
\]
Since $g_i^{(n)}(\cdot,\gamma^{(n)})=G(\cdot,\gamma^{(n)},w_i^{(n)})$, the SBV decomposition applied to each summand of $\alpha_{t+1}^{(n)}(b)+c_{t+1}^{(n)}(b)$ gives,
\[
  S_n(b):=
  \alpha_{t+1}^{(n)}(b)+c_{t+1}^{(n)}(b)
  =
  \frac1n\sum_{i=1}^n
  \mc D_b(r_i^{(n)},\gamma^{(n)},w_i^{(n)}).
\]
Gaussian integration by parts, justified with the same smooth compact cutoffs as in \eqref{eq:kde_gaussian_derivative_bound},   gives
\begin{align}
  \mc D_b(r,\gamma,w)
  =
  \int_E
  \varphi_{B_b(\gamma)}(v-r)\,
  \bigl((v-r)B_b(\gamma)^{-1}\bigr)
  \otimes G(v,\gamma,w)\,\mc L(\dif v).
  \label{eq:kde_score_representation}
\end{align}
Here, the growth   bound makes the boundary terms vanish and gives dominated convergence for the integral against $\mc L$, while \eqref{eq:kde_gaussian_derivative_bound} with $T=B_b(\gamma)$ gives dominated convergence for the integral against $\mu_{\gamma,w}$. Substituting $v=r+b\Xi\gamma^{-1/2}$ with $\Xi\sim\mc N(0_E,I_E)$ in \eqref{eq:kde_score_representation} gives
\begin{align}
    \label{eq:simplified_Db}\mc D_b(r,\gamma,w)=b^{-1}\gamma^{1/2}\,\mathbb E\big[\Xi\otimes G\big(r+b\Xi\gamma^{-1/2},\gamma,w\big)\big].
\end{align}
This form gives two properties of $\mc D_b$.
First, $\mc D_b$ is jointly continuous. Indeed, if $(r_j,\gamma_j,w_j)\to(r,\gamma,w)$, the integrand in \eqref{eq:simplified_Db} converges for almost every $\Xi$, because $r+b\Xi\gamma^{-1/2}$ has a density and so almost surely avoids the discontinuity set of $G$. The growth bound  supplies an integrable envelope, so dominated convergence applies. Second, applying \eqref{eq:G_growth_rate} in \eqref{eq:simplified_Db}  gives, for every compact $K\subset\mc E_+$,
\begin{align}
  \|\mc D_b(r,\gamma,w)\|
  \le C_{b,K}\bigl(1+\|r\|+(1+\|w\|)^{k_w/k_x}\bigr),
  \qquad \gamma\in K,\label{eq:bound_Db}
\end{align}
where $C_{b,K}:=b^{-1}\big(\sup_{\gamma\in K}\|\gamma^{1/2}\|_{\rm op}\big)C_{K,1}\big(\sqrt{\dim E}+b \sup_{\gamma\in K}\|\gamma^{-1/2}\|_{\rm op}\dim E\big)$.

The empirical weak convergence and moment convergence therefore give
\[
  S_n(b)\cto
  S(b):=\Exp\mc D_b(R,\bar\gamma,W)
  \qquad\text{for every fixed }b>0.
\]
Indeed, one first restricts $\gamma^{(n)}$ to a compact
neighbourhood $K$ of $\bar\gamma$, whose complement $\{\gamma^{(n)}\notin K\}$ has summable
probabilities due to \eqref{eq:hypothesis_onsager_gamma}. Bounded continuous truncations are handled by
weak empirical convergence \eqref{eq:hypothesis_onsager_weak} and parameter convergence \eqref{eq:hypothesis_onsager_gamma}.
The moment assumptions \eqref{eq:assumption_W} and \eqref{eq:hypothesis_onsager_r}, with $k_x>1$, make the truncation
error small: by the bound on $\|\mc{D}_b(r, \gamma, w)\|$ in \eqref{eq:bound_Db}, the rows removed by the truncation contribute at most $C_{b,K}$ times the average of $1+\|r_i\|+(1+\|w_i\|)^{k_w/k_x}$ over those rows, and this average vanishes as the truncation level grows.

\emph{The  limit of $S(b)$ as $b\downarrow 0$.} Having proved $S_n(b)\cto S(b)$, we now prove $\lim_{b\downarrow 0}S(b)=A_{t+1}(0^+)+C_{t+1}(0^+)$. 
Write $\mu_W:=\mu_{\bar\gamma,W}$ and $m:=X_0\Sigma_0^{-1}\rho$.
Conditioning on $(X_0,W)$ and using Gaussian convolution yields
\begin{align}
  S(b)
  &=
  \Exp\int_E
  \varphi_{\tau+b^2\bar\gamma^{-1}}(v-m)\,
  \mu_W(\dif v).
  \label{eq:kde_population_convolution}
\end{align}
Absolute integrability and the interchange of integrals follow
from \eqref{eq:kde_gaussian_derivative_bound}.

Set $T:=\tau+\bar\gamma^{-1}$. For $0<b\le1$,
\[
  \varphi_{\tau+b^2\bar\gamma^{-1}}(z)
  \le
  \sqrt{\frac{\det T}{\det\tau}}\,
  \varphi_T(z).
\]
Consequently, the integrated random quantities in
\eqref{eq:kde_population_convolution} are dominated in norm by
\[
  \sqrt{\frac{\det T}{\det\tau}}\,
  \int_E\varphi_T(v-m)\,|\mu_W|(\dif v)
  \le C\bigl(1+\|m\|+(1+\|W\|)^{k_w/k_x}\bigr),
\]
where $C:=\sqrt{\det T/ \det \tau} \cdot C_{\bar{\gamma}, T}$ with $C_{\bar\gamma,T}$ being the constant from \eqref{eq:kde_gaussian_derivative_bound}, evaluated at $\gamma=\bar\gamma$ and $T=\tau+\bar\gamma^{-1}$. 
This bound is integrable: $m=\Exp[R\mid X_0,W]$, so
$\Exp\|m\|\le\Exp\|R\|<\infty$, and
$$\Exp (1+\|W\|)^{k_w/k_x} \le \Exp(1+\|W\|)^{k_w}<\infty$$ by the moment assumption \eqref{eq:assumption_W} since $k_x\ge2>1$.

For each fixed $(m,w)$, the same Gaussian domination establishes
continuity of the derivative-measure integral as its covariance
converges to $\tau$. Dominated convergence for the outer
expectation therefore gives
\begin{align*}
  \lim_{b\downarrow0}S(b)
  &=
  \Exp\int_E
  \varphi_\tau(v-m)\,\mu_W(\dif v)\\
  &=
  \Exp G'(R,\bar\gamma,W)
  +
  \Exp\mc C\bigl(G(\cdot,\bar\gamma,W),m,\tau\bigr)\\
  &=A_{t+1}(0^+)+C_{t+1}(0^+).
\end{align*}
The two terms are separately finite by
\eqref{eq:kde_gaussian_derivative_bound} and the SBV decomposition.

\emph{Selection of the bandwidth.} Let $S:=A_{t+1}(0^+)+C_{t+1}(0^+)$, so that $S(b) \to S$ as $b\downarrow 0$ by the previous step. By fixed-bandwidth complete convergence, $\sum_{n=1}^\infty\mathbb P(\|S_n(b)-S(b)\|>\delta)<\infty$ for every fixed $b,\delta>0$. Hence, 
for each integer $j\ge1$, we can  choose strictly increasing deterministic integers
$N_1< N_2<\dots$ such that
\[
  \sum_{n\ge N_j}
  \Prob\left(
    \|S_n(1/j)-S(1/j)\|>1/j
  \right)<2^{-j}.
\]
Set  $b^{(n)}=1$
for $n<N_1$ and $b^{(n)}=1/j$ for $N_j\le n<N_{j+1}$. Then $b^{(n)}\downarrow0$.

Fix $\varepsilon>0$.
Since $S(1/j)\to S$, there is $j_0$ such that
\[
  \|S(1/j)-S\|+1/j<\varepsilon
  \qquad\text{for all }j\ge j_0.
\]
Let $n\ge N_{j_0}$. Then $N_j\le n<N_{j+1}$ for some $j\ge j_0$, so $b^{(n)}=1/j$. By the triangle inequality, $\|S_n(b^{(n)})-S\|>\varepsilon$ implies
$$\|S_n(1/j)-S(1/j)\|>\varepsilon-\|S(1/j)-S\|>1/j.$$
Bounding each of the terms with $n<N_{j_0}$ by 1 gives
\begin{align*}
  \sum_{n=1}^{\infty}
  \Prob\bigl(\|S_n(b^{(n)})-S\|>\varepsilon\bigr)
  &\le
  N_{j_0}
  +
  \sum_{j\ge j_0}\sum_{n=N_j}^{N_{j+1}-1}
  \Prob\left(
    \|S_n(1/j)-S(1/j)\|>1/j
  \right)\\
  &\le N_{j_0}+\sum_{j\ge j_0}2^{-j}<\infty.
\end{align*}
This proves the claimed complete convergence.
\end{proof}

\section{Some Useful  Results}
\subsection{from probability theory}
\label{section:useful probability}

For proving some integrability claims for functions growing slow enough, we will need to assume the existence of some smooth cutoff functions. We provide a construction below, although any smooth cutoff suffice for all the arguments used.
\begin{remark}[Smooth cutoff functions]\label{remark:smooth cutoff}
Define the smooth function
\[
\eta(t)=
\begin{cases}
e^{-1/t},&t>0,\\
0,&t\le0,
\end{cases}
\qquad
\theta(s)=
\frac{\eta(4-s)}
{\eta(4-s)+\eta(s-1)}.
\]
The denominator is everywhere positive. Moreover, \(\theta\in C^\infty(\mathbb M)\) is nonincreasing, equals \(1\) on \((-\infty,1]\), and equals \(0\) on \([4,\infty)\).
Now take
\begin{align}\label{def:smooth cutoff}
\chi_n(r)=\theta\!\left(\frac{\|r\|^2}{n^2}\right),
\quad n\ge1.
\end{align}
Then \(\chi_n\in C_c^\infty(E)\), with \(0\le\chi_n\le1\).
 \(\chi_n=1\) on \(\{\|r\|\le n\}\) and \(\chi_n=0\) on \(\{\|r\|\ge2n\}\).
  Since \(\theta\) is nonincreasing, \(\chi_n(r)\uparrow1\) for every \(r\).

Finally,
\[
\Dif \chi_n(r)
=\frac{2r}{n^2}\,
\theta'\!\left(\frac{\|r\|^2}{n^2}\right).
\]
This vanishes unless \(n<\|r\|<2n\), so
\[
\|\Dif \chi_n\|_\infty
\le \frac{4\|\theta'\|_\infty}{n}.
\]
Here \(\|\theta'\|_\infty<\infty\), because \(\theta'\) is smooth and supported in \([1,4]\).
    
\end{remark}

\propPLk*
\begin{proof}[Proof of \Cref{proposition:plk equivalence}]
Let $P_n$ be the expectation operator of the empirical distribution from the rows of $x^{(n)}$ and $P$ be the same for the distribution of $X$.
Let $\psi\in\text{PL}(k;E;E')$. Its growth satisfies
\[
\|\psi(v)\|\le Cg(v),
\qquad g(v):=1+\|v\|^k.
\]
Choose a Lipschitz cutoff \(\chi_M:E'\to[0,1]\) equal to \(1\) on \(\{\|v\|\le M\}\) and \(0\) on \(\{\|v\|\ge2M\}\). Both \(\chi_M\psi\) and \(\chi_Mg\) are bounded Lipschitz. Consequently, for each fixed \(M\),
\[
P_n(\chi_M\psi)\cto P(\chi_M\psi),
\qquad
P_n(\chi_Mg)\cto P(\chi_Mg).
\]
Together with \(P_ng\cto Pg\), this gives
\[
T_{n,M}:=P_n\big((1-\chi_M)g\big)
\cto 
P\big((1-\chi_M)g\big)=:t_M.
\]
Moreover, \(t_M\to0\) by dominated convergence, since \(Pg<\infty\). Now
\[
\begin{aligned}
\|P_n\psi-P\psi\|
&\le
\|P_n(\chi_M\psi)-P(\chi_M\psi)\|
+C(T_{n,M}+t_M)\\
&\le
\|P_n(\chi_M\psi)-P(\chi_M\psi)\|
+C|T_{n,M}-t_M|+2Ct_M.
\end{aligned}
\]
Given \(\varepsilon>0\), fix \(M\) sufficiently large so that \(2Ct_M\le\varepsilon/2\). Then
\[
\begin{aligned}
\sum_{n=1}^{\infty}
\mathbb P\!\left(\|P_n\psi-P\psi\|>\varepsilon\right)
&\le
\sum_{n=1}^{\infty}
\mathbb P\!\left(
\|P_n(\chi_M\psi)-P(\chi_M\psi)\|>\frac{\varepsilon}{4}
\right)+
\sum_{n=1}^{\infty}
\mathbb P\!\left(
|T_{n,M}-t_M|>\frac{\varepsilon}{4C}
\right)
<\infty.
\end{aligned}
\]
Thus \(P_n\psi\cto P\psi\), proving the claim.
\end{proof}

\begin{proposition}[Dominated convergence]\label{proposition:moments durrett}
Suppose $X_n\dto X$ in a finite-dimensional Euclidean space $E$.
Let $g:E\to[0,\infty)$ be continuous and let $h:E\to\Real$ be measurable,
with $\abs{h(u)}\le g(u)$ for all $u\in E$. Suppose $\Prob(X\in D_h)=0$,
where $D_h$ is the set of discontinuity points of $h$.
If either of the following conditions holds, then
$\Exp h(X_n)\to\Exp h(X)$.
\begin{enumerate}[label=\textup{(\roman*)}]
\item $\sup_n\Exp g(X_n)<\infty$ and,  $\forall\,\varepsilon>0$,
 $\exists\,R<\infty$ such that
$\abs{h(u)}\le\varepsilon g(u)$ whenever $\norm{u}\ge R$.
\item\label{proposition:moments durrett:converging g} $\Exp g(X_n)\to\Exp g(X)<\infty$.
\end{enumerate}
\end{proposition}
\begin{proof}
We need the following result.

\begin{proposition}[\!\texorpdfstring{\cite[Theorem 1.6.8]{durrett2019probability}}{Durrett}]\label{proposition: durrett domination}
Suppose $X_n\xrightarrow{a.s.} X$, and $g,h\in C$  satisfying
\begin{enumerate}[label=\textup{(\roman*)}]
\item $\norm{g(x)}\to\infty$ as $\norm{x}\to\infty$,
\item $\frac{\norm{h(x)}}{\norm{g(x)}}\to 0$ as $\norm{x}\to\infty$, and
\item $\sup_n \Exp \norm{g(X_n)}\le K<\infty$ for all $n$
\end{enumerate}
Then $\Exp h(X_n)\to \Exp h(X).$
\end{proposition}

By the Skorokhod representation theorem, we may work with copies for which
$X_n\to X$ almost surely. Then $h(X_n)\to h(X)$ and $g(X_n)\to g(X)$
almost surely. Fatou's lemma gives $\Exp g(X)\le\liminf_n\Exp g(X_n)$.

Under (i), for every $\varepsilon>0$ there is a constant $C_\varepsilon$
such that $\abs{h}\le C_\varepsilon+\varepsilon g$ everywhere: the
little-$o$ condition gives the bound outside a ball, and $g$ is bounded
on the ball. Applying Fatou's lemma to the two nonnegative functions
$C_\varepsilon+\varepsilon g\pm h$ gives
\[
    \limsup_n \abs{\Exp h(X_n)-\Exp h(X)}
    \le \varepsilon\bigl(\limsup_n\Exp g(X_n)-\Exp g(X)\bigr).
\]
The quantity in parentheses is finite, so letting $\varepsilon\downarrow0$
proves (i).

Under (ii), $\abs{h}\le g$ gives the same inequality with
$\varepsilon=1$ and $C_\varepsilon=0$. Its right-hand side is zero.
\end{proof}

\corollaryEmpiricalConvergence*
\begin{proof}[Proof of \Cref{corollary:empirical dominated convergence}]
Consider a metric $d$ that metrizes weak convergence and define
\[
    d_g(\nu,\mu)=d(\nu,\mu)+\abs{\nu g-\mu g}
\]
on the probability measures with finite $g$-moment, i.e., $\mu g<\infty$.
By \Cref{proposition:moments durrett}\ref{proposition:moments durrett:converging g}, the map $\nu\mapsto\nu f$
is continuous at $\mu$ in $d_g$. Weak complete convergence gives
$d(\mu_n,\mu)\cto0$ by the weak-neighbourhood characterization.
Together with $\mu_n g\cto\mu g$, this gives $d_g(\mu_n,\mu)\cto0$.
For each $\varepsilon>0$, continuity therefore gives $\delta>0$ such that
\[
    \Prob\bigl(\abs{\mu_n f-\mu f}>\varepsilon\bigr)
    \le \Prob\bigl(d_g(\mu_n,\mu)>\delta\bigr).
\]
The RHS, and hence the LHS, is summable in $n$, which concludes the proof.
\end{proof}

\propCMT*
\begin{proof}[Proof of \Cref{proposition:continuous mapping}]
Let $\mu_n$ be the empirical law of $x^{(n)}$ and let $\mu$ be the law
of $X$. Fix $\psi\in C_b(E_x\times E_y,\Real)$ and set
$f(u)=\psi(u,h(u))$. Then $f$ is bounded and $\mu$-almost everywhere
continuous, since $D_f\subseteq D_h$. By the Portmanteau lemma,
$\nu\mapsto\nu f$ is continuous at $\mu$ in the weak topology.

Thus, for every $\varepsilon>0$, there exists $m$ large enough such that the tests
$g_1,\ldots,g_m\in C_b(E_x,\Real)$ satisfy the following for $\delta>0$
\[
    \max_{1\le j\le m}\abs{\nu g_j-\mu g_j}\le\delta
    \quad\Longrightarrow\quad
    \abs{\nu f-\mu f}<\varepsilon.
\]
Since $x^{(n)}\cweak X$, we have $\mu_n g_j\cto\mu g_j$ for each $g_j$.
Hence
\[
    \sum_{n=1}^{\infty}\Prob\bigl(\abs{\mu_n f-\mu f}>\varepsilon\bigr)
    \le \sum_{j=1}^{m}\sum_{n=1}^{\infty}
    \Prob\bigl(\abs{\mu_n g_j-\mu g_j}>\delta\bigr)<\infty.
\]
Therefore
\[
    \frac1n\sum_{i=1}^n\psi(x_i^{(n)},y_i^{(n)})
    =\mu_n f\cto\mu f=\Exp\psi(X,h(X)).
\]
Since $\psi$ was arbitrary, $(x^{(n)},y^{(n)})\cweak(X,h(X))$.
\end{proof}

\begin{lemma}[Gebelein's inequality]\label{lemma:Gebelein inequality} Suppose $A,B$ are jointly Gaussian with $\Exp A=0_E$ and $\Exp B=0_{E'}$ and denote $\Sigma_a=\Exp A^{\otimes 2},\Sigma_b=\Exp B^{\otimes 2}$ and $\rho=\Exp A\otimes B\in L(E;E')$. For $f:E\to E''$ such that $\norm{\Exp f(A)^{\otimes 2}}_{\sf op}<\infty$ let $g(B)=\Exp[f(A)|B]$. Then 
\begin{align}\label{eq:conditional_cov}
\Cov(g(B))\preceq\norm{\Sigma_a\isqrtm\Exp (A\otimes B)\Sigma_b\isqrtm}_{\sf op}\Cov(f(A))
\end{align}
\end{lemma}
\begin{proof} 
By whitening $A\mapsto A\Sigma_a\isqrtm$ and $B\mapsto B\Sigma_b\isqrtm$, it suffices to consider identity marginal covariances. Since \(\Exp V=\Exp U\), we may also center \(f\). Conditioning on \(B\) then yields,
\[
\Cov(V)
=\Exp[g(B)\otimes g(B)]
=\Exp[f(A)\otimes g(B)].
\]
Choose singular-vector coordinates for \(\rho\), with positive singular values \(s_1,\ldots,s_r\). For each \(i\le r\), the conditional law of \(A_i\) given \(B\) is \(N(s_iB_i,1-s_i^2)\). Hence the normalized Hermite polynomials \(h_k=\mathrm{He}_k/\sqrt{k!}\) satisfy
\[
\mathbb E[h_k(A_i)| B]=s_i^k h_k(B_i).
\]
Write \(h_\alpha(x)=\prod_i h_{\alpha_i}(x_i)\) and expand \(f(A)=\sum_{\alpha\ne0}h_\alpha(A)u_\alpha\), with \(u_\alpha\in E''\). The $\alpha=0$ term can be omitted because $f$ is centered. The coordinates of \(A\) are conditionally independent given \(B\), and those beyond \(r\) are independent of \(B\). Thus
\[
g(B)=\sum_{\substack{\nu\in\mathbb N_0^r\\\nu\ne0}}
\left(\prod_{i=1}^r s_i^{\nu_i}\right)h_\nu(B_1,\ldots,B_r)\,u_{(\nu,0)},
\]
where \((\nu,0)\) extends \(\nu\) by zeros to \(\dim E\) coordinates. By orthonormality,
\begin{align*}
\Cov(V)
&=\sum_{\nu\ne0}\left(\prod_{i=1}^r s_i^{2\nu_i}\right)u_{(\nu,0)}^{\otimes2}\preceq \|\rho\|_{\mathrm{op}}^2\sum_{\alpha\ne0}u_\alpha^{\otimes2}
=\|\rho\|_{\mathrm{op}}^2\operatorname{Cov}(f(A)).
\end{align*}
The inequality uses \(s_i\le\|\rho\|_{\mathrm{op}}\le1\) and \(|\nu|\ge1\).

For the lemma as stated with general marginal covariances, take \(\rho=\Sigma_a^{-1/2}\mathbb E[A\otimes B]\Sigma_b^{-1/2}\), with inverses on the covariance ranges.
\end{proof}

\subsection{on proximal operators}

\begin{proposition}[Derivatives of proximal selections]
\label{proposition:asplund}
Let $\gamma\in\mc E_+$ and let $\psi:E\to(-\infty,+\infty]$ be proper,
lower semicontinuous, and bounded below. For any selection
$\zeta(\cdot,\gamma)$ of $\prox_\psi(\cdot,\gamma)$, define the so-called Asplund function
\[
h_\gamma(v):=\round{\psi+\tfrac12\norm{\cdot}_\gamma^2}^*(v\gamma).
\]
where $^*$ indicates the Fenchel conjugate of the function. Then $h_\gamma$ is finite and convex, and $\nabla h_\gamma(v)=\zeta(v,\gamma)\gamma$ for $\mc L$-almost every $v$.
Consequently, $\zeta(\cdot,\gamma)\in\BVloc(E;E)$ and
$(\Dif\zeta(\cdot,\gamma))\gamma=\Dif^2h_\gamma$ is a locally finite
$\mc E_\ge$-valued measure.
\end{proposition}

\begin{proof}
Completing the square gives
\[
h_\gamma(v)
=\sup_{u\in E}\set{\inner{v,u\gamma}-\psi(u)-\tfrac12\norm{u}_\gamma^2}
=\tfrac12\norm{v}_\gamma^2
 -\inf_{u\in E}\set{\psi(u)+\tfrac12\norm{u-v}_\gamma^2}.
\]
The infimum is finite and attained because the objective is proper,
lower semicontinuous, and coercive. The supremum shows that $h_\gamma$
is convex. Fix $v$ and set $u=\zeta(v,\gamma)$. By proximal optimality, this $u$
attains the supremum defining $h_\gamma(v)$, so
\[
h_\gamma(v)=\inner{v,u\gamma}-\psi(u)-\tfrac12\norm{u}_\gamma^2.
\]
For any $w\in E$, using the same $u$ in the supremum defining
$h_\gamma(w)$ gives
\[
h_\gamma(w)\ge \inner{w,u\gamma}-\psi(u)-\tfrac12\norm{u}_\gamma^2
=h_\gamma(v)+\inner{w-v,u\gamma}.
\]
\[
h_\gamma(w)\ge h_\gamma(v)+\inner{w-v,\zeta(v,\gamma)\gamma}.
\]
Thus $\zeta(v,\gamma)\gamma\in\partial h_\gamma(v)$, the subdifferential set, for every $v$.
At every point of differentiability of $h_\gamma$, this subdifferential
is the singleton $\{\nabla h_\gamma(v)\}$. A finite convex function is
locally Lipschitz and differentiable almost everywhere, and its
distributional Hessian is a locally finite positive semidefinite measure.
Hence $\Dif\zeta(\cdot,\gamma)=(\Dif^2h_\gamma)\gamma^{-1}$, which
also gives $\zeta(\cdot,\gamma)\in\BVloc(E;E)$.
\end{proof}

\begin{lemma}\label{lemma:proximal derivative}
Let $R=X+N$, where $N\sim\mathcal N(0,\tau)$ is independent of $(X,W)$,
and fix $\gamma,\tau\in\mc E_+$. Write $D$ for the Gaussian average of
the full distributional derivative of $G(\cdot,\gamma,w)=\prox_{F_w}(\cdot,\gamma)$:
\[
D=\Exp_{X,W}\!\left[\int_E\varphi_\tau(r-X)\,\Dif G(\cdot,\gamma,W)(\dif r)\right].
\]
Assume the integrals and second moments below are finite. In the
$\SBVloc$ setting, $D=A+C$. We use the distributional Stein identity
$\operatorname{Cov}(N,G(R,\gamma,W))=\tau D$ from \Cref{lemma:stein}.
\begin{enumerate}[label=(\alph*)]
\item \textbf{Strict positivity of averaged proximal derivative.}
\label{lemma:proximal derivative positive}
Suppose that, for almost every $w$, $G(\cdot,\gamma,w)$ is a proximal
selection of a proper, lower semicontinuous, bounded-below function $F_w$
with $\operatorname{aff}(\operatorname{dom}F_w)=E$. Then $D\gamma\succ0$.
\item \textbf{Strict upper bound for averaged proximal derivative.}
\label{lemma:proximal derivative upper bound}
Suppose $D\gamma$ is self-adjoint and
\[
\Exp\!\left[\operatorname{Cov}\bigl(G(R,\gamma,W)\mid X,W\bigr)\right]\prec\tau.
\]
Then $D\gamma\prec\gamma$. In particular, the covariance assumption holds
if $\operatorname{Cov}(G(R,\gamma,W)-X)\prec\tau$.
\end{enumerate}
\end{lemma}

\begin{proof}
\begin{enumerate}[label=(\alph*)]
\item 
Fix such a $w$ and let $\Phi_w$ be the potential in
\Cref{proposition:asplund} for $F_w$. Thus
$(\Dif G(\cdot,\gamma,w))\gamma=\Dif^2\Phi_w\succeq0$.
For $v\ne0$, the positive measure $\partial^2_{vv}\Phi_w$ is nonzero.
Indeed, if it vanished, each smooth mollification of $\Phi_w$ would be
affine on the line $t\mapsto tv$. Passing to the limit gives
$\Phi_w(tv)=a+bt$ for all $t\in\mathbb R$. The definition of $\Phi_w$
then gives, for every $x\in\operatorname{dom}F_w$,
\[
a+bt\ge t\inner{v,x\gamma}-F_w(x)-\tfrac12\norm{x}_\gamma^2
\qquad(t\in\mathbb R).
\]
Letting $t\to\pm\infty$ forces $\inner{v,x\gamma}=b$ throughout
$\operatorname{dom}F_w$, contradicting its full affine hull.
Since $\varphi_\tau$ is everywhere positive,
\[
\inner{v,vD\gamma}
=\Exp\!\left[\int_E\varphi_\tau(r-X)\,
\partial^2_{vv}\Phi_W(\dif r)\right]>0 .
\]

\item Write $Z=G(R,\gamma,W)$ and $Y=Z-\Exp[Z\mid X,W]$. By independence,
$\operatorname{Cov}(N,Y)=\tau D$. Hence
\[
0\preceq\operatorname{Cov}(Y-ND)=\operatorname{Cov}(Y)-D^*\tau D,
\qquad
\operatorname{Cov}(Y)=\Exp[\operatorname{Cov}(Z\mid X,W)]\prec\tau.
\]
Thus $D^*\tau D\prec\tau$. The operator
$S=\gamma^{-1/2}(D\gamma)\gamma^{-1/2}$ is self-adjoint.
If $vS=\lambda v$ for $v\ne0$, then $u=v\gamma^{1/2}$ satisfies
$uD^*=\lambda u$, since $D\gamma=\gamma D^*$. It follows that
$\lambda^2\norm{u}_\tau^2=\norm{uD^*}_\tau^2<\norm{u}_\tau^2$.
Thus $S\prec I$, and conjugating by $\gamma^{1/2}$ yields $D\gamma\prec\gamma$.

Finally, the law of total covariance gives
\[
\operatorname{Cov}(Z-X)
=\Exp[\operatorname{Cov}(Z\mid X,W)]
+\operatorname{Cov}(\Exp[Z\mid X,W]-X).
\]
The last term is positive semidefinite, which proves the final assertion.
\end{enumerate}
\end{proof}

\subsection{on conditional distribution of Haar distributed random matrices under linear constraints}
\label{section:haar conditioning}

\begin{lemma}[Haar conditioning
\texorpdfstring{\cite[Lemmas 3--4]{rangan2019vector}}{}]\label{lemma:haar conditioning}
Let $E$ be a real Euclidean space of dimension $d$, and let $dt<n$.
Regard $A,B\in E^{n\times t}$ as linear operators
$E^t\to\Real^n$ using row-wise inner products. Suppose that
$A^*A=B^*B\in \mc E_\ge^t$, and let
\(
    r=\operatorname{rank}(A)=\operatorname{rank}(B)\leq dt.
\)
Let $U$ be a Haar distributed orthogonal random matrix of size $n\times n$. Choose
$\Pi_{A^\perp},\Pi_{B^\perp}\in\Real^{n\times(n-r)}$
whose columns form orthonormal bases of
null$(A^*)$ and null$(B^*)$, respectively.
Then for all $n$,
\[
    U|\{UB=A\}
    \overset{d}{=}
    U_0+\Pi_{A^\perp}\widetilde U\Pi_{B^\perp}^*,
\]
where
\(
    U_0=AB^\dagger=(A^*)^\dagger B^*
    \in\Real^{n\times n},
\)
and $\widetilde U$ is an independent (of the event $UB=A$) Haar distributed orthogonal random matrix of size $n-r\times n-r$.
\end{lemma}
In the lemma above, conditioning is understood
as the canonical regular conditional distribution of $U$ given $UB$,
evaluated at $A$.

\subsection{The \texorpdfstring{$T$}{} transform and the Bias and Variance functions}

We introduce the following $T$-transform which is closely related to the $\eta$-transform and the Stieltjes transform.
For a non-negative random variable $\Sigma^2$, we define the $T$ transform 
\begin{align}
T_{\Sigma^2}(\lambda) := \Exp\frac{\lambda}{\Sigma^2+\lambda}\quad \lambda>0
\end{align}

We define the closely related bias $B$ and variance $V$ functions
\begin{align*}
B_{\Sigma^2}(\lambda)&:=\Exp\round{\frac{\lambda}{\Sigma^2+\lambda}}^2 = T_{\Sigma^2}(\lambda)-\lambda\cdot T'_{\Sigma^2}(\lambda)\\
V_{\Sigma^2}(\lambda)&:= \Exp\round{\frac{\Sigma}{\Sigma^2+\lambda}}^2 = T_{\Sigma^2}'(\lambda)
\end{align*}

\paragraph{Calculations for the Marchenko Pastur law:}
For a matrix $\Psi\in\Real^{m\times p}$ with the type-2 normalization\footnote{the usual normalization is $\mc N(0,\frac1m)$ which we refer to as the type-1 normalization.} $\Psi_{ij}\overset{\sf i.i.d.}\sim\mc N(0,\frac1p)$, in the proportional asymptotic regime where $m,p\to\infty$ and $\frac{p}{m}\cdot a\in(0,\infty)$, its singular values have a limiting distribution given by the Marchenko Pastur law which satisfies
\begin{align}
T_{\Sigma^2}(\lambda)
&=\frac{a-1-a\lambda+\Delta(\lambda)}{2a},\qquad \Delta(\lambda)=\sqrt{(1+a+a\lambda)^2-4a},\\
B_{\Sigma^2}(\lambda)
&=\frac{1}{2a}\left[
a-1+\frac{(a-1)^2+a(1+a)\lambda}{\Delta(\lambda)}
\right],\\[3pt]
V_{\Sigma^2}(\lambda)
&=\frac12\left[
\frac{1+a+a\lambda}{\Delta(\lambda)}-1
\right].
\end{align}

\end{document}